\documentclass[10pt,twocolumn,letterpaper]{article}

\usepackage{iccv}
\usepackage{times}
\usepackage{epsfig}
\usepackage{graphicx}
\usepackage{amsmath}
\usepackage{amssymb}
\usepackage{subfigure}
\usepackage{amsmath}
\usepackage{amssymb}
\usepackage{booktabs}
\usepackage{color}
\usepackage{makecell}
\usepackage{colortbl}
\usepackage{times}
\usepackage{epsfig}
\usepackage{graphicx}
\usepackage{amsmath}
\usepackage{amssymb}
\usepackage{graphicx}
\usepackage{amsmath}
\usepackage{amssymb}
\usepackage{booktabs}
\usepackage{subfigure}
\usepackage{multirow}
\usepackage{amssymb}

\usepackage{bbding}

\usepackage{colortbl}  
\usepackage{xcolor}
\usepackage{array}   

\definecolor{LightGrey}{rgb}{.9,.9,.9}
\definecolor{White}{rgb}{1.,0.,1.}
\definecolor{first}{rgb}{.8,.0,.0}
\definecolor{second}{rgb}{.0,.6,.0}
\definecolor{third}{rgb}{.0,.0,.8}

\definecolor{nbarrier}{RGB}{255, 120, 50}
\definecolor{nbicycle}{RGB}{255, 192, 203}
\definecolor{nbus}{RGB}{255, 255, 0}
\definecolor{ncar}{RGB}{0, 150, 245}
\definecolor{nconstruct}{RGB}{0, 255, 255}
\definecolor{nmotor}{RGB}{200, 180, 0}
\definecolor{npedestrian}{RGB}{255, 0, 0}
\definecolor{ntraffic}{RGB}{255, 240, 150}
\definecolor{ntrailer}{RGB}{135, 60, 0}
\definecolor{ntruck}{RGB}{160, 32, 240}
\definecolor{ndriveable}{RGB}{255, 0, 255}
\definecolor{nother}{RGB}{139, 137, 137}
\definecolor{nsidewalk}{RGB}{75, 0, 75}
\definecolor{nterrain}{RGB}{150, 240, 80}
\definecolor{nmanmade}{RGB}{213, 213, 213}
\definecolor{nvegetation}{RGB}{0, 175, 0}

\definecolor{car}{rgb}{0.39215686, 0.58823529, 0.96078431}
\definecolor{bicycle}{rgb}{0.39215686, 0.90196078, 0.96078431}
\definecolor{motorcycle}{rgb}{0.11764706, 0.23529412, 0.58823529}
\definecolor{truck}{rgb}{0.31372549, 0.11764706, 0.70588235}
\definecolor{other-vehicle}{rgb}{0.39215686, 0.31372549, 0.98039216}
\definecolor{person}{rgb}{1.        , 0.11764706, 0.11764706}
\definecolor{bicyclist}{rgb}{1.        , 0.15686275, 0.78431373}
\definecolor{motorcyclist}{rgb}{0.58823529, 0.11764706, 0.35294118}
\definecolor{road}{rgb}{1.        , 0.        , 1.        }
\definecolor{parking}{rgb}{1.        , 0.58823529, 1.        }
\definecolor{sidewalk}{rgb}{0.29411765, 0.        , 0.29411765}
\definecolor{other-ground}{rgb}{0.68627451, 0.        , 0.29411765}
\definecolor{building}{rgb}{1.        , 0.78431373, 0.        }
\definecolor{fence}{rgb}{1.        , 0.47058824, 0.19607843}
\definecolor{vegetation}{rgb}{0.        , 0.68627451, 0.        }
\definecolor{trunk}{rgb}{0.52941176, 0.23529412, 0.        }
\definecolor{terrain}{rgb}{0.58823529, 0.94117647, 0.31372549}
\definecolor{pole}{rgb}{1.        , 0.94117647, 0.58823529}
\definecolor{traffic-sign}{rgb}{1.        , 0.        , 0.    }   

\makeatletter
\newcommand{\car@semkitfreq}{3.92}
\newcommand{\bicycle@semkitfreq}{0.03}
\newcommand{\motorcycle@semkitfreq}{0.03}
\newcommand{\truck@semkitfreq}{0.16}
\newcommand{\othervehicle@semkitfreq}{0.20}
\newcommand{\person@semkitfreq}{0.07}
\newcommand{\bicyclist@semkitfreq}{0.07}
\newcommand{\motorcyclist@semkitfreq}{0.05}
\newcommand{\road@semkitfreq}{15.30}  %
\newcommand{\parking@semkitfreq}{1.12}
\newcommand{\sidewalk@semkitfreq}{11.13}  %
\newcommand{\otherground@semkitfreq}{0.56}
\newcommand{\building@semkitfreq}{14.1}  %
\newcommand{\fence@semkitfreq}{3.90}
\newcommand{\vegetation@semkitfreq}{39.3}  %
\newcommand{\trunk@semkitfreq}{0.51}
\newcommand{\terrain@semkitfreq}{9.17} %
\newcommand{\pole@semkitfreq}{0.29}
\newcommand{\trafficsign@semkitfreq}{0.08}
\newcommand{\semkitfreq}[1]{{\csname #1@semkitfreq\endcsname}}

\newcommand{\barrier@nuscenesfreq}{11.79}
\newcommand{\bicycle@nuscenesfreq}{0.18}
\newcommand{\bus@nuscenesfreq}{5.83}
\newcommand{\car@nuscenesfreq}{48.27}
\newcommand{\construction@nuscenesfreq}{1.92}
\newcommand{\motorcycle@nuscenesfreq}{0.54}
\newcommand{\pedestrian@nuscenesfreq}{2.93}
\newcommand{\trafficcone@nuscenesfreq}{0.93}
\newcommand{\trailer@nuscenesfreq}{6.22}
\newcommand{\truck@nuscenesfreq}{20.07}
\newcommand{\driveable@nuscenesfreq}{28.64}
\newcommand{\other@nuscenesfreq}{0.77}
\newcommand{\sidewalk@nuscenesfreq}{6.34}
\newcommand{\terrain@nuscenesfreq}{6.35}
\newcommand{\manmade@nuscenesfreq}{16.10}
\newcommand{\vegetation@nuscenesfreq}{11.08}
\newcommand{\nuscenesfreq}[1]{{\csname #1@nuscenesfreq\endcsname}}

\usepackage[pagebackref=true,breaklinks=true,letterpaper=true,colorlinks,bookmarks=false]{hyperref}

\iccvfinalcopy 

\ificcvfinal\fi

\begin{document}
	
	\title{SurroundOcc: Multi-Camera 3D Occupancy Prediction for Autonomous Driving}
	
	
	\author{
		Yi Wei\textsuperscript{1,2}\thanks{Equal contribution.},
		Linqing Zhao\textsuperscript{3}\footnotemark[1],
		Wenzhao Zheng\textsuperscript{1,2},
		Zheng Zhu\textsuperscript{4}, 
		Jie Zhou\textsuperscript{1,2},
		Jiwen Lu\textsuperscript{1,2}\thanks{Corresponding author.}
		\\ \\
		\textsuperscript{1}Beijing National Research Center for Information Science and Technology, China \\
		\textsuperscript{2}Department of Automation, Tsinghua University, China \\
		\textsuperscript{3}School of Electrical and Information Engineering, Tianjin University, China \\
		\textsuperscript{4}PhiGent Robotics \\
		{\tt\small \{y-wei19,zhengwz18\}@mails.tsinghua.edu.cn; linqingzhao@tju.edu.cn; } \\
		{\tt\small zhengzhu@ieee.org; 
			\tt\small \{jzhou, lujiwen\}@tsinghua.edu.cn} \\
	}

	\maketitle
	\ificcvfinal\thispagestyle{empty}\fi

	\begin{abstract}
		3D scene understanding plays a vital role in vision-based autonomous driving. While most existing methods focus on 3D object detection, they have difficulty describing real-world objects of arbitrary shapes and infinite classes. Towards a more comprehensive perception of a 3D scene, in this paper, we propose a SurroundOcc method to predict the 3D occupancy with multi-camera images. We first extract multi-scale features for each image and adopt spatial 2D-3D attention to lift them to the 3D volume space.
		Then we apply 3D convolutions to progressively upsample the volume features and impose supervision on multiple levels. To obtain dense occupancy prediction, we design a pipeline to generate dense occupancy ground truth without expansive occupancy annotations. Specifically, we fuse multi-frame LiDAR scans of dynamic objects and static scenes separately. Then we adopt Poisson Reconstruction to fill the holes and voxelize the mesh to get dense occupancy labels. Extensive experiments on nuScenes and SemanticKITTI datasets demonstrate the superiority of our method. Code and dataset are available at \href{https://github.com/weiyithu/SurroundOcc}{\color{cyan}{https://github.com/weiyithu/SurroundOcc}}.

	\end{abstract}
	
	\section{Introduction}
	\label{sec:intro}
	Understanding the 3D geometry of the surrounding scene serves as the basic procedure in an autonomous driving system.
	While LiDAR is a direct solution to capture this geometric information, it suffers from high-cost sensors and sparse scanned points, limiting its further application.
	Recently, vision-centric autonomous driving has attracted extensive attention as a promising direction.
	Taking multi-camera images as inputs, it has demonstrated competitive performance in various 3D perception tasks including depth estimation~\cite{guizilini2022full,wei2022surrounddepth}, 3D object detection~\cite{bevformer,bevdepth,bevfusion,liu2022petr,bevdet}, and semantic map construction~\cite{beverse,fiery,stretchbev}.
	
	\begin{figure}[tb]
		\centering
		\includegraphics[width=0.95\linewidth]{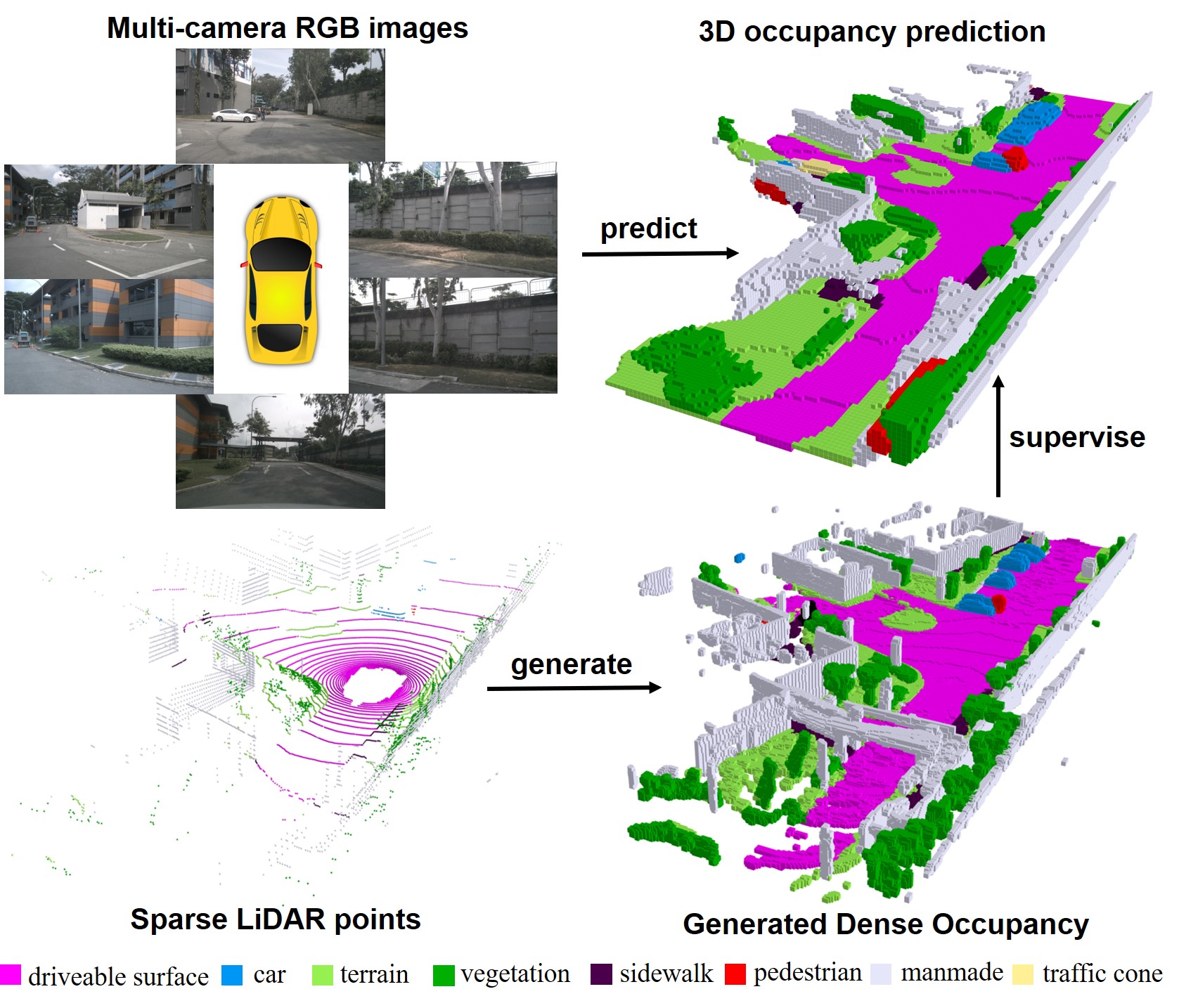}
		\caption{The overview of SurroundOcc. Given multi-camera images, our method can predict volumetric occupancy of surrounding 3D scenes. To train the network, we design a pipeline to generate dense occupancy labels with sparse LiDAR points. \textbf{Better viewed when zoomed in.}}
		\label{fig:qualitative}
		\vspace{-5mm}
	\end{figure}

	Although multi-camera 3D object detection plays an important role ~\cite{bevformer,bevdepth,liu2022petr,bevdet} in vision-based 3D perception, it is easy to suffer from the long-tail problem and difficult to recognize all classes of objects in the real world. Complementary to 3D detection, reconstructing surrounding 3D scenes can better help the downstream perception tasks. Recent works~\cite{guizilini2022full, wei2022surrounddepth} incorporate information from multiple views and predict surrounding depth maps. However, depth maps only predict the nearest occupied point in each optical ray and are unable to recover the occluded parts of the 3D scene. Different from depth-based methods, another trend \cite{monoscene,huang2023tri} is to directly predict the 3D occupancy of the scene. It describes a 3D scene by assigning an occupied probability to each voxel in the 3D space. We advocate 3D occupancy to be a good 3D representation for multi-camera scene reconstruction, which naturally guarantees the multi-camera geometry consistency and is able to recover occluded parts. Also, it is flexible to extend to other 3D downstream tasks such as 3D semantic segmentation~\cite{cylinder3d,cylindrical,rangedet}. As one of the pioneering works, MonoScene \cite{monoscene} infers the dense 3D voxelized semantic scene with monocular images. However, simply fusing multi-camera results with cross-camera post-processing will lead to low performance \cite{bevformer}. TPVFormer \cite{huang2023tri} uses sparse LiDAR points as supervision, which results in sparse occupancy prediction.

	To address this, we propose a SurroundOcc method, which aims to predict dense and accurate 3D occupancy with multi-camera images input. We first use a 2D backbone network to extract multi-scale feature maps from each image. Then we perform 2D-3D spatial attention to lift multi-camera image information to 3D volume features instead of BEV features. A 3D convolution network is then employed to progressively upsample the low-resolution volume features and fuse them with high-resolution ones to obtain fine-grained 3D representations. At each level, we use a decayed weighted loss to supervise the network. To get dense predictions, we need dense occupancy labels. However, 
	the mainstream multi-camera dataset nuScenes \cite{caesar2020nuscenes} only provides sparse LiDAR points. To avoid expensive occupancy annotations, we devise a pipeline to generate dense occupancy ground truth only with the existing 3D detection and 3D semantic segmentation labels. Specifically, we first combine multi-frame points of dynamic objects and static scenes respectively. Then we leverage Poisson Reconstruction \cite{kazhdan2006poisson} algorithm to further fill the holes. Finally, NN and voxelization are used to obtain dense 3D occupancy labels. 
	
	With the dense occupancy ground truth, we train the model and compare it with other state-of-the-art methods on nuScenes \cite{caesar2020nuscenes} dataset. Both the quantitative results and visualizations demonstrate the effectiveness of our method. Moreover, we further conduct experiments on SemanticKITTI dataset \cite{behley2019semantickitti}. Although our method is not designed for the monocular setting, it achieves state-of-the-art performance on the monocular 3D semantic scene completion benchmark.

	\section{Related Work}

	\textbf{Voxel-based Scene Representation:}
	How to effectively represent a 3D scene lies at the core of autonomous driving perception. Voxel-based scene representation voxelizes the 3D space into discretized voxels and describes each voxel by a vector feature. It has empowered the success of numerous methods on the lidar segmentation~\cite{amvnet,spvnas,af2s3net,drinet++,ye2022lidarmultinet} and 3D scene completion~\cite{monoscene,lmscnet,dsketch,aicnet,js3c,li2023voxformer, li2023sscbench} tasks. For the 3D occupancy prediction task, we also advocate the voxel representation as it is more suitable to model the occupancy field of a 3D scene. MonoScene~\cite{monoscene} is the first work to reconstruct outdoor scenes using only RGB inputs. TPVFormer~\cite{huang2023tri} further generalizes it to multi-camera 3D semantic occupancy prediction. However, its lack of dense supervision results in sparse occupancy prediction. Differently, we devise a pipeline to generate dense occupancy ground truth for training and our occupancy prediction is much denser.

	\textbf{3D Scene Reconstruction:}
	3D reconstruction \cite{peng2020convolutional,song2017semantic,saito2019pifu,park2019deepsdf,zhou2017unsupervised,godard2019digging,bian2019unsupervised,atlas,neuralrecon,transformerfusion} is a traditional but important topic in computer vision. 
	One way is through depth estimation which predicts a depth value for each pixel in the image.
	While early methods require full depth annotations to supervise the depth estimation model~\cite{eigen2015predicting, lee2019big, zhang2018progressive}, later research focuses on self-supervised depth estimation as it does not require intensive human annotations~\cite{zhou2017unsupervised,godard2019digging,bian2019unsupervised,zhao2020towards,yin2018geonet,wang2017orientation,zhou2019moving,ranjan2019competitive,chang2018pyramid}. Recently, SurroundDepth~\cite{wei2022surrounddepth} further incorporates the interactions between surrounding views to capture more spatial correlations. 
	Different from depth estimation, 3D scene reconstruction methods ~\cite{surfacenet,atlas,neuralrecon,transformerfusion,monoscene} directly reconstruct a comprehensive and accurate 3D geometry of a scene. 
	SurfaceNet~\cite{surfacenet} employs a 3D convolutional network to transform RGB colors to 3D surface occupancy from two images.
	Atlas~\cite{atlas} further extends it to the multi-view setting and utilized learned features to predict occupancy.
	NeuralRecon~\cite{neuralrecon} and TransformerFusion~\cite{transformerfusion} fuse the learned image features from different views in an online manner for more accurate 3D reconstructions. However, most of these 3D scene reconstruction methods are designed for indoor scenes, which are different from the multi-camera setting in the outdoor environment.

	\textbf{Vision-based 3D Perception:}
	The lack of direct geometric input demands vision-based 3D surround perception methods to infer the 3D scene geometry based on semantic cues.
	Depth-based methods explicitly predict depth maps for image inputs to extract 3D geometric information of the scene before perception~\cite{pseudo-lidar,AM3D,caddn,dd3d,bevdepth,bevfusion,bevfusion2,lss,bevdet,beverse}.
	The widely adopted pipeline is to predict categorical depth distributions and leverage them to project semantic features into 3D space~\cite{lss}.
	Other methods implicitly learn 3D features without producing explicit depth maps~\cite{fcos3d,bevformer,detr3d,petr,PGD,monopair}.
	For example, BEVFormer~\cite{bevformer} adopts cross attention to progressively refine BEV grid features from 2D image features. While most existing works employ BEV representation to describe a scene, we propose to reconstruct a 3D scene using volumetric occupancy representation, which provides the more fine-grained and comprehensive modeling of the scene.

	\begin{figure*}[t]
		\centering
		\includegraphics[width=0.95\linewidth]{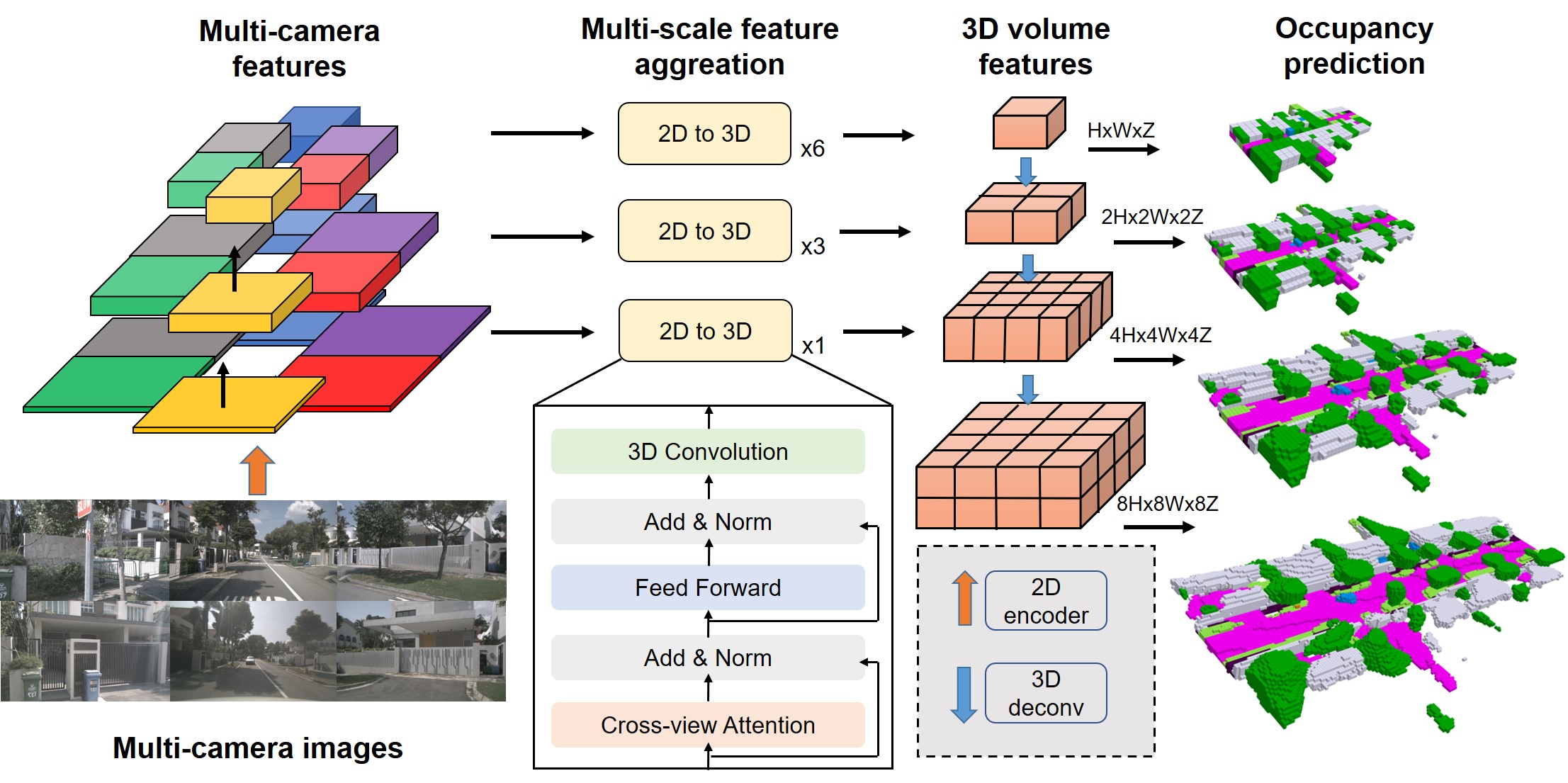}
		\caption{The pipeline of the proposed method. First, we use a backbone to extract multi-scale features of multi-camera images. Then we adopt 2D-3D spatial attention to fuse multi-camera information and construct 3D volume features in a multi-scale fashion. Finally, the 3D deconvolution layer is used to upsample 3D volumes and occupancy prediction is supervised in each level.
		}
		\vspace{-1mm}
		\label{fig:pipeline}
	\end{figure*}
	
	\begin{figure}[t]
		\centering
		\includegraphics[width=0.9\linewidth]{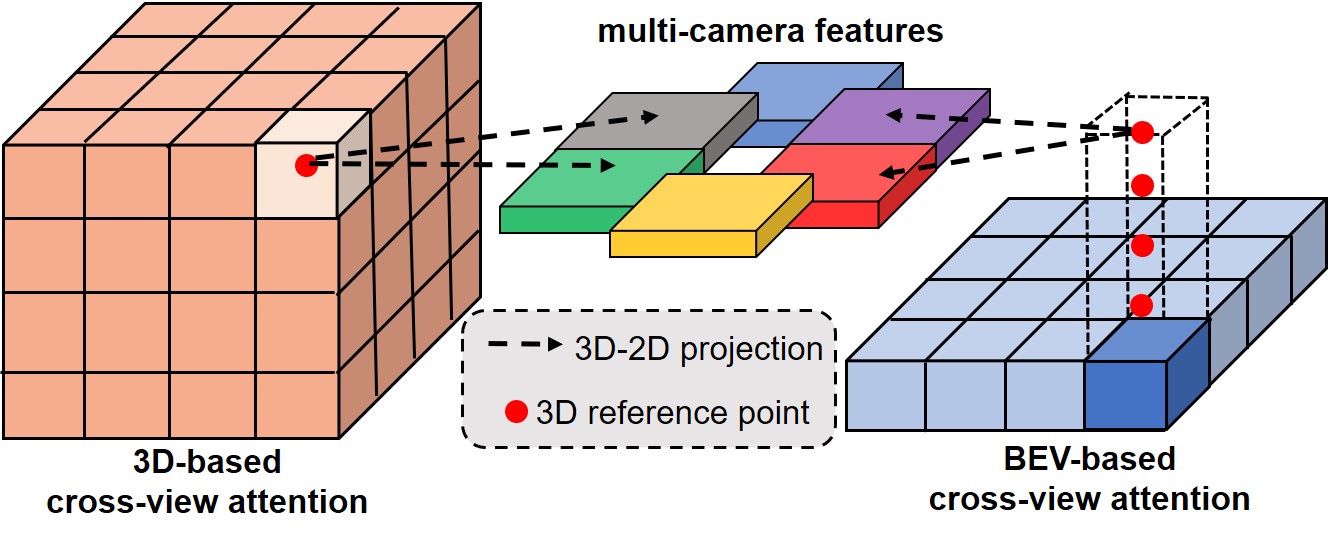}
		\caption{The comparison of 3D-based and BEV-based cross-view attention. The 3D-based attention can better preseve 3D information. For each 3D volume query, we project it to the corresponding 2D views to sample features. 
		}
		\vspace{-3mm}
		\label{fig:spatial}
	\end{figure}
	
	\section{Approach}
	\subsection{Problem Formulation}
	In this work, we aim to predict the 3D occupancy of surrounding scenes with multi-camera images ${I} = \{{I}^{1}, {I}^{2}, \cdots  {I}^{N}\}$. Formally, the 3D occupancy predcition is represented as:
	\begin{equation}
		V = G({I}^{1}, {I}^{2}, \cdots  {I}^{N})
	\end{equation}
	where $G$ is an neural network and $V \in  \mathbb{R}^{H \times W \times Z}$ is the 3D occupancy. The value of $V$ is between 0 and 1, representing the occupied probability of the grids. Lifting $V$ to an $(L, H, W, Z)$ tensor, we can obtain the 3D semantic occupancy, where $L$ is the class number and class 0 means non-occupied grids. 
	
	3D occupancy is a good representation for multi-camera 3D scene reconstruction. First, since 3D occupancy is predicted in 3D space, it  theoretically satisfies the multi-camera consistency. Second, it is possible for networks to predict occluded areas according to the surrounding semantic information, which is unavailable in depth estimation. Third, 3D occupancy is easy to extend to other downstream tasks, such as 3D semantic segmentation and scene flow estimation.

	\begin{figure*}[t]
		\centering
		\includegraphics[width=0.95\linewidth]{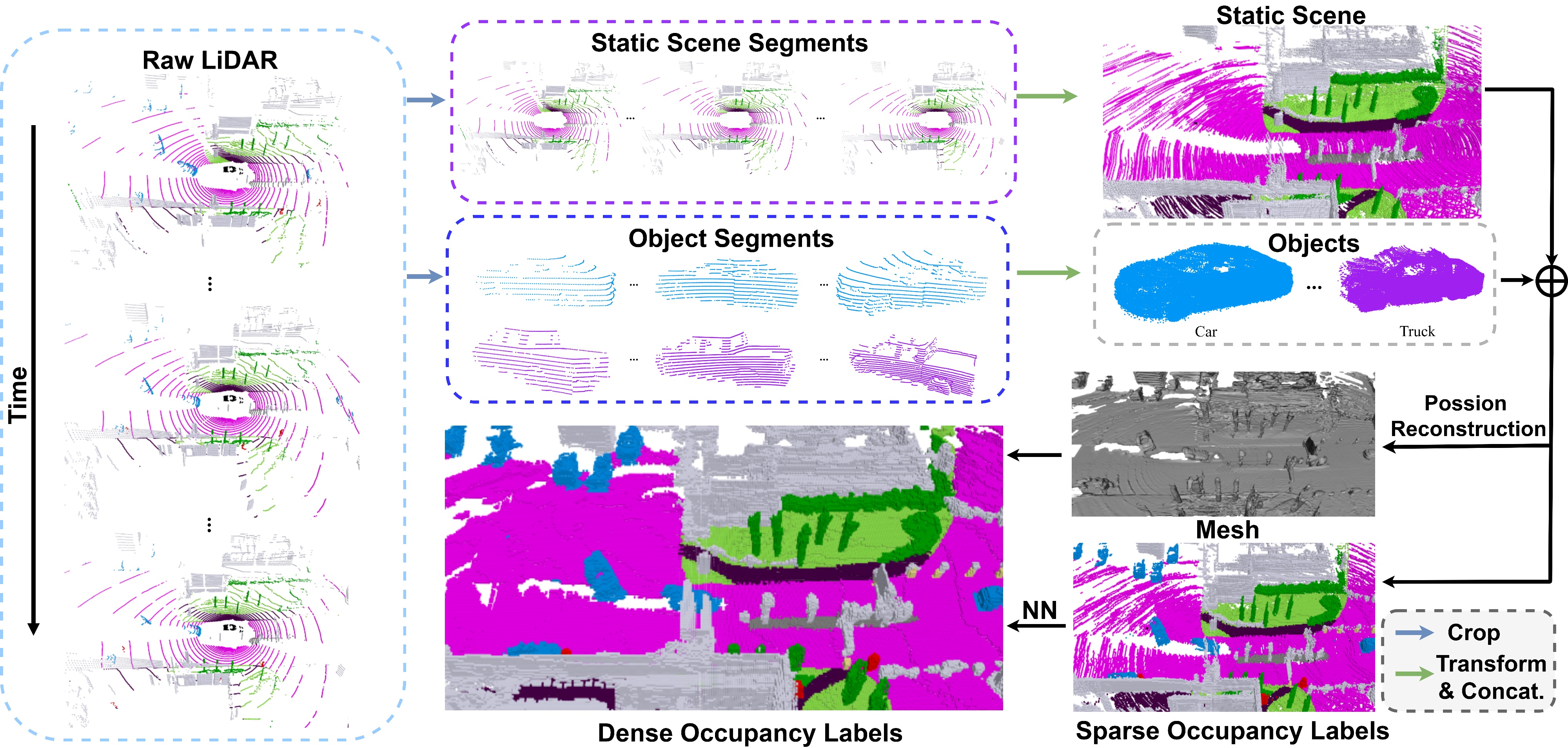}
		\caption{Dense occupancy ground truth generation. We first traverse all frames to stitch the multi-frame LiDAR points of dynamic objects and static scenes separately, and then merge them into a complete scene. Subsequently, we employ Poisson Reconstruction to densify the points and voxelize the resulting mesh to obtain a dense 3D occupancy. Finally, we use the Nearest Neighbor (NN) algorithm to assign semantic labels to dense voxels. }

		\vspace{-3mm}
		\label{fig:gt_pipeline}
	\end{figure*}
	
	\subsection{Overview}
	Figure \ref{fig:pipeline} shows the pipeline of our method. Given a set of surrounding multi-camera images, we first use a backbone network (\eg ResNet-101 \cite{he2016deep}) to extract $N$ cameras' and $M$ levels' multi-scale features $X = \{\{X_i^j\}_{i=1}^{N}\}_{j=1}^{M}$. For each level, we use a transformer to fuse multi-camera features with spatial cross attention. The output of the 2D-3D spatial attention layer is a 3D volume feature instead of the BEV feature. Then the 3D convolution network is utilized to upsample and combine multi-scale volume features. The occupancy prediction in each level is supervised by the generated dense occupancy ground truth with a decayed loss weight.
	
	\subsection{2D-3D Spatial Attention}
	Many 3D scene reconstruction methods \cite{monoscene,atlas} integrate multi-view 2D features into 3D space by reprojecting 2D features back to the 3D volumes with known poses. The grid feature is calculated by simply averaging all 2D features in this grid. However, this kind of method assumes that different views contribute equally to the 3D volumes, which does not always hold true, especially when some views are occluded or blurred. 
	
	To tackle the issue, we leverage cross-view attention to fuse multi-camera features. We project 3D reference points to 2D views and use deformable attention \cite{zhu2020deformable,bevformer} to query points and aggregate information. As shown in Figure \ref{fig:spatial}, instead of 2D BEV queries, we build 3D volume queries to further reserve 3D space information. Specifically,  3D volume queries are defined as $Q \in  \mathbb{R}^{C \times H \times W \times Z}$. For each query, we project its corresponding 3D point to 2D views according to the given intrinsic and extrinsic. We
	only use the views that the 3D reference point hits. Then we sample 2D features around these projected 2D positions. The output $F \in  \mathbb{R}^{C \times H \times W \times Z}$ of the cross-view attention layer is a weighted sum of sampled features according to the deformable attention mechanism:
	\begin{equation}
		\begin{aligned}
			\text{DeformAttn}(q, p, x) &= \sum_{i=1}^{N_\text{head}} \mathcal{W}_i\sum_{j=1}^{N_\text{key}} \mathcal{A}_{ij} \cdot \mathcal{W}'_i x(p+ \Delta p_{ij})  \\
			F^p = \frac{1}{|\mathcal{V}_\text{hit}|} \sum\limits_{i\in \mathcal{V}_\text{hit}} &\text{DeformAttn}(Q^p, \mathcal{P}(q^p,i), X_i)
		\end{aligned}  
	\end{equation}
	where $F^p$ and $Q^p$ indicate the $p$th element of the output features and 3D volume queries. $q^p$ is the corresponding 3D positions of queries and $P$ is the 3D to 2D project function. $\mathcal{V}_\text{hit}$ represents the hit views of 3D query points. $W_i$ and $W'_i$ are the learnable weights and $A_{ij}\!\in\![0,1]$ is the attention weight calculated by the dot product of query and key. $x(p + \Delta p_{ij})$ is the 2D feature at location $p + \Delta p_{ij}$. Instead of performing expensive 3D self-attention, we use the 3D convolution to interact features between neighboring 3D voxels.

	\subsection{Multi-scale Occupancy Prediction}
	We further extend the 2D-3D spatial attention to a multi-scale fashion. Different from 3D detection task, 3D scene reconstruction needs more low-level features to help the network learn fine-grained details. To tackle the issue, we design a 2D-3D U-Net architecture. Specifically, given multi-scale 2D features $\{\{X_i^j\}_{i=1}^{N}\}_{j=1}^{M}$, we adopt different number of 2D-3D spatial attention layers to extract multi-scale 3D volume features $\{F_j \in  \mathbb{R}^{C_j \times H_j \times W_j \times Z_j}\}_{j=1}^{M}$. Then we upsample $j-1$th level 3D volume features $Y_{j-1}$ with 3D deconvolution layer and fuse it with $F_j$: 
	
	\begin{equation}
		Y_j = F_j + \text{Deconv}(Y_{j-1})
	\end{equation}

	For each level, the network outputs an occupancy prediction result with different resolution $V_j \in \mathbb{R}^{C_j \times H_j \times W_j \times Z_j}$. To get powerful both high-level and low-level 3D features, the network is supervised at each scale. Specifically, we use the cross-entropy loss and scene-class affinity loss introduced in \cite{monoscene} as the supervision signals. For 3D semantic occupancy prediction, we adopt a multi-class cross-entropy loss and for 3D scene reconstruction we change it to a two-class formulation. Since the high-resolution prediction is more important, we use a decayed loss weight $\alpha_j = \frac{1}{2^j}$ for $j$th level supervision.

	\section{Dense Occupancy Ground Truth}
	
	\begin{figure}[t]
		\centering
		\subfigure[RGB image]{
			\includegraphics[width=0.45\linewidth]{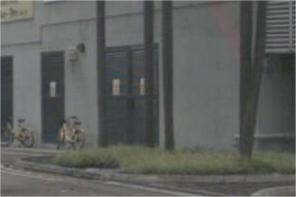}
		}
		\subfigure[single-frame LiDAR points]{
			\includegraphics[width=0.45\linewidth]{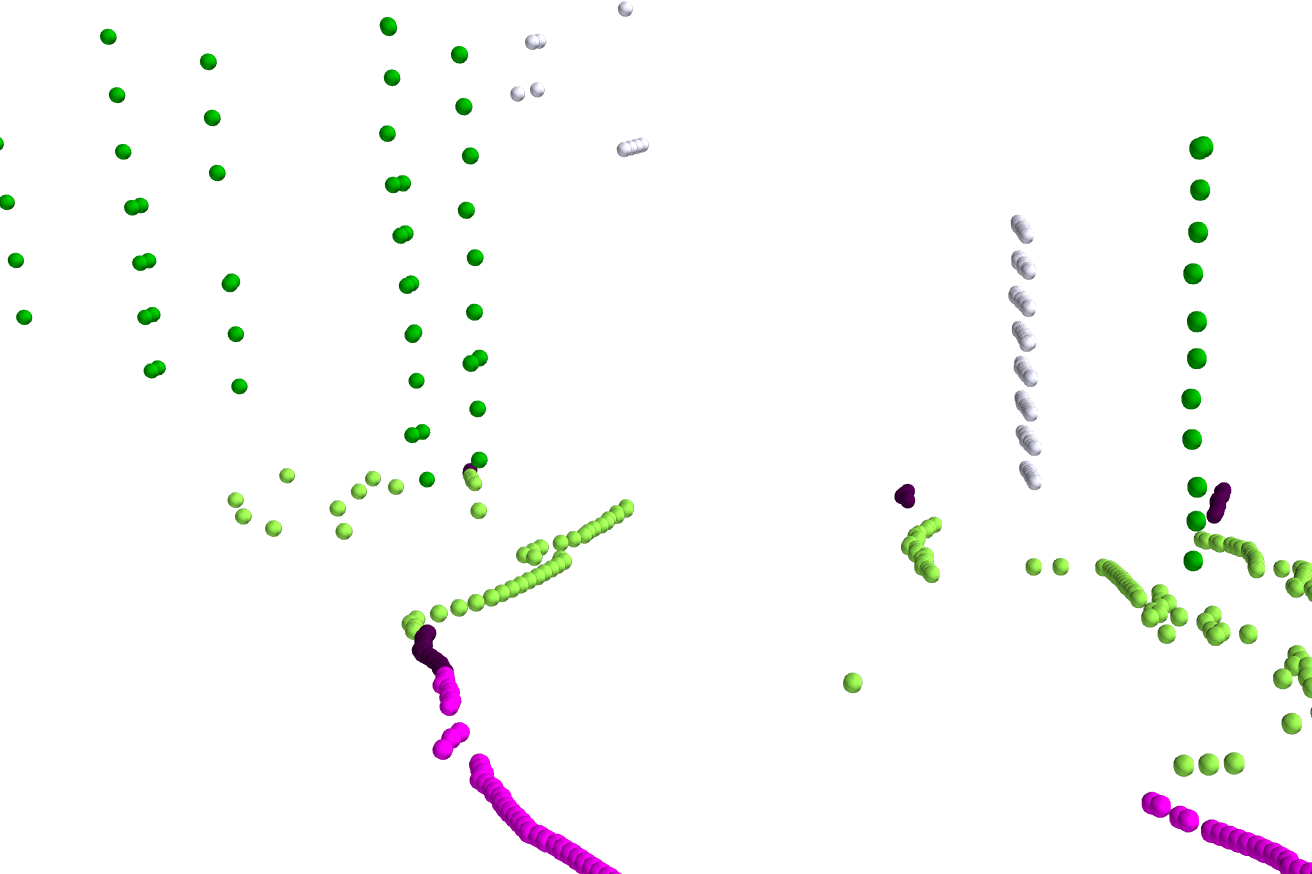}
		}\\
		\subfigure[Sparse occupancy labels]{
			\includegraphics[width=0.45\linewidth]{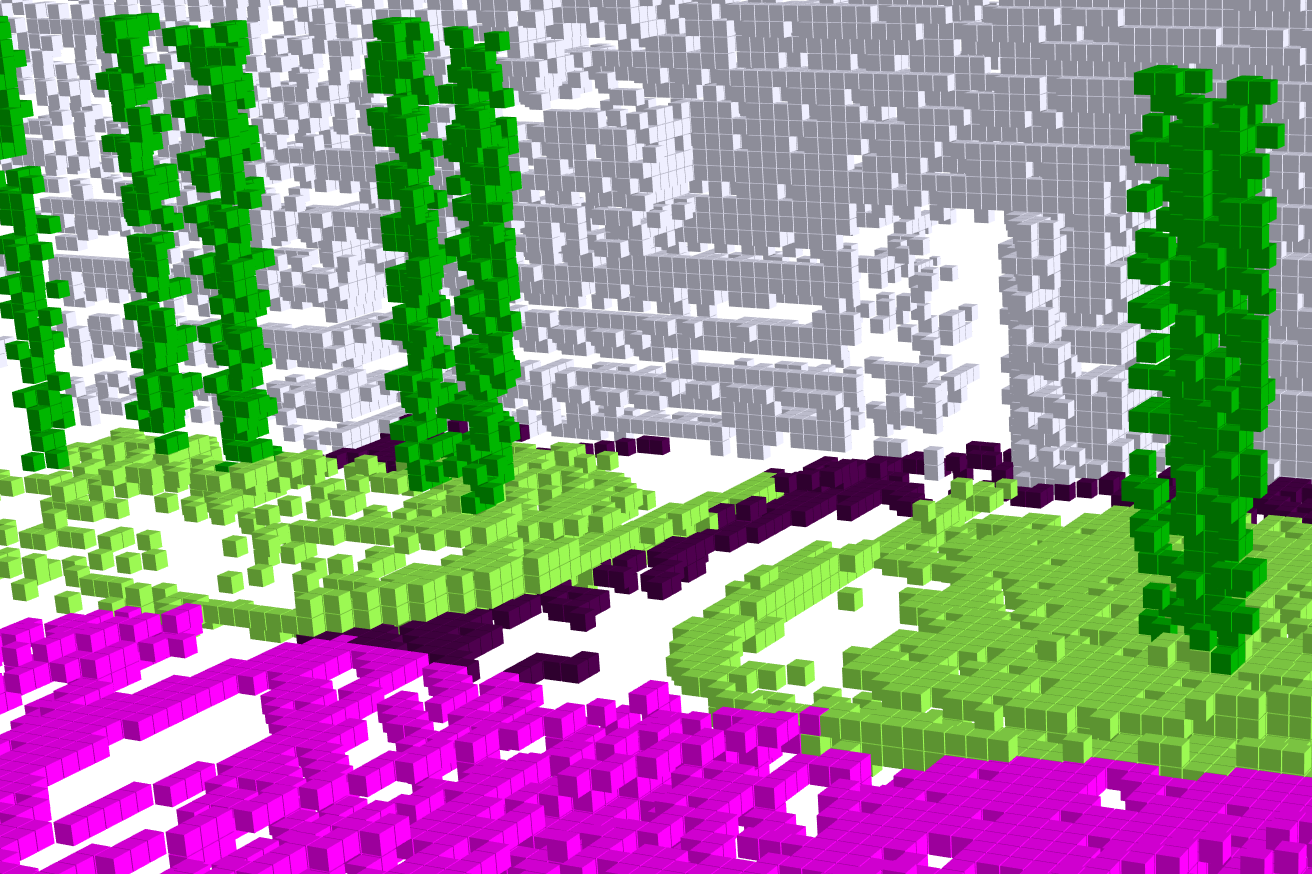}
		}
		\subfigure[Dense occupancy labels]{
			\includegraphics[width=0.45\linewidth]{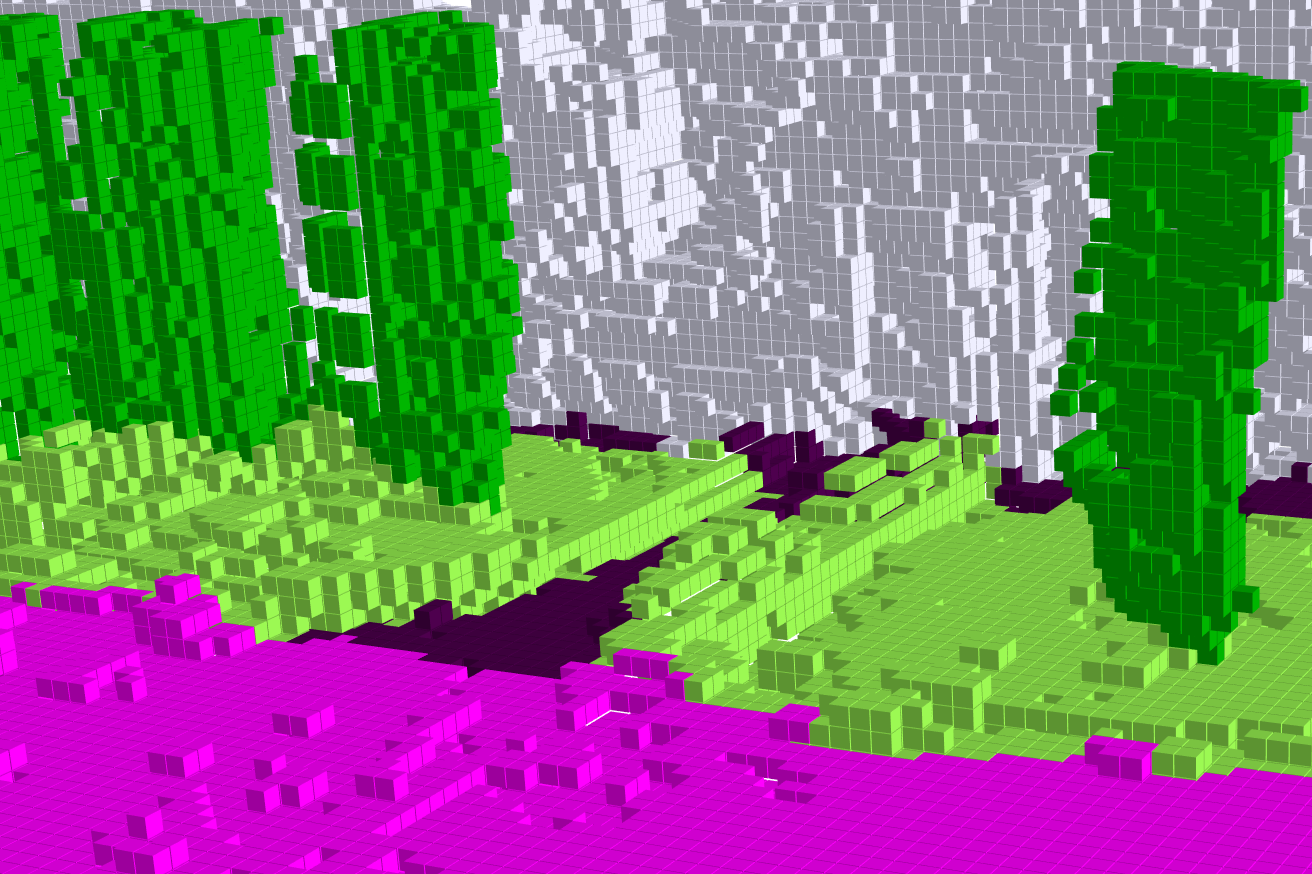}
		}
		\caption{Comparison on different occupancy labels. Compared with single-frame LiDAR points and the sparse occupancy converted from multi-frame points, our dense voxels are able to provide more realistic occupancy labels.
		} 
		\vspace{-3mm}
		\label{fig:detail_comparison}
	\end{figure}
	
	
	In our experiments, we find that the network supervised by sparse LiDAR points is unable to predict dense enough occupancy. Thus, it is necessary to generate dense occupancy labels. However, as mentioned in SemanticKITTI \cite{behley2019semantickitti}, it is complex and needs huge human efforts to annotate the dense occupancy of a 3D scene which has millions of voxels. To this end, we design a pipeline to generate dense occupancy ground truth leveraging existing 3D detection and 3D semantic segmentation labels without extra human annotations, which is shown in Figure~\ref{fig:gt_pipeline}. An intuitive way is to directly transform a multi-frame LiDAR point sequence into a unified coordinate system, and then voxelize the concatenated dense points into voxel grids. However, such a straightforward solution is only applicable to completely static scenes and ignores moving objects. Moreover, multi-frame point clouds are not dense enough and there still exist many holes, which will result in wrong occupancy labels. To address these issues, we propose to stitch multi-frame LiDAR points of dynamic objects and static scenes separately. In addition, we adopt Poisson Reconstruction \cite{kazhdan2006poisson} to fill up the holes and voxelize the obtained mesh to get dense volumetric occupancy. Since the LiDAR scans surface points, our method also generates surface occupancy. 
	
	\subsection{Multi-frame Point Cloud Stitching}
	We propose a two-stream pipeline to stitch static scenes and objects separately and merge them into a complete scene before voxelization. Specifically, for each frame, we first cut out movable objects from the LiDAR points according to 3D bounding box labels, so that we can obtain the 3D points of a static scene and movable objects. After traversing all frames in the scene, we integrate the collected static scene segments and object segments into a set respectively. To combine the multi-frame segments, we then transform their coordinates into the world coordinate system via the known calibrated matrices and ego-poses. We denote the transformed static scene segments and object segments as $P_{ss} = \{P^{1}_{ss}, P^{2}_{ss}, \cdots P^{n}_{ss}\}$ and $P_{os} = \{P^{1}_{os}, P^{2}_{os}, \cdots P^{m}_{os}\}$, where $n$ and $m$ are the numbers of frames and objects in the sequence, respectively. Note that the same objects in different frames can be recognized according to the bounding box index. Therefore, we can represent the whole static scene as $P_{s} = [P^{1}_{ss}, P^{2}_{ss}, \cdots P^{n}_{ss}]$ while the objects as $P_{o} = [P^{1}_{os}, P^{2}_{os}, \cdots P^{m}_{os}]$, respectively, where $[ \cdot ]$ is the concatenation operator.
	Finally, according to the objects' locations and ego-pose of the current frame, the 3D points of this frame can be obtained by merging static scene and objects: $P = [T_s(P{s}),T_o(P{o})]$, where $T_s$ and $T_o$ are the transformations of static scenes and objects from world coordinate system to the current frame coordinate system. In this way, the occupancy labels of the current frame leverage the LiDAR points of all frames in the sequence. 
	
	\subsection{Densifying with Poisson Reconstruction}
	While the density of $P$ is much larger than a single-frame LiDAR, there still exists many interspaces and the points are not evenly distributed, which is caused by the limited LiDAR beams. To address this, we first compute the normal vectors according to the spatial distribution in local neighborhoods. Then we reconstruct $P$ to a triangular mesh $\mathcal{M}$ via Poisson Surface Reconstruction~\cite{kazhdan2006poisson}, whose input is point cloud with normal vectors, and the output is a triangular mesh (see Figure~\ref{fig:gt_pipeline}). The obtained mesh $\mathcal{M} = \{\mathcal{V},\mathcal{E}\}$ fills up the holes of point clouds with evenly distributed vertices $\mathcal{V}$, so that we can further convert the mesh into dense voxel $V_d$.

	\begin{table*}[t]
		\footnotesize
		\setlength{\tabcolsep}{0.006\linewidth}
		
		\newcommand{\classfreq}[1]{{~\tiny(\nuscenesfreq{#1}\%)}}  %
		\centering
		\begin{tabular}{l|c c | c c c c c c c c c c c c c c c c}
			\toprule
			Method
			&  \makecell{SC\\ IoU} & \makecell{SSC \\ mIoU}
			& \rotatebox{90}{\textcolor{nbarrier}{$\blacksquare$} barrier}
			& \rotatebox{90}{\textcolor{nbicycle}{$\blacksquare$} bicycle}
			& \rotatebox{90}{\textcolor{nbus}{$\blacksquare$} bus}
			& \rotatebox{90}{\textcolor{ncar}{$\blacksquare$} car}
			& \rotatebox{90}{\textcolor{nconstruct}{$\blacksquare$} const. veh.}
			& \rotatebox{90}{\textcolor{nmotor}{$\blacksquare$} motorcycle}
			& \rotatebox{90}{\textcolor{npedestrian}{$\blacksquare$} pedestrian}
			& \rotatebox{90}{\textcolor{ntraffic}{$\blacksquare$} traffic cone}
			& \rotatebox{90}{\textcolor{ntrailer}{$\blacksquare$} trailer}
			& \rotatebox{90}{\textcolor{ntruck}{$\blacksquare$} truck}
			& \rotatebox{90}{\textcolor{ndriveable}{$\blacksquare$} drive. suf.}
			& \rotatebox{90}{\textcolor{nother}{$\blacksquare$} other flat}
			& \rotatebox{90}{\textcolor{nsidewalk}{$\blacksquare$} sidewalk}
			& \rotatebox{90}{\textcolor{nterrain}{$\blacksquare$} terrain}
			& \rotatebox{90}{\textcolor{nmanmade}{$\blacksquare$} manmade}
			& \rotatebox{90}{\textcolor{nvegetation}{$\blacksquare$} vegetation}
			\\
			\midrule
			MonoScene \cite{monoscene} & 23.96 & 7.31 & 4.03 &	0.35& 8.00& 8.04&	2.90& 0.28& 1.16&	0.67&	4.01& 4.35&	27.72&	5.20& 15.13&	11.29&	9.03&	14.86 \\
			
			Atlas \cite{atlas} & 28.66 & 15.00 & 10.64&	5.68&	19.66& 24.94& 8.90&	8.84&	6.47& 3.28&	10.42&	16.21&	34.86&	15.46&	21.89&	20.95&	11.21&	20.54
			\\
			
			BEVFormer \cite{bevformer} & 30.50 & 16.75 & 14.22 &	6.58 & 23.46 & 28.28& 8.66 &10.77& 6.64& 4.05& 11.20&	17.78 & 37.28 & 18.00 & 22.88 & 22.17 & 13.80 &	22.21
			\\
			
			TPVFormer \cite{huang2023tri} & 11.51 & 11.66 & 16.14&	7.17& 22.63	& 17.13 & 8.83 & 11.39 & 10.46 & 8.23&	9.43 & 17.02 & 8.07 & 13.64 & 13.85 & 10.34 & 4.90 & 7.37
			
			\\
			
			TPVFormer*  & 30.86 & 17.10 & 15.96&	 5.31& 23.86	& 27.32 & 9.79 & 8.74 & 7.09 & 5.20& 10.97 & 19.22 & \textbf{38.87} & 21.25 & 24.26 & \textbf{23.15} & 11.73 & 20.81
			
			\\

			\midrule

			SurroundOcc & \textbf{31.49} & \textbf{20.30}  & \textbf{20.59} & \textbf{11.68} & \textbf{28.06} & \textbf{30.86} & \textbf{10.70} & \textbf{15.14} & \textbf{14.09} & \textbf{12.06} & \textbf{14.38} & \textbf{22.26} & 37.29 & \textbf{23.70} & \textbf{24.49} & 22.77 & \textbf{14.89} & \textbf{21.86}  \\ %
			
			\bottomrule
		\end{tabular}
		
		\vspace{2mm}
		\caption{\textbf{3D semantic occupancy prediction results on nuScenes validation set.} Except TPVFormer \cite{huang2023tri}, all methods are trained with dense occupancy labels. To fairly compare, we further use dense ground truth to train the TPVFormer, which is denoted as TPVFormer*.
		}
		\label{tab:nuscseg}
		\vspace{-2mm}
	\end{table*}
	
	\begin{table}[]
		\centering
		\resizebox{0.4\textwidth}{!}{
			\begin{tabular}{l|l}
				\hline
				Acc & $\mbox{mean}_{p \in P}(\min_{p^*\in P^*}||p-p^*||)$ \\
				Comp & $\mbox{mean}_{p^* \in P^*}(\min_{p\in P}||p-p^*||)$ \\
				Prec & $\mbox{mean}_{p \in P}(\min_{p^*\in P^*}||p-p^*||<0.5)$ \\
				Recal & $\mbox{mean}_{p^* \in P^*}(\min_{p\in P}||p-p^*||<0.5)$ \\
				CD & $\text{Acc} + \text{Comp}$ \\
				F-score & $(2 \times \text{Prec} \times \text{Recal}) / (\text{Prec} + \text{Recal})$ \\
				\hline
		\end{tabular}}
		\vspace{1mm}
		\caption{Evaluation metrics for 3D scene reconstruction. $p$ and $p^*$ are the predicted and ground truth point clouds.
		}
		\vspace{-3mm}
		\label{tab:metric_defs}
	\end{table}
	
	\subsection{Semantic Labeling with NN Algorithm}
	Having obtained the occupancy of dense voxels $V_d$, we aim to assign semantic labels to each voxel, since position reconstruction can only be applied to 3D space, not semantic space. To this end, we propose to leverage Nearest Neighbors (NN) algorithm to search the nearest semantic label for each voxel. Specifically, we first voxelize $P$ with semantics into voxels $V_s$, which are sparser than $V_d$ due to limited LiDAR beams. Then for each occupied voxel in $V_d$, we use NN to search the nearest voxel in $V_s$ and assign the semantic label to it. Note that this process can be accelerated by parallel computing on the GPU. Thus, all occupied voxels in $V_d$ can obtain their semantic labels from $V_s$.
	
	Figure~\ref{fig:detail_comparison} shows a detailed visual comparison between single-frame LiDAR points, sparse occupancy labels and dense occupancy labels. We observe that our dense voxels can provide much more realistic occupancy labels with clear semantic boundaries. 
	
	We think it is not trivial to propagate the sparse semantic label to a dense one since it is a ill-posed problem. The proposed multi-frame point cloud stitching can aggregate multi-frame semantic information and provide dense enough reference points for NN. However, we find that NN is sensitive to the annotation noise in
	original LiDAR semantic labels, and we will try to solve it
	as the future work.

	

	\begin{table*}
		\footnotesize
		\setlength{\tabcolsep}{0.004\linewidth}
		
		\newcommand{\classfreq}[1]{{~\tiny(\semkitfreq{#1}\%)}}  %
		\centering
		\begin{tabular}{l|c c | c c c c c c c c c c c c c c c c c c c}
			\toprule
			Method
			& \makecell{SC\\ IoU} & \makecell{SSC \\ mIoU}
			& \rotatebox{90}{road}
			\rotatebox{90}{\ \ \ \classfreq{road}} 
			& \rotatebox{90}{sidewalk}
			\rotatebox{90}{\ \ \ \classfreq{sidewalk}}
			& \rotatebox{90}{parking}
			\rotatebox{90}{\ \ \ \classfreq{parking}} 
			& \rotatebox{90}{other-grnd}
			\rotatebox{90}{\ \ \ \classfreq{otherground}} 
			& \rotatebox{90}{ building}
			\rotatebox{90}{\ \ \ \classfreq{building}} 
			& \rotatebox{90}{ car}
			\rotatebox{90}{\ \ \ \classfreq{car}} 
			& \rotatebox{90}{ truck}
			\rotatebox{90}{\ \ \ \classfreq{truck}} 
			& \rotatebox{90}{ bicycle}
			\rotatebox{90}{\ \ \ \classfreq{bicycle}} 
			& \rotatebox{90}{motorcycle}
			\rotatebox{90}{\ \ \ \classfreq{motorcycle}} 
			& \rotatebox{90}{ other-veh.}
			\rotatebox{90}{\ \ \  \classfreq{othervehicle}} 
			& \rotatebox{90}{vegetation}
			\rotatebox{90}{\ \ \ \classfreq{vegetation}} 
			& \rotatebox{90}{ trunk}
			\rotatebox{90}{\ \ \ \classfreq{trunk}} 
			& \rotatebox{90}{terrain}
			\rotatebox{90}{\ \ \ \classfreq{terrain}} 
			& \rotatebox{90}{ person}
			\rotatebox{90}{\ \ \ \classfreq{person}} 
			& \rotatebox{90}{ bicyclist}
			\rotatebox{90}{\ \ \ \classfreq{bicyclist}} 
			& \rotatebox{90}{ motorcyclist.}
			\rotatebox{90}{\ \ \ \classfreq{motorcyclist}} 
			& \rotatebox{90}{ fence}
			\rotatebox{90}{\ \ \ \classfreq{fence}} 
			& \rotatebox{90}{ pole}
			\rotatebox{90}{\ \ \ \classfreq{pole}} 
			& \rotatebox{90}{traf.-sign}
			\rotatebox{90}{\ \ \ \classfreq{trafficsign}} 
			\\
			\midrule
			LMSCNet~\cite{roldao2020lmscnet}  &  31.38 & 7.07 &  46.70 & 19.50 & 13.50 & 3.10 & 10.30 & 14.30 & 0.30 & 0.00 & 0.00 & 0.00 & 10.80 & 0.00 & 10.40 & 0.00 & 0.00 & 0.00 & 5.40 & 0.00 & 0.00   \\
			
			3DSketch~\cite{chen20203d}  & 26.85 & 6.23 & 37.70 & 19.80 & 0.00 & 0.00 & 12.10 & 17.10 & 0.00 & 0.00 & 0.00 & 0.00 & 12.10 & 0.00 & 16.10 & 0.00 & 0.00 & 0.00 & 3.40 & 0.00 & 0.00  \\
			
			AICNet~\cite{li2020anisotropic}  & 23.93 & 7.09	& 39.30	& 18.30 & 19.80 & 1.60 & 9.60	& 15.30	& 0.70	& 0.00	& 0.00	& 0.00	& 9.60	& 1.90	& 13.50	& 0.00	& 0.00	& 0.00	& 5.00	& 0.10	& 0.00 \\
			
			JS3C-Net~\cite{yan2021sparse}  & 34.00 & 8.97 & 47.30 & 21.70 & 19.90 & 2.80 & 12.70 & 20.10 & 0.80 & 0.00 & 0.00 & 4.10 & 14.20& 3.10 & 12.40 & 0.00 & 0.20 & 0.20 & 8.70 & 1.90 & 0.30  \\
			
			MonoScene~\cite{monoscene}  & 34.16 & 11.08 & 54.70 & 27.10 & 24.80 & 5.70 & 14.40 & 18.80 & 3.30 & 0.50 & 0.70 & \textbf{4.40} & \textbf{14.90} & 2.40 & 19.50 & 1.00 & 1.40 & \textbf{0.40} & 11.10 & 3.30 & 2.10  \\
			
			TPVFormer \cite{huang2023tri}  & 34.25 & 11.26 & 55.10 & 27.20 & 27.40 & 6.50 & 14.80 & 19.20 & \textbf{3.70} & 1.00 & 0.50 & 2.30 & 13.90 & 2.60 & \textbf{20.40} & 1.10 & \textbf{2.40} & 0.30 & 11.00 & 2.90 & 1.50  \\
			\midrule
			
			SurroundOcc & \textbf{34.72} & \textbf{11.86} & \textbf{56.90} & \textbf{28.30} & \textbf{30.20} & \textbf{6.80} & \textbf{15.20} & \textbf{20.60} & 1.40 & \textbf{1.60} & \textbf{1.20} & \textbf{4.40} & \textbf{14.90} & \textbf{3.40} & 19.30 & \textbf{1.40} & 2.00 & 0.10 & \textbf{11.30}& \textbf{3.90} & \textbf{2.40} \\ 
			\bottomrule
		\end{tabular}
		
		\vspace{1mm}
		\caption{\textbf{Monocular Semantic scene completion results on SemanticKITTI test set.} For fair comparison, we use the performances of RGB-inferred versions of the first four methods, which are reported in MonoScene~\cite{monoscene}. Although our method is not designed for monocular perception, we still outperform other methods for a large margin.}
		\label{tab:kittiseg}
		
	\end{table*}
	
	\section{Experiments}
	\subsection{Experimental Setup}
	\noindent \textbf{Dataset:}
	We conduct multi-camera experiments on nuScenes dataset \cite{caesar2020nuscenes}, which is a large-scale autonomous driving dataset. Since the 3D semantic and 3D detection labels are unavailable in test set and we cannot generate dense occupancy labels, we use the training set to train the model and validation set for evaluation. The occupancy prediction range is set as $[-50m, 50m]$ for $X,Y$ axis and $[-5m, 3m]$ for $Z$ axis. The final output occupancy has the shape 200x200x16 with 0.5m voxel size. The input image resolution is 1600x900. 
	
	To further demonstrate the effectiveness of our method, we conduct monocular semantic scene completion experiment on SemanticKITTI dataset \cite{behley2019semantickitti}. SemanticKITTI has annotated outdoor LiDAR scans with 21 semantic labels. The ground truth is voxelized as 256x256x32 grid with 0.2m voxel size. We evaluate our model on the test set. 
	
	
	\noindent \textbf{Implementation Details:}
	The whole network architecture obtains $M=4$ levels and we do not add the skip connection in level 0. For nuScenes dataset, we adopt ResNet101-DCN \cite{he2016deep, dai2017deformable} with the initial weight from FCOS3D \cite{fcos3d} as the backbone to extract image features. The features of stage 1,2,3 are fed to FPN \cite{lin2017feature} and used as multi-scale image features. The number of 2D-3D spatial attention layers are set as 1, 3, 6 for three levels. For SemanticKITTI dataset, following MonoScene \cite{monoscene}, we use a pretrained EfficientNetB7 \cite{tan2019efficientnet} as the backbone to generate multi-scale image features. We also adopt FPN to further fuse the features of different levels. We set the number of 2D-3D spatial attention layers as 1, 3, 8. All experiments are conducted on 8 RTX 3090s. For Possion Reconstruction, we accumulate both key-frame and non
	key-frame data. In details, we use LiDAR frames at 20Hz, resulting in 400 frames and around 13 million points for a 20 second sequence.

	\begin{figure*}[t]
		\centering
		\includegraphics[width=0.98\linewidth]{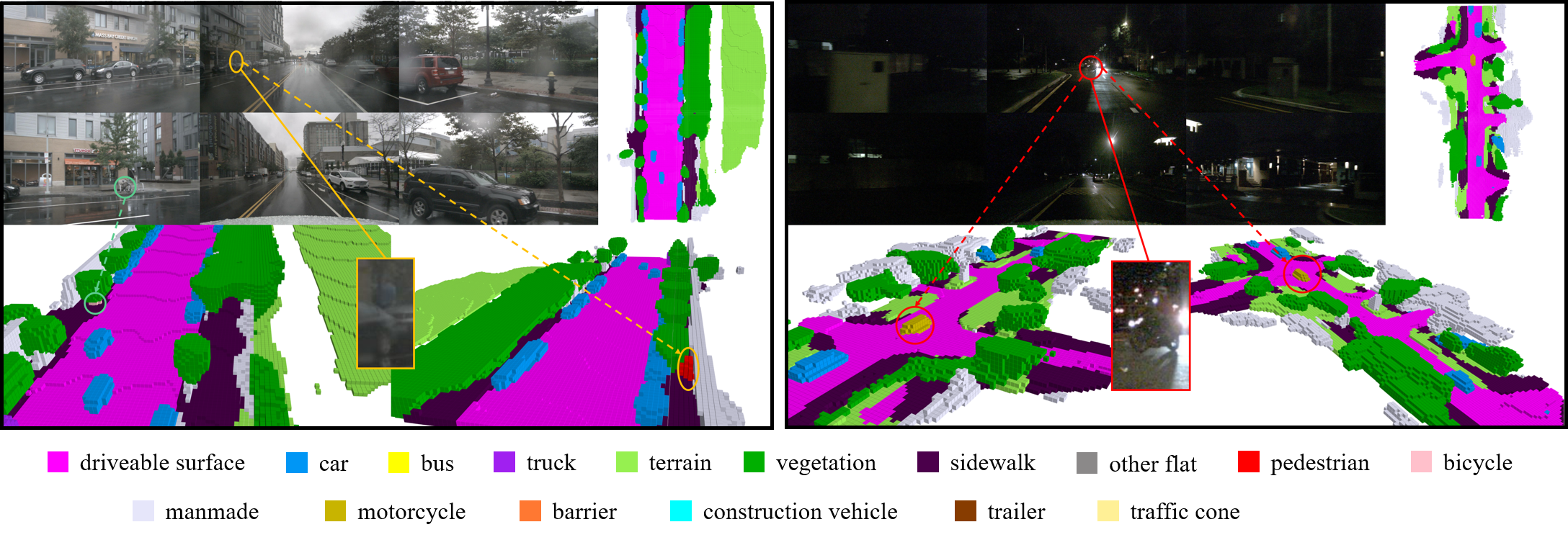}
		\caption{An example of challenging scenes. Although the quality of RGB images degrades in rainy days and nights, our method can still predict detailed occupancy.  \textbf{Better viewed when zoomed in.}
		}
		\vspace{-3mm}
		\label{fig:seg1}
	\end{figure*}

	\begin{table}[t]
		\centering
		\setlength\tabcolsep{2pt}
		\resizebox{0.49\textwidth}{!}{
			\begin{tabular}{l|cccccc}
				\toprule
				Method  & Acc $\downarrow$  & Comp $\downarrow$  & Prec $\uparrow$& Recall $\uparrow$ & \textbf{CD} $\downarrow$  & \textbf{F-score} $\uparrow$  \\
				\midrule
				SurroundDepth \cite{wei2022surrounddepth}& 1.747& 1.384& 0.261& 0.353 & 3.130 & 0.293\\
				AdaBins \cite{bhat2021adabins}& 1.989 & 1.287 & 0.233& 0.347& 3.275 & 0.271 \\
				NeWCRFs \cite{yuan2022new}&2.163& 1.233&0.214& 0.348 & 3.396 & 0.257  \\
				Atlas \cite{atlas} &\textbf{0.679}& 1.685&0.407& 0.546 & 2.365 & 0.458  \\
				TransformerFusion \cite{transformerfusion} &0.771& 1.434&0.375& 0.591 & 2.205 & 0.453  \\
				\midrule
				SurroundOcc &0.724 & \textbf{1.226} &\textbf{0.414}& \textbf{0.602} & \textbf{1.950} & \textbf{0.483}  \\
				\bottomrule
				\hline
			\end{tabular}
		}
		\vspace{1mm}
		\caption{\textbf{3D scene reconstruction results on nuScenes validation set}. F-score and CD are the main metrics.}
		\vspace{-3mm}
		\label{tab:occ}
	\end{table}

	\noindent \textbf{Evaluation Metrics:}
	For 3D semantic occupancy prediction, we use the intersection over union (IoU) of occupied voxels, ignoring their
	semantic class as the evaluation metric of the scene completion (SC) task and the mIoU of all semantic classes for the SSC task.
	\begin{equation}
		\begin{aligned}
			\text{IoU} = &\frac{TP}{TP + FP + FN} \\
			\text{mIoU} = \frac{1}{C} & \sum_{i=1}^{C} \frac{TP_{i}}{TP_{i} + FP_{i} + FN_{i}}
		\end{aligned}
	\end{equation}
	where $TP$, $FP$, $FN$ indicate the number of true positive, false positive, and false negative predictions. $C$ is the class number.
	
	For 3D scene reconstruction, we first convert occupancy prediction to point clouds. Following \cite{atlas,transformerfusion}, we use 3D metrics for evaluation, which is shown in Table \ref{tab:metric_defs}. Chamfer distance (CD) and F-score are the main metrics since they consider both precision and recall. Please refer to the supplementary for more evaluation metric details. For all evaluation, we adopt dense occupancy as ground truth since sparse LiDAR points cannot fully evaluate the quality of occupancy reconstruction.

	\subsection{3D Semantic Occupancy Prediction}
	We first conduct multi-camera 3D semantic occupancy prediction on nuScenes \cite{caesar2020nuscenes} dataset and compare with several state-of-the-art methods \cite{monoscene,bevformer,huang2023tri,atlas}. For MonoScene \cite{monoscene}, we concatenate the
	multi-camera occupancy predictions and voxelize them with the 0.5m voxel size, which is same as our setting. For BEVFormer \cite{bevformer}, we add a 3D segmentation head to predict semantic occupancy.  As shown in Table \ref{tab:nuscseg}, our method achieves state-of-the-art performance. We also show some qualitative results in Figure \ref{fig:seg1} and Figure \ref{fig:seg2}.  See supplementary material for more video demos and qualitative comparisons. 
	
	Especially in Figure  \ref{fig:seg1}, we show rainy day and night visualization. Although the quality of RGB images degrades in these two challenging scenes, our method can still predict fine details. The LiDAR sensor suffers from rainy days and easily
	misses points. With multi-frame aggregation and Possion
	Reconstruction, we dramatically densify the labels and provide strong supervision, which is crucial for the challenging
	scenarios. Moreover, we conduct color jitter augmentation,
	which increases the robustness of brightness change. We also note that some parts cannot be observed by LiDAR sensor, such as the back side of the car. Surprisingly, our model can predict the complete shape according to the surrounding information. 
	
	To further demonstrate the superiority of our method, we also conduct monocular 3D semantic scene completion on SemanticKITTI dataset \cite{behley2019semantickitti}. Table \ref{tab:kittiseg} shows the results. Although our method is not designed for monocular perception and cross-view attention will be ineffective for the monocular setting, our method still achieves state-of-the-art performance on this benchmark.

	\subsection{3D Scene Reconstruction}
	Another important application of 3D occupancy prediction is 3D scene reconstruction. Due to this reason, we further evaluate 3D reconstruction performance without using multi-class semantic labels. We do the comparison with state-of-the-art multi-camera depth estimation methods (SurroundDepth \cite{wei2022surrounddepth}), monocular depth estimation methods (AdaBins \cite{bhat2021adabins} and NeWCRFs \cite{yuan2022new}) and 3D reconstruction method (Atlas \cite{atlas} and TransformerFusion \cite{transformerfusion}). For the self-supervised methods SurroundDepth \cite{wei2022surrounddepth}, we use depth ground truth to supervise them. To evaluate depth estimation methods in 3D space, following \cite{atlas,transformerfusion}, we run TSDF fusion \cite{curless1996volumetric, newcombe2011kinectfusion} to fuse multi-camera depth as point clouds. Note that
	we fuse the multi-camera depth maps of the same timestamp
	but not the multi-frame depths. Thus, there is no need to
	specially deal with movable objects. As shown in Table \ref{tab:occ}, SurroundOcc achieves state-of-the-art performance on most metrics, and outperforms other methods by a large margin, which verifies the effectiveness of the proposed method.

	\begin{table}[t]
		\centering
		\resizebox{0.37\textwidth}{!}{
			\begin{tabular}{c|cc}
				\toprule
				Method & SC IoU & SSC  mIoU  \\ \midrule
				w/o spatial attention & 29.78 & 17.34\\
				BEV-based attention &30.45 &18.94 \\
				Ours &\bf{31.49} &\bf{20.30} \\
				\bottomrule
			\end{tabular}
		}
		\vspace{2mm}
		\caption{The ablation study of 2D-3D spatial attention. ``w/o spatial attention" indicates that we average all multi-camera features in a grid.}
		\label{tab:ab_1}
	\end{table}

	\begin{table}[t]
		\centering
		\resizebox{0.43\textwidth}{!}{
			\begin{tabular}{c|cc}
				\toprule
				Method & SC IoU & SSC  mIoU \\ \midrule
				w/o multi-scale structure & 30.41 & 18.22\\
				w/o multi-scale supervision &31.16 &19.73 \\
				Ours &\bf{31.49} &\bf{20.30} \\
				\bottomrule
			\end{tabular}
		}
		\vspace{2mm}
		\caption{The ablation study of multi-scale occupancy prediction. ``w/o multi-scale structure" means that we do not add multi-scale skip connection.}
		\label{tab:ab_2}
	\end{table}

	\begin{table}[t]
		\centering
		\resizebox{0.4\textwidth}{!}{
			\begin{tabular}{c|cc}
				\toprule
				Supervision & SC IoU & SSC  mIoU \\ \midrule
				sparse LiDAR points & 11.96 & 12.17\\
				sparse occupancy labels &30.58 &18.83 \\
				dense occupancy labels  &\bf{31.49} &\bf{20.30} \\
				\bottomrule
			\end{tabular}
		}
		\vspace{2mm}
		\caption{The ablation study of dense occupancy supervision. The model trained with our dense occupancy ground truth is much better than that trained with sparse LiDAR points.}
		\label{tab:ab_3}
		\vspace{-3mm}
	\end{table}

	\begin{figure*}[tb]
		\centering
		\setlength\tabcolsep{1.0pt} 
		\renewcommand{\arraystretch}{1.0}
		\vspace{-3mm}
		\begin{tabular}{cccccc}
			
			\footnotesize FRONT LEFT & \footnotesize FRONT &  \footnotesize FRONT RIGHT & \footnotesize BACK RIGHT & \footnotesize BACK &\footnotesize BACK LEFT \\
			{\includegraphics[width=0.166\linewidth]{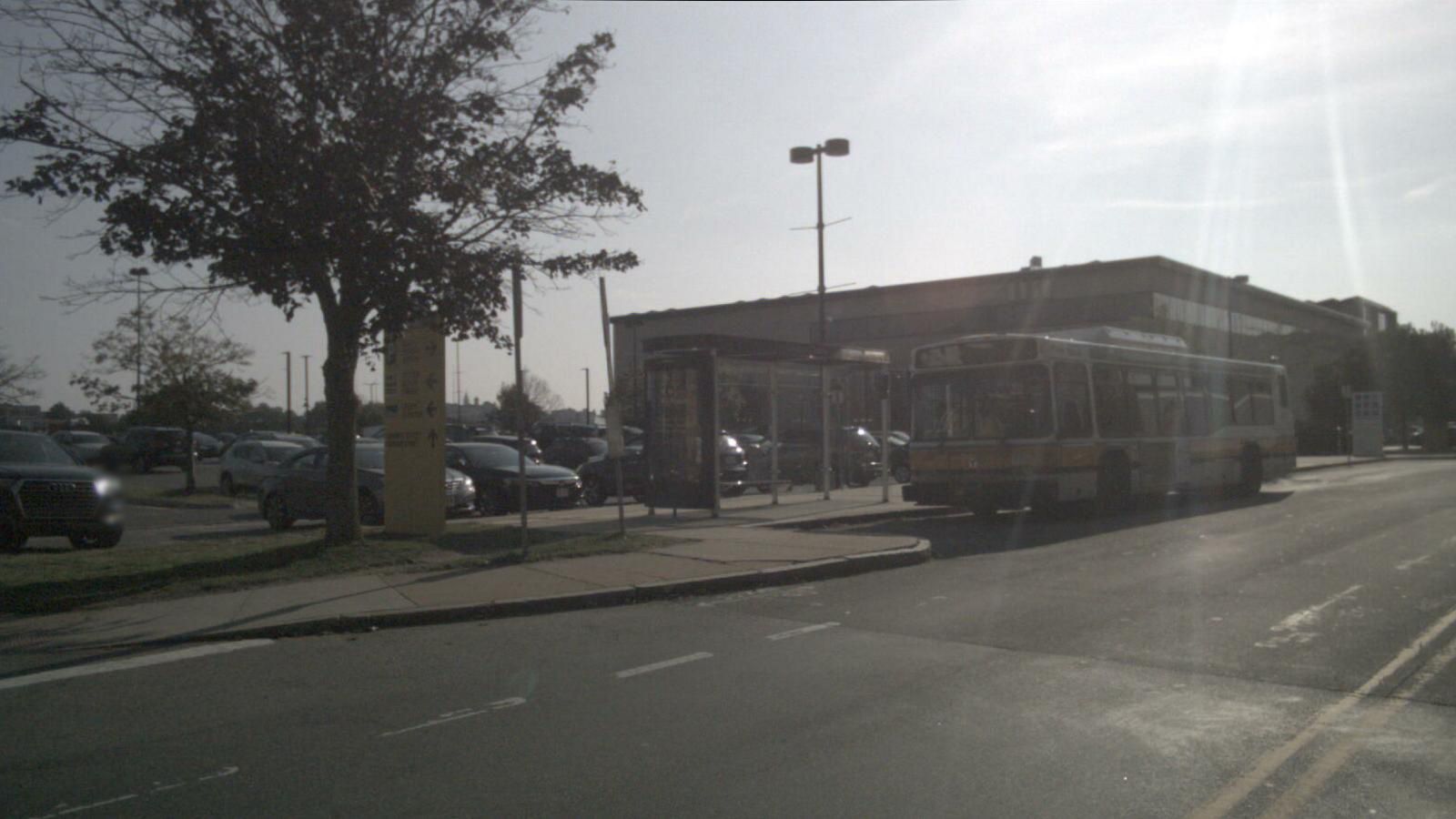}} &
			{\includegraphics[width=0.166\linewidth]{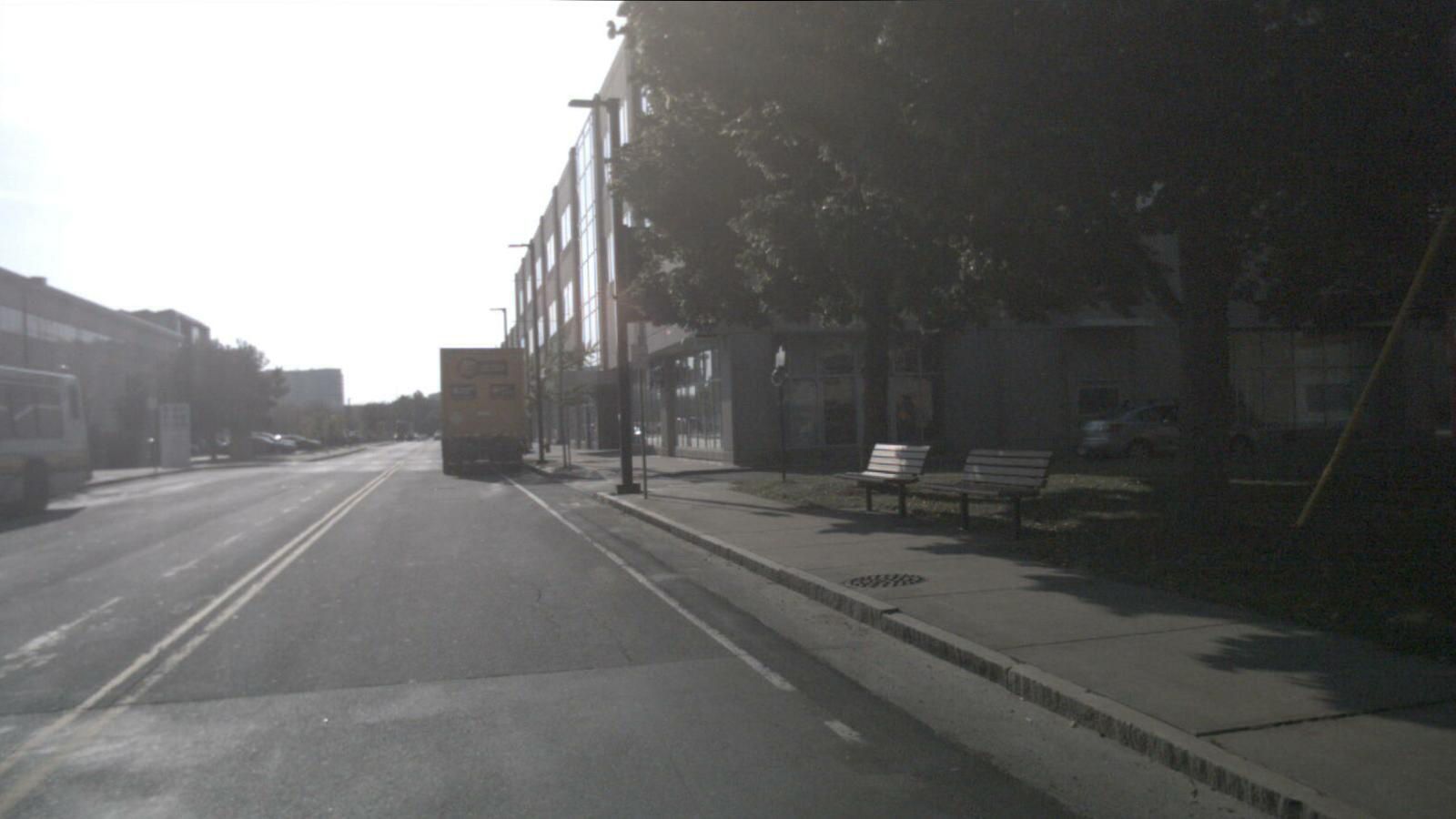}} &
			{\includegraphics[width=0.166\linewidth]{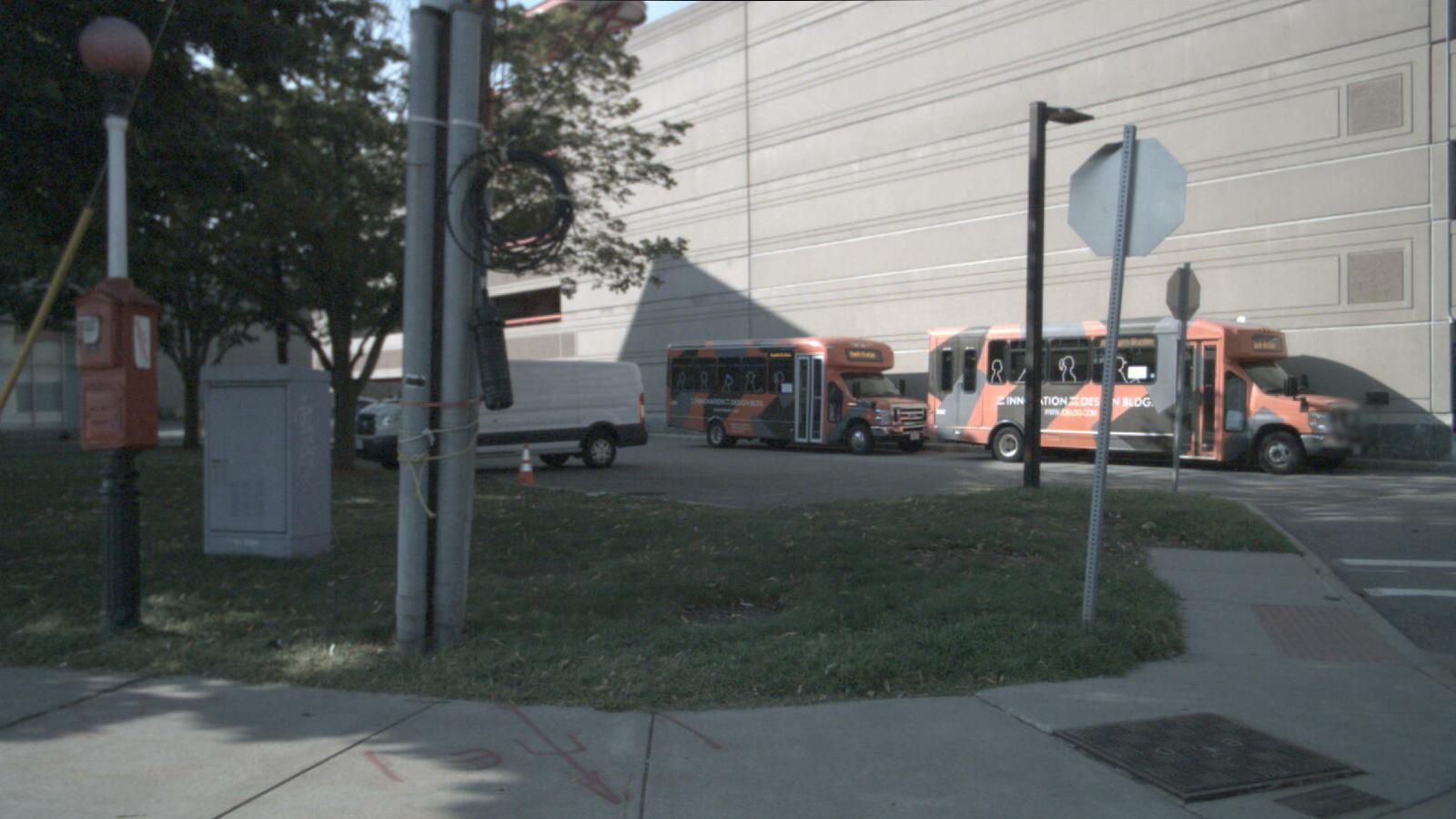}} &
			{\includegraphics[width=0.166\linewidth]{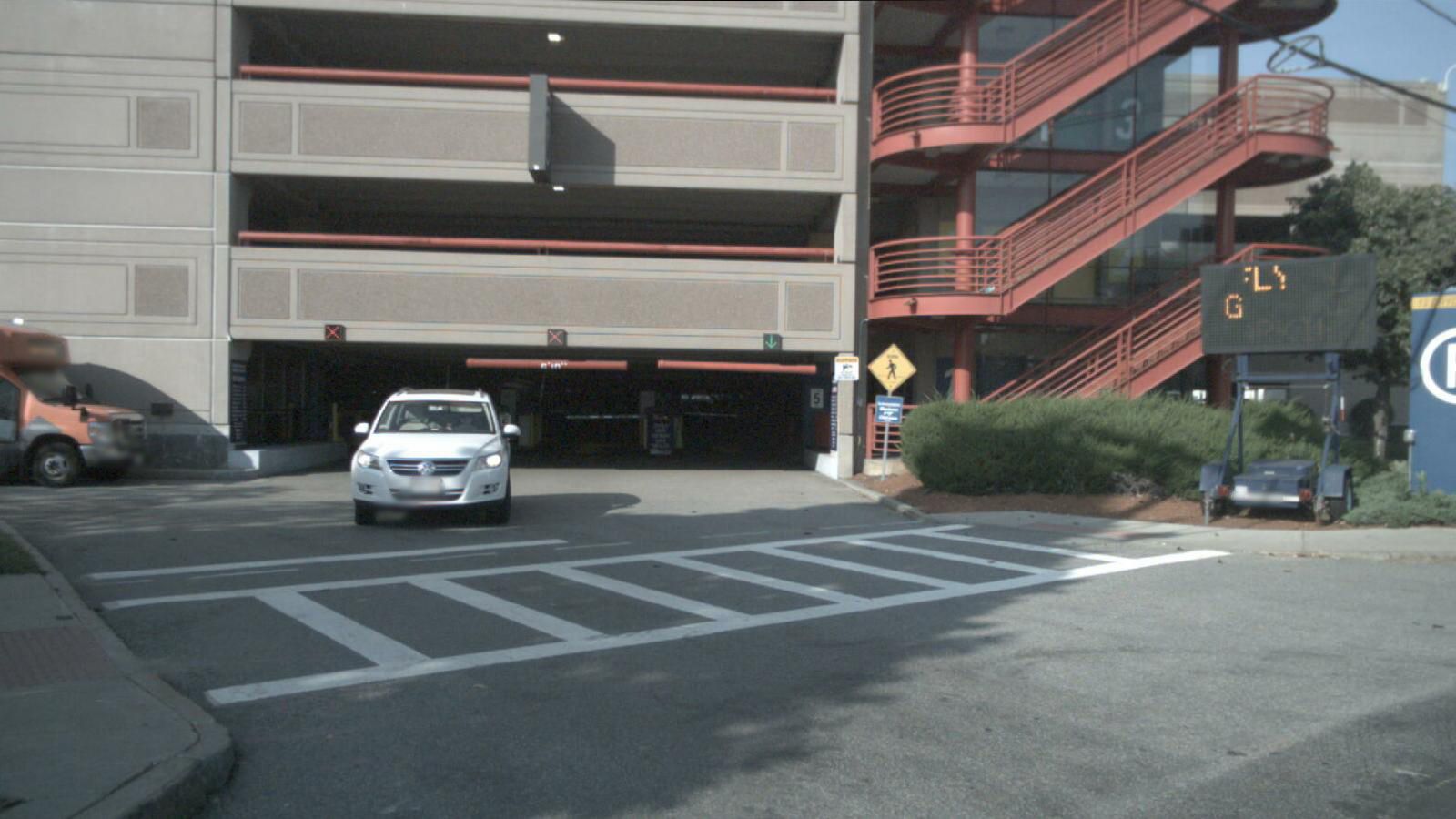}} &
			{\includegraphics[width=0.166\linewidth]{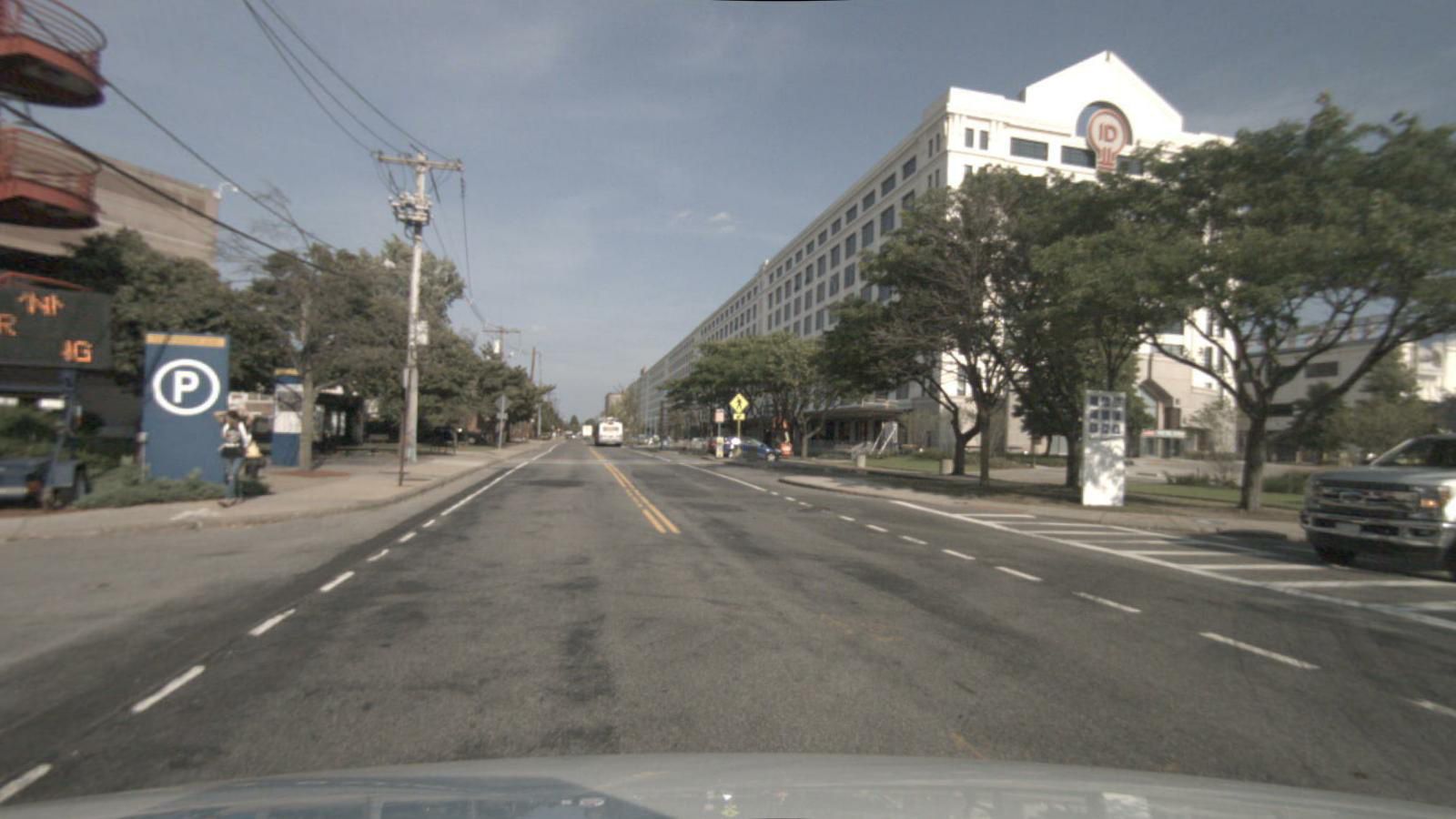}} &
			{\includegraphics[width=0.166\linewidth]{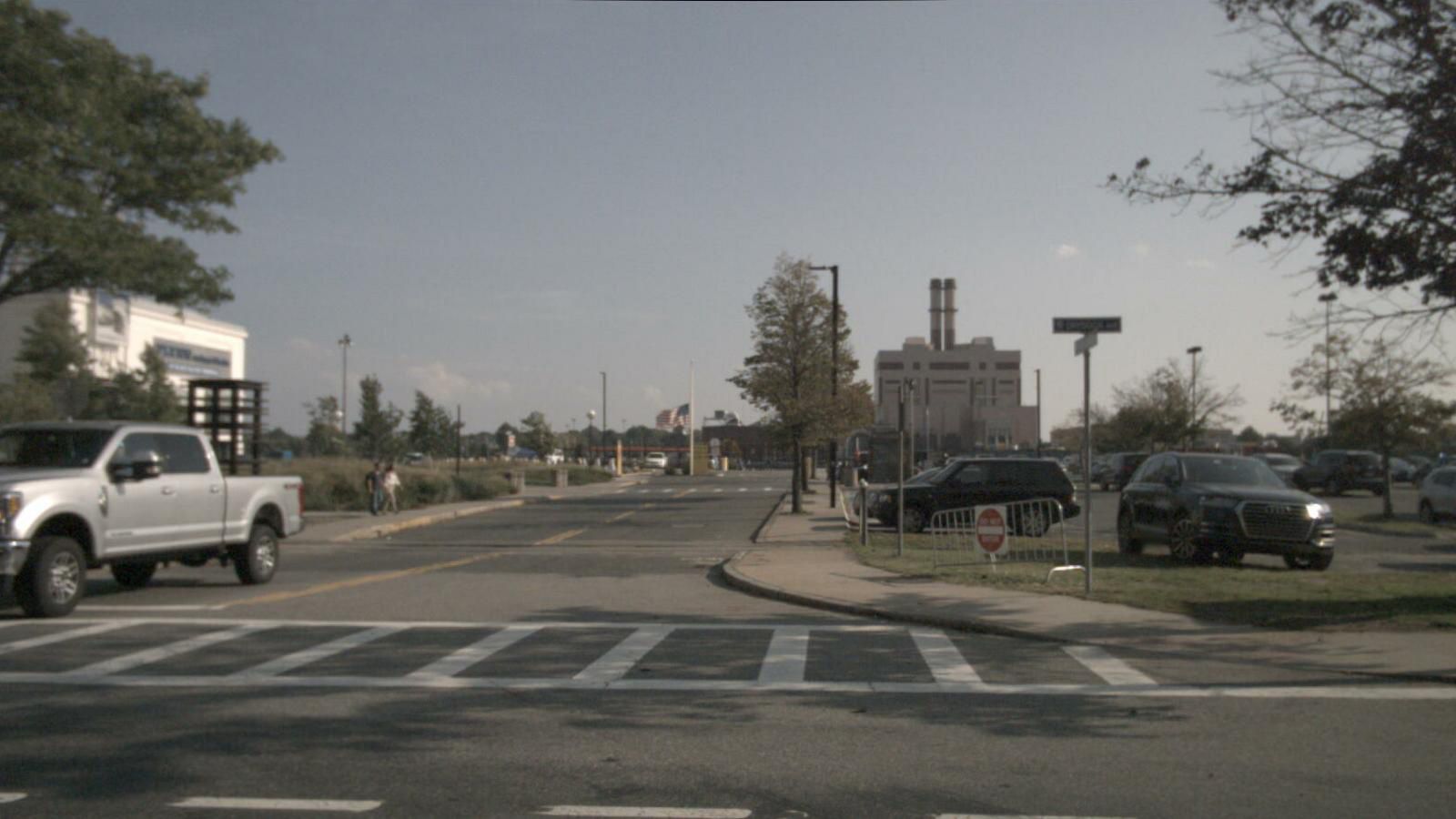}} \\
		\end{tabular}
		
		\begin{tabular}{cccc}
			{\includegraphics[width=0.25\linewidth]{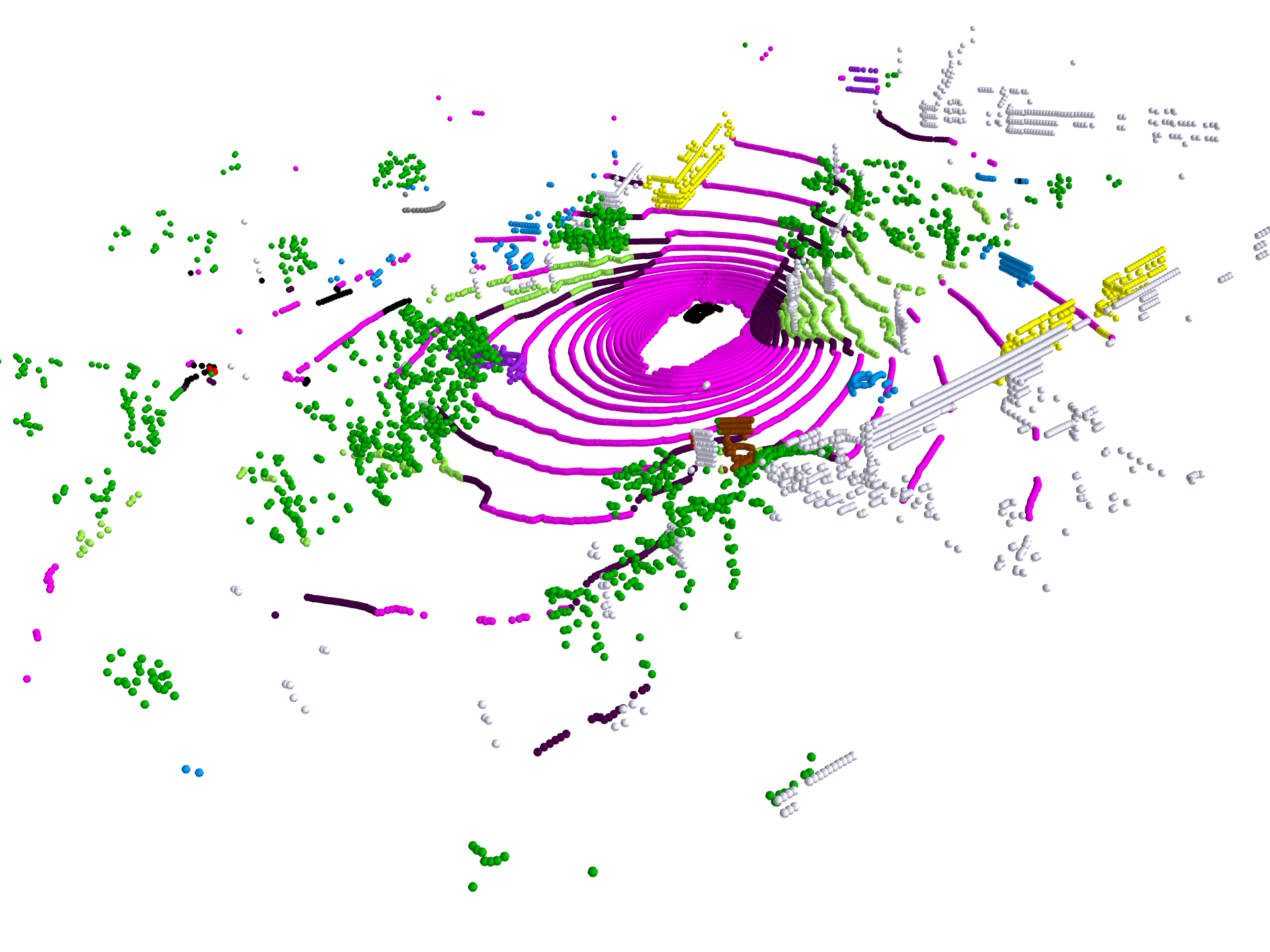}} &
			{\includegraphics[width=0.25\linewidth]{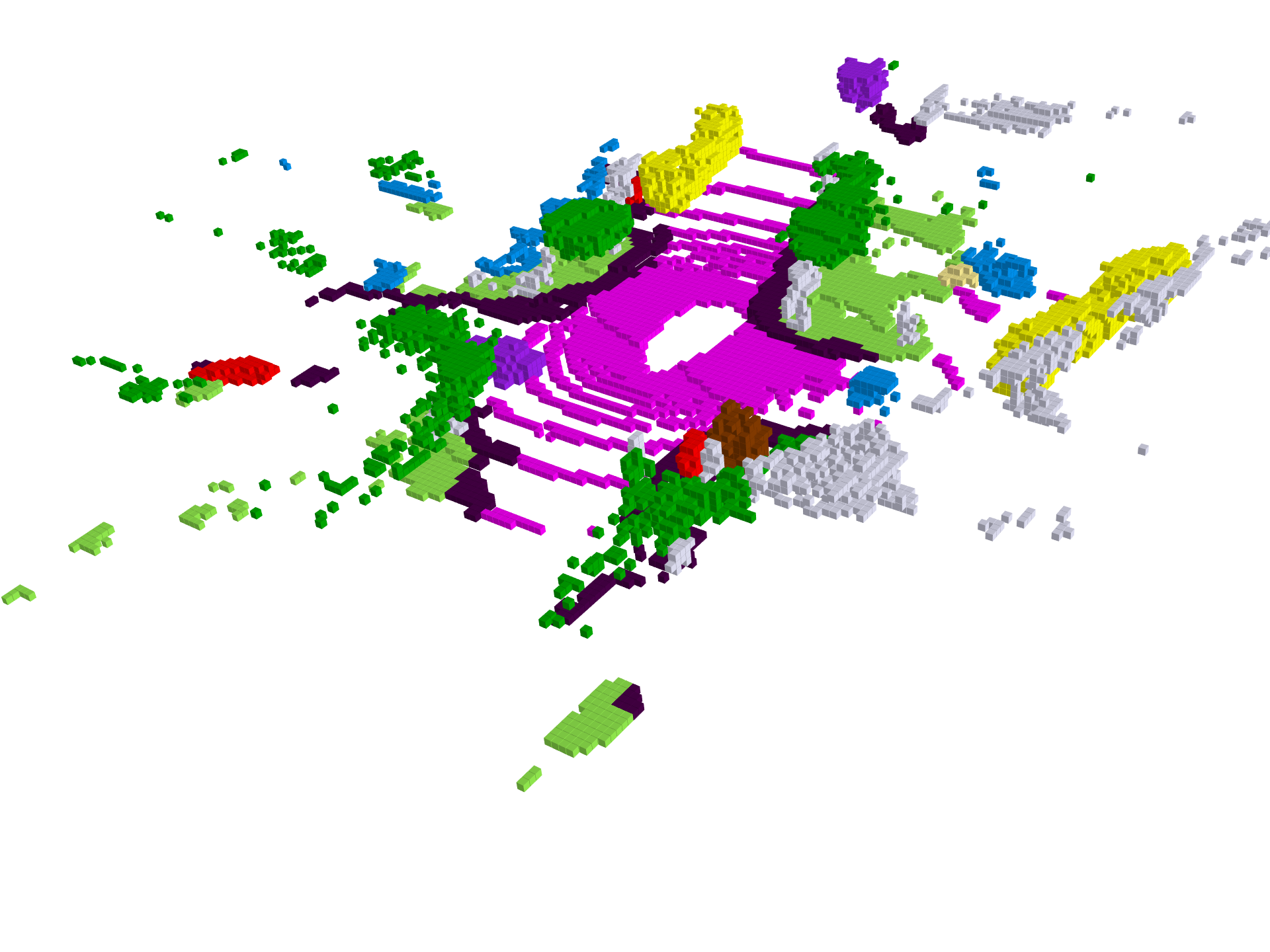}} &
			{\includegraphics[width=0.25\linewidth]{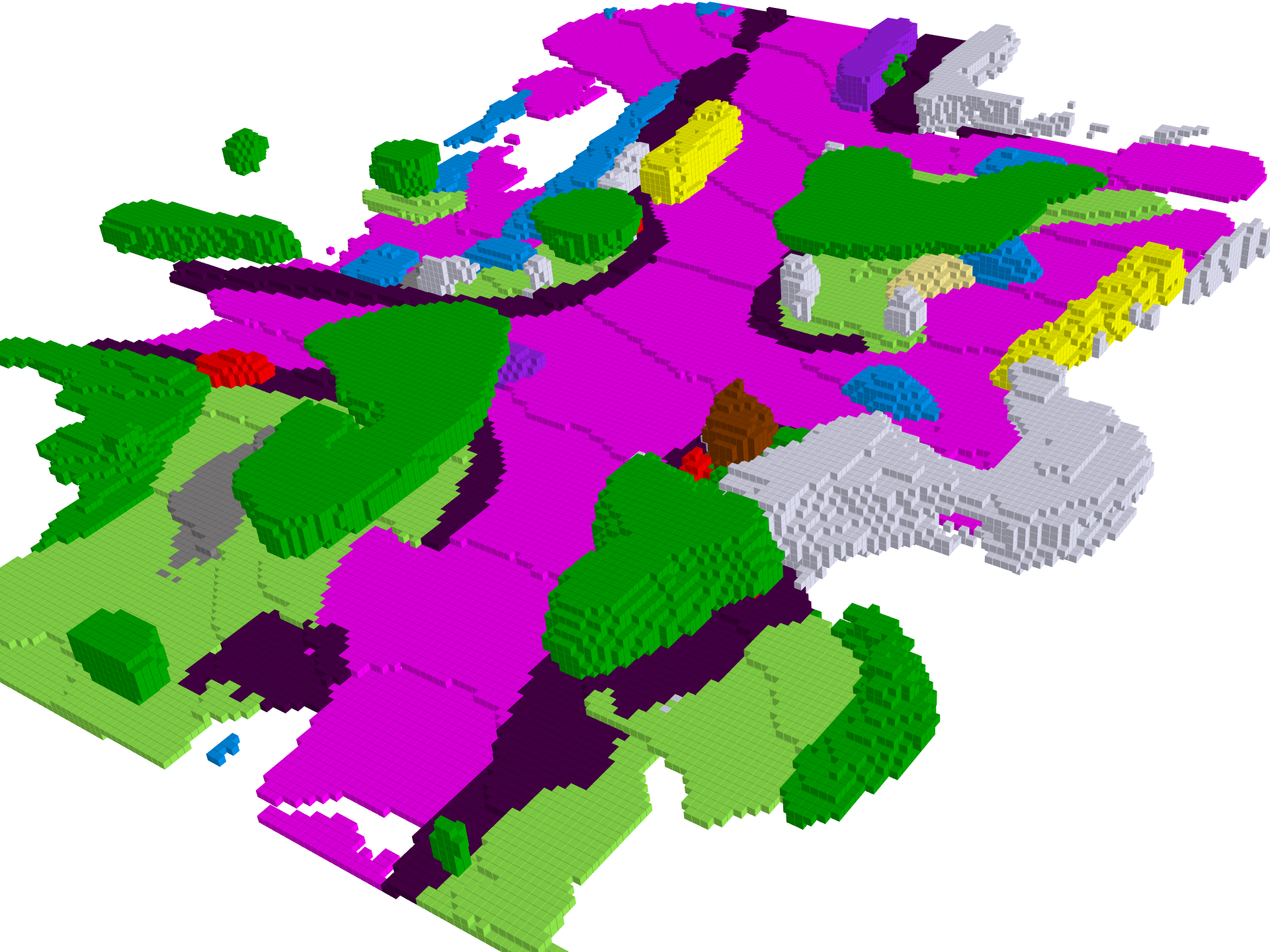}} &
			{\includegraphics[width=0.25\linewidth]{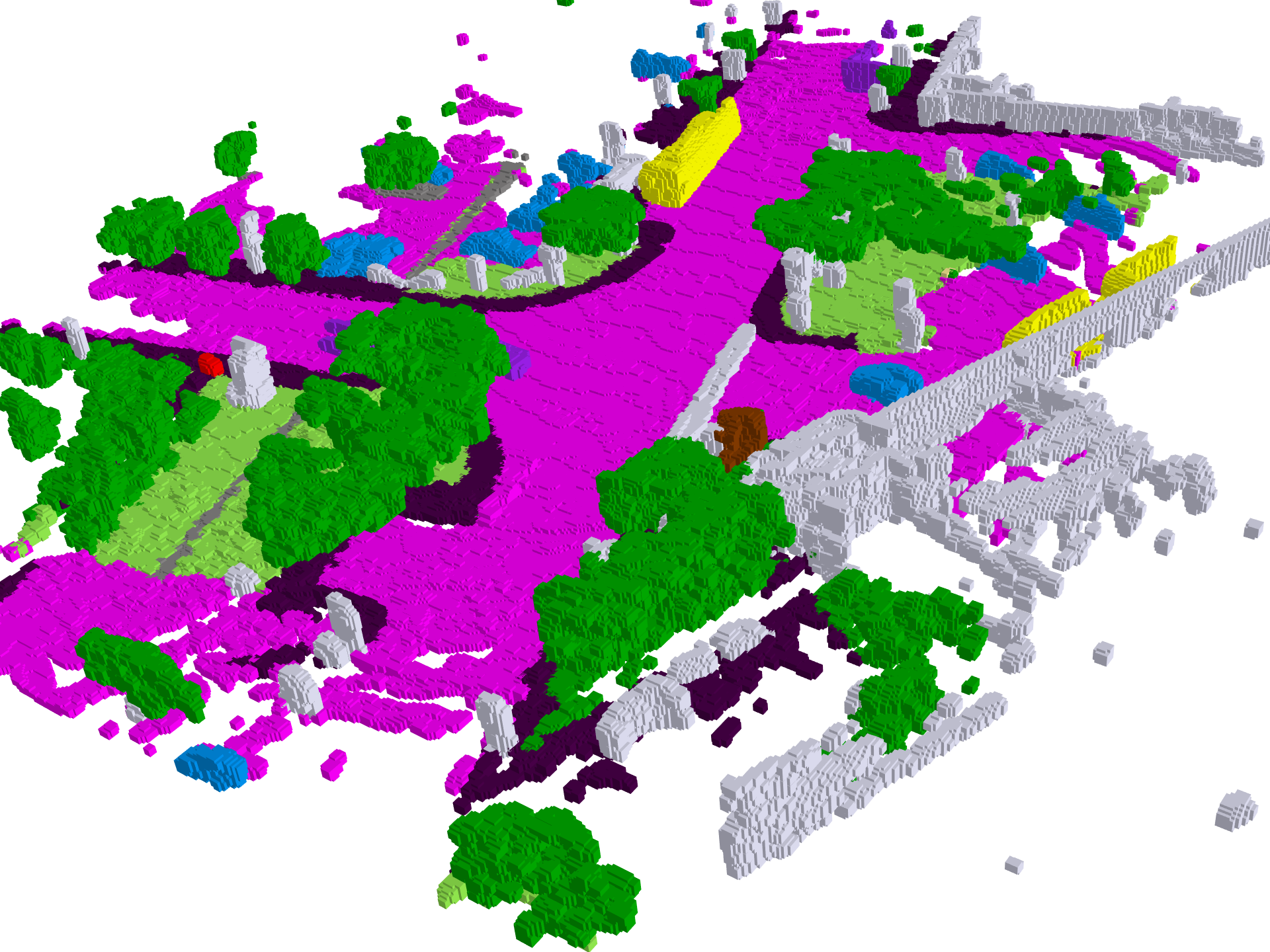}} \\
		\end{tabular}
		
		\begin{tabular}{cccccc}
			{\includegraphics[width=0.166\linewidth]{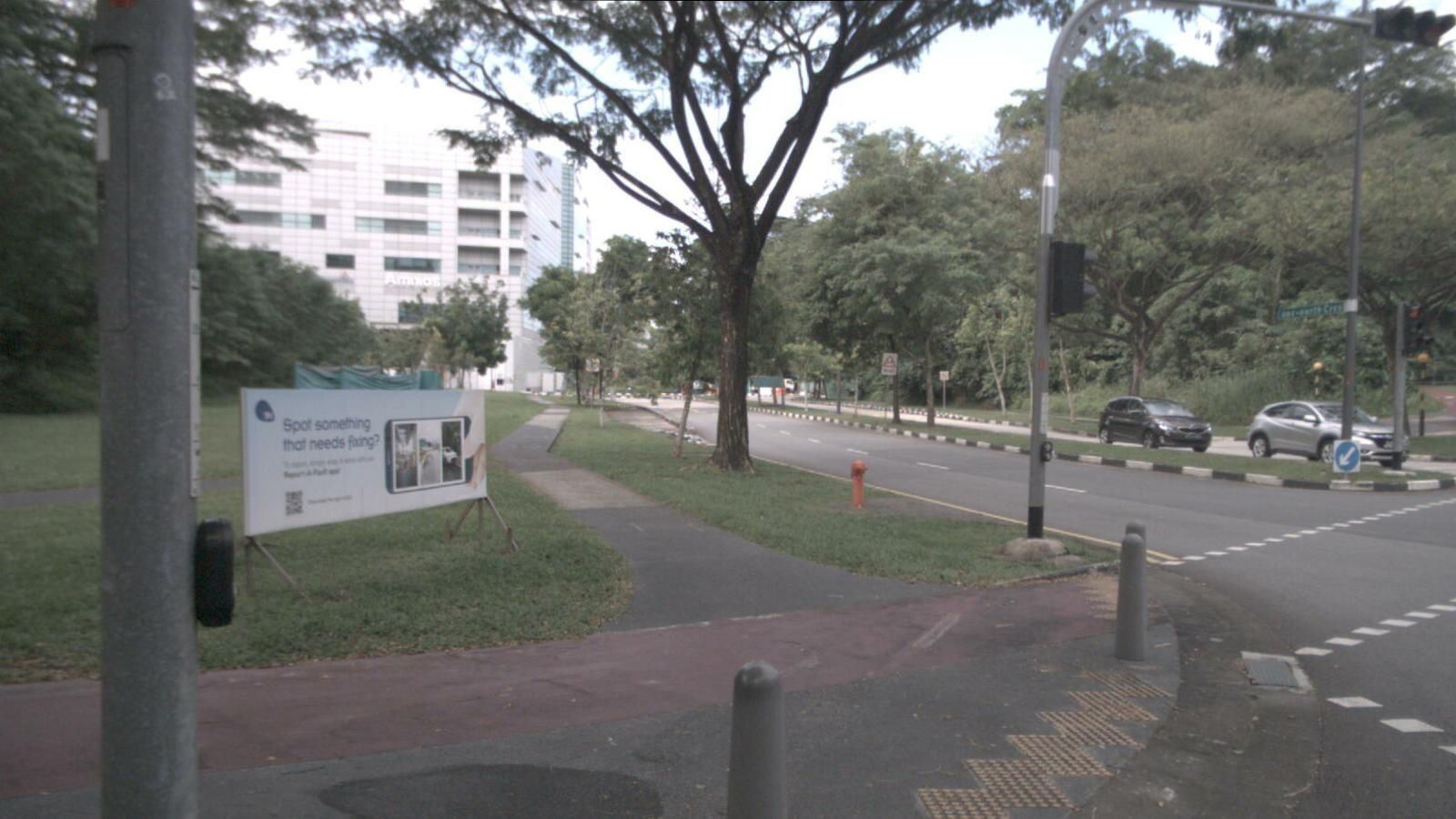}} &
			{\includegraphics[width=0.166\linewidth]{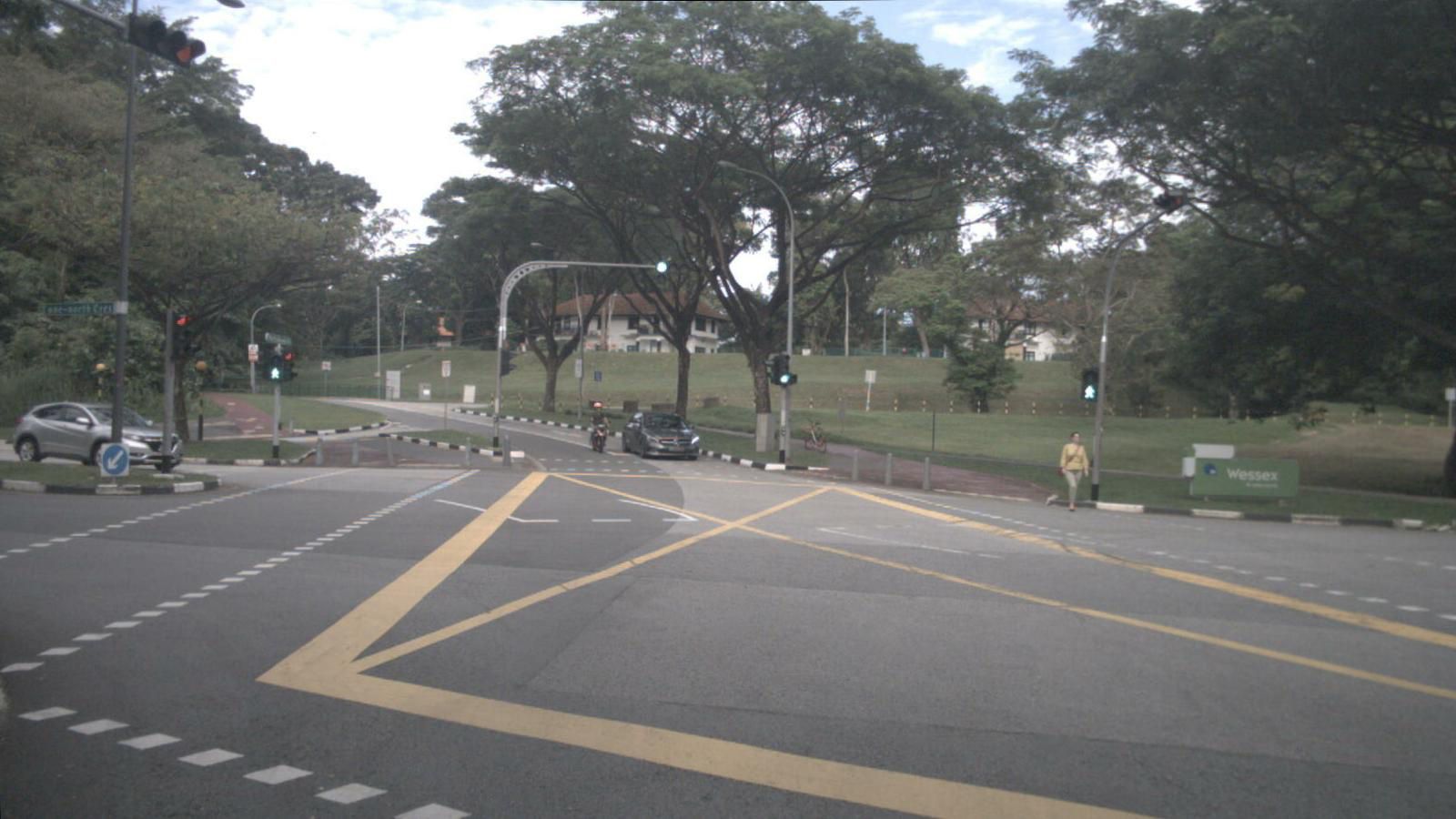}} &
			{\includegraphics[width=0.166\linewidth]{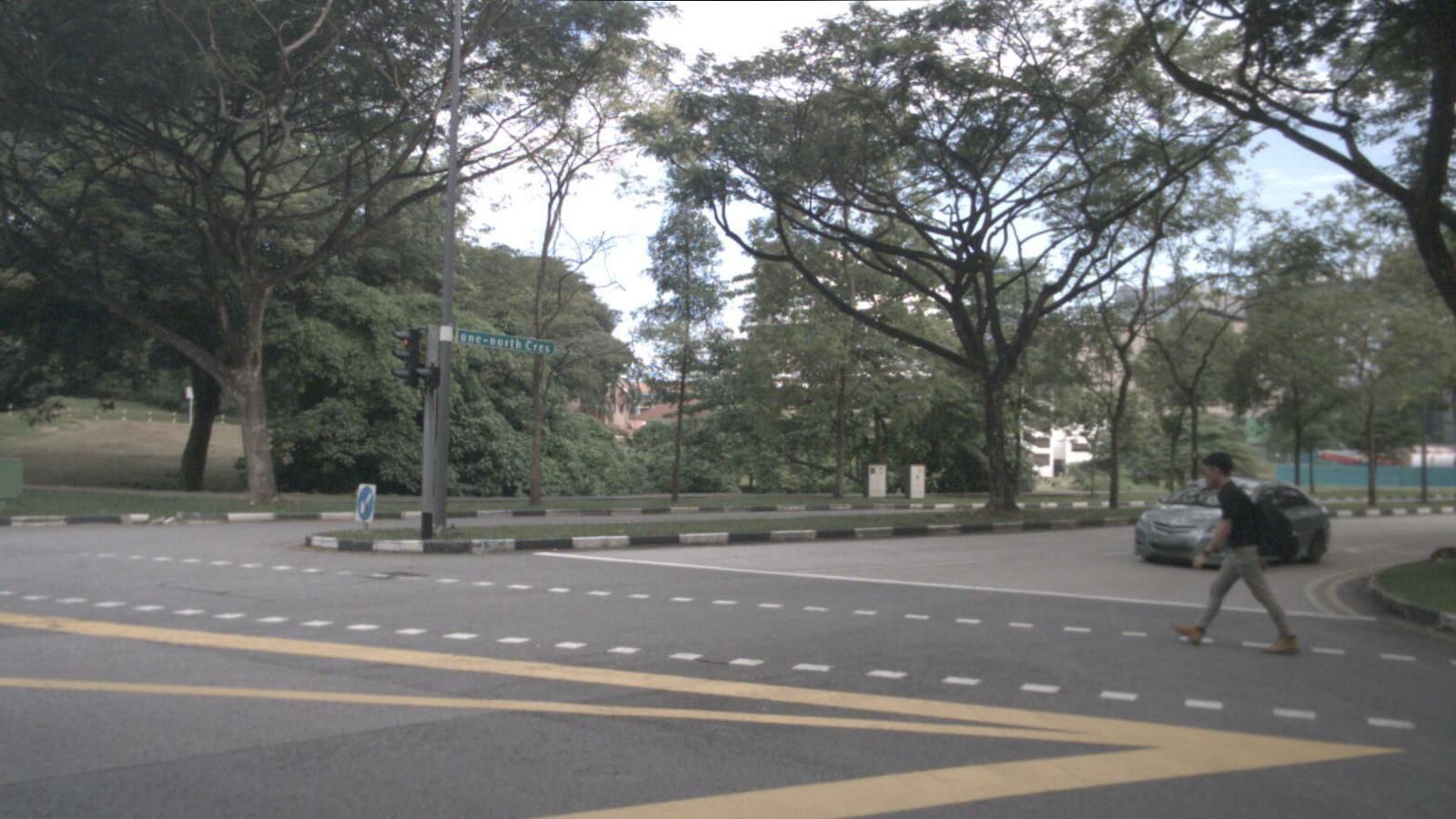}} &
			{\includegraphics[width=0.166\linewidth]{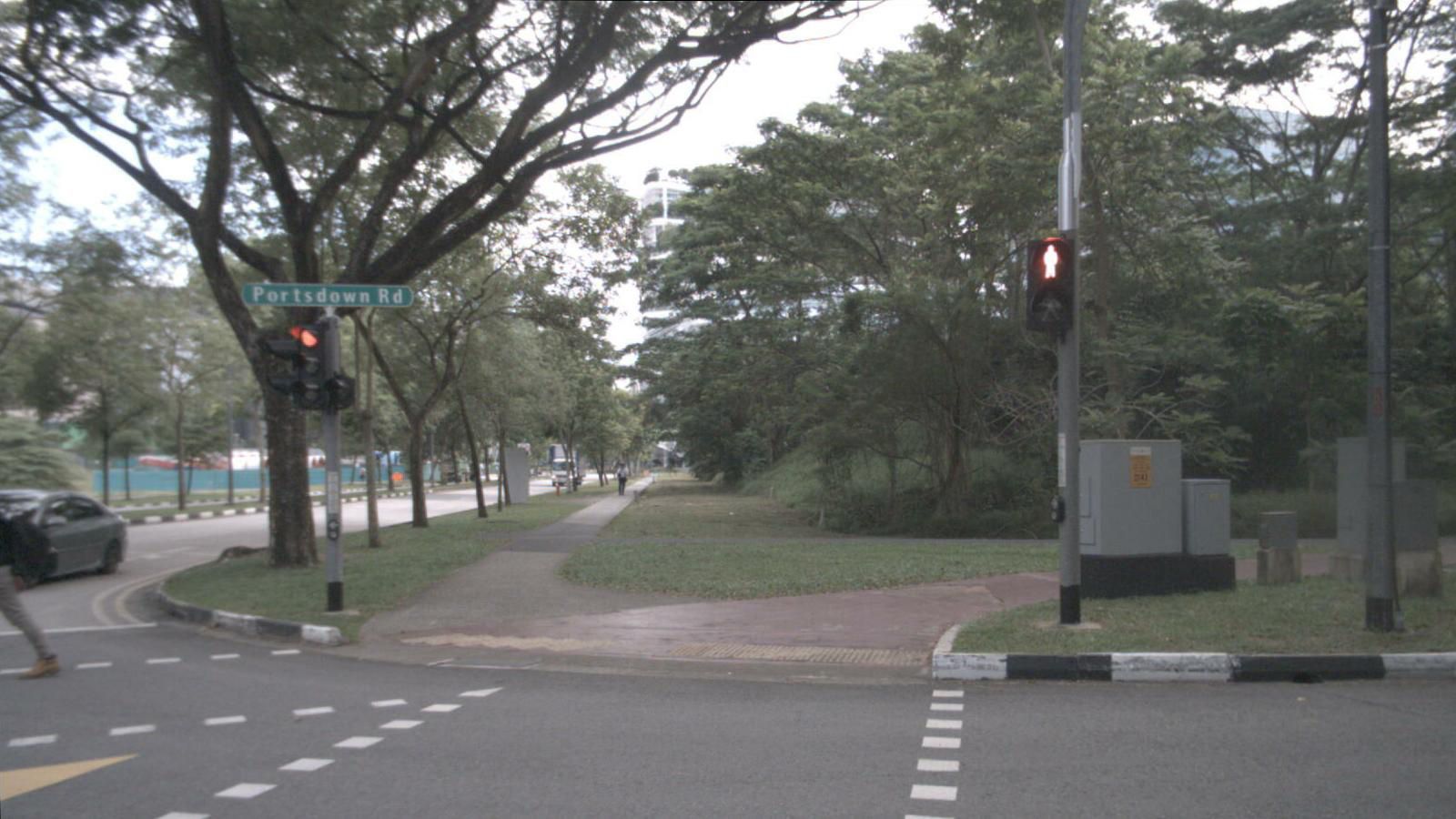}} &
			{\includegraphics[width=0.166\linewidth]{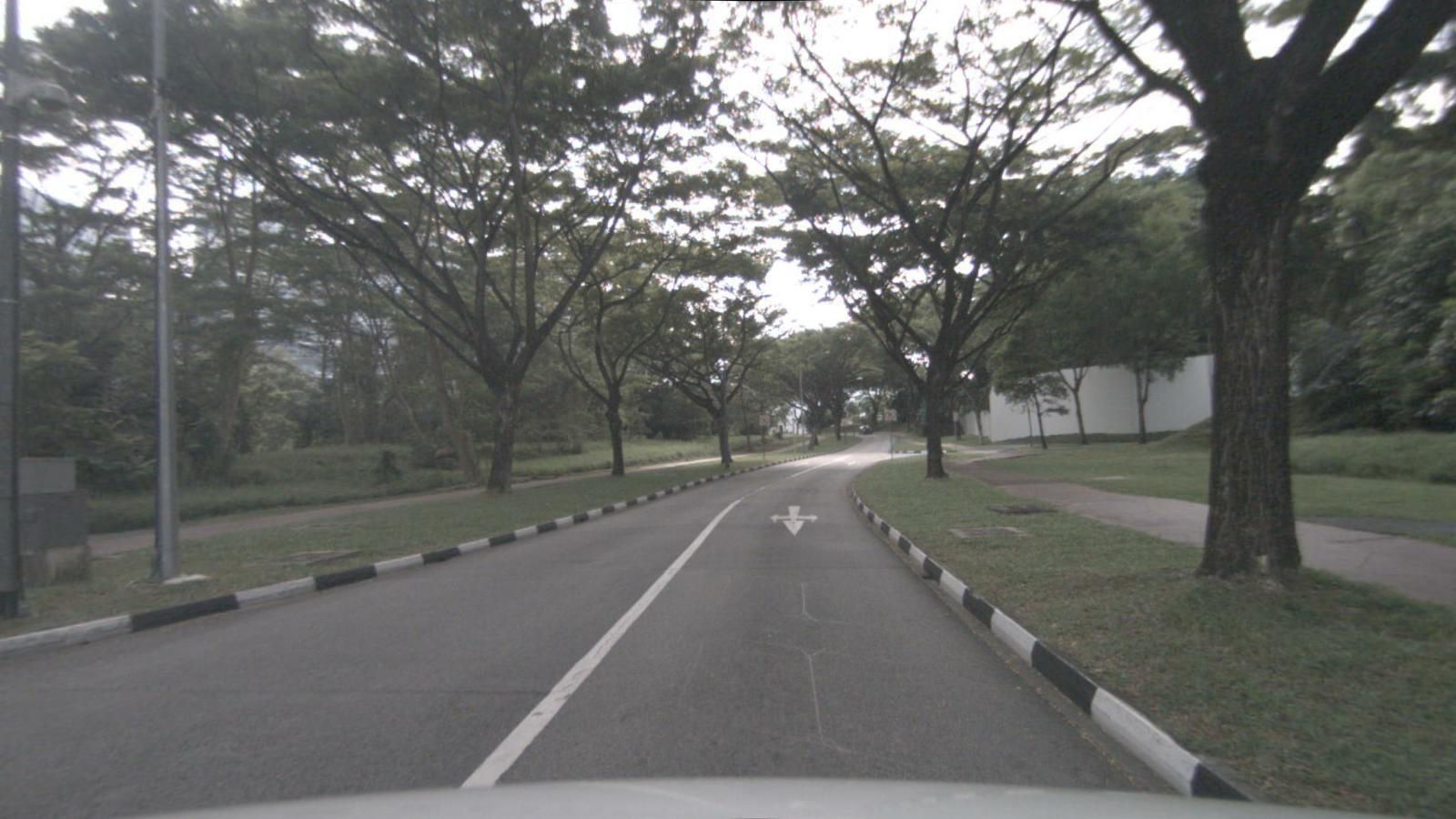}} &
			{\includegraphics[width=0.166\linewidth]{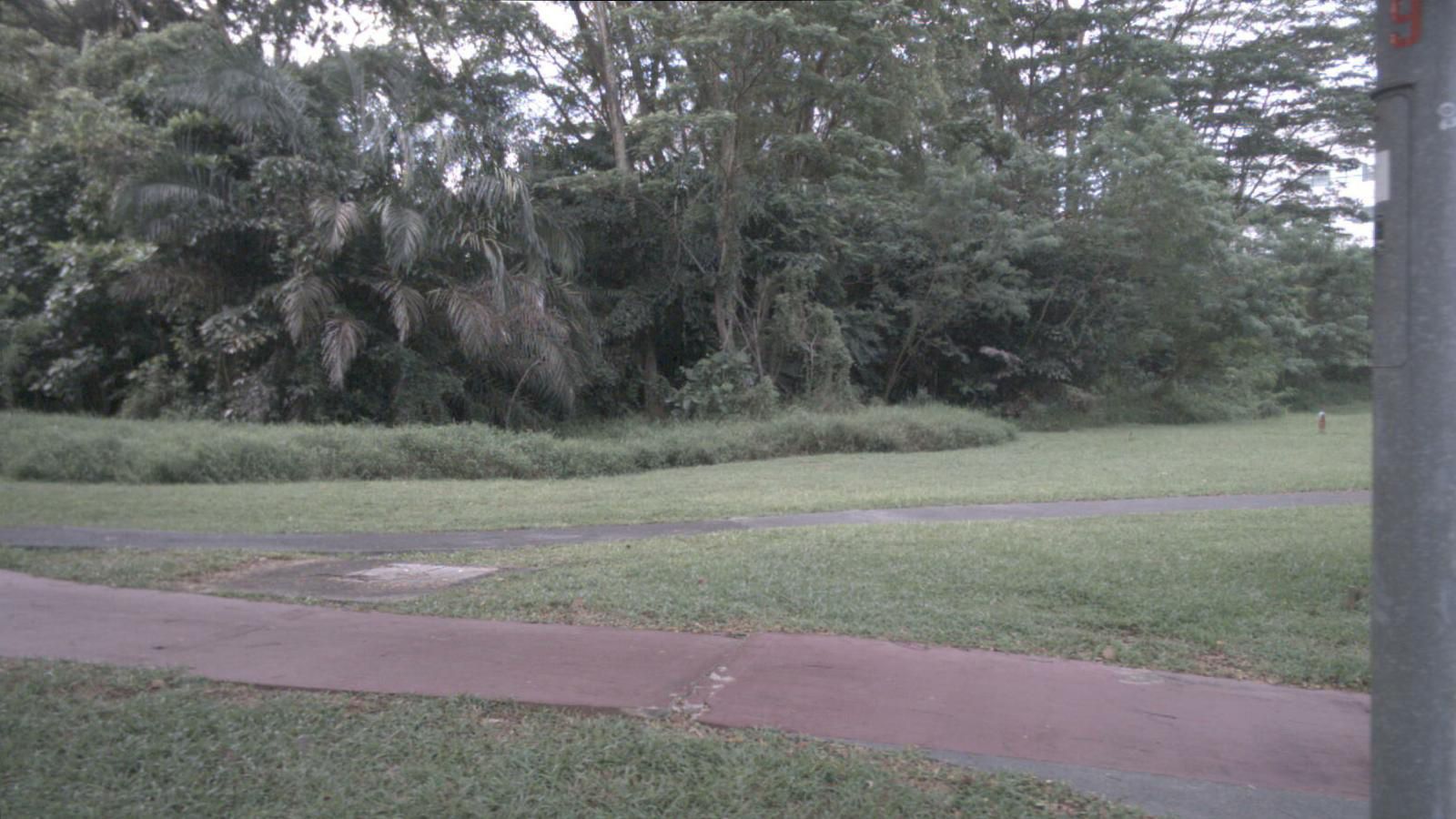}} \\
		\end{tabular}
		
		\begin{tabular}{cccc}
			{\includegraphics[width=0.25\linewidth]{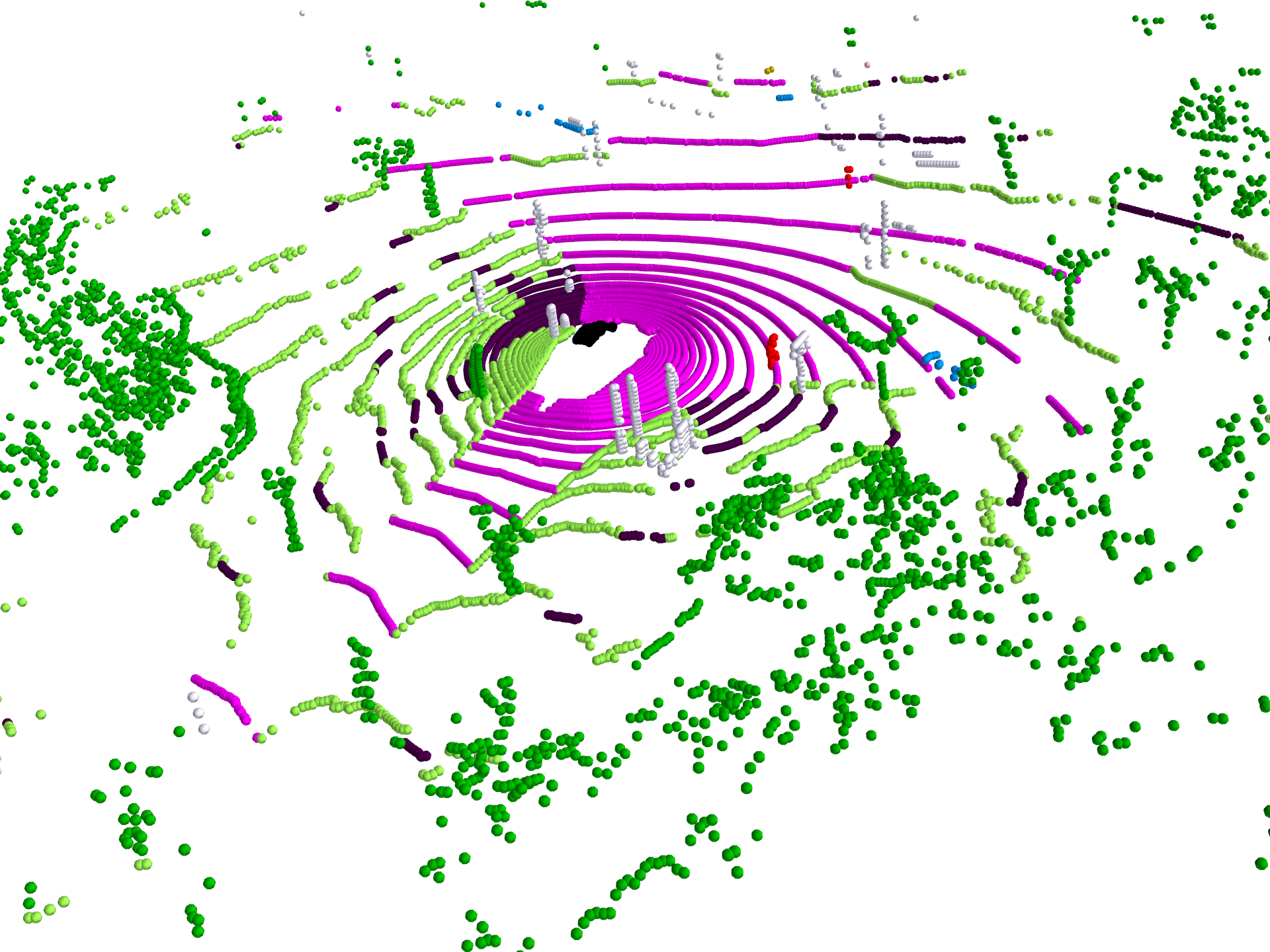}} &
			{\includegraphics[width=0.25\linewidth]{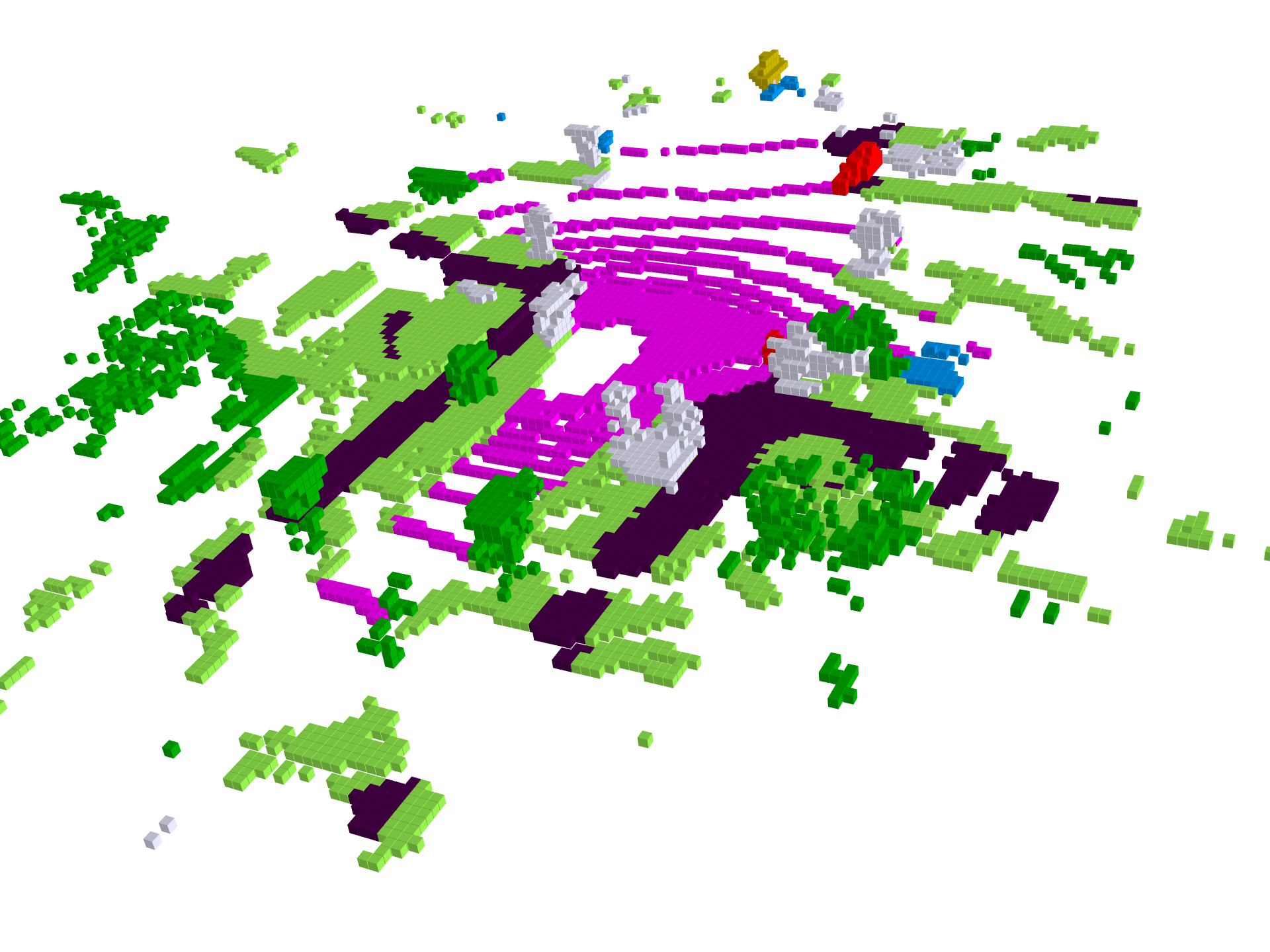}} &
			{\includegraphics[width=0.25\linewidth]{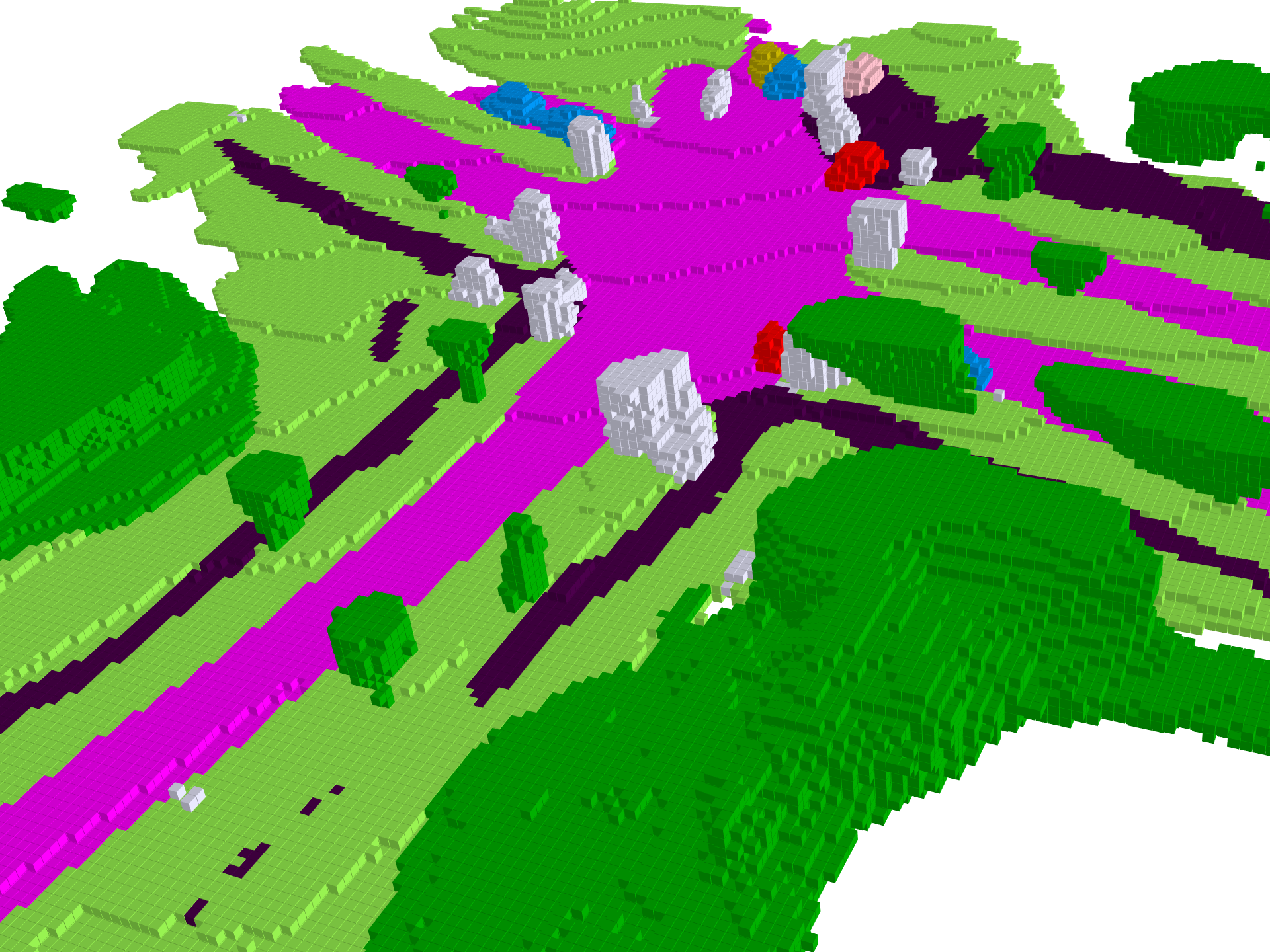}} &
			{\includegraphics[width=0.25\linewidth]{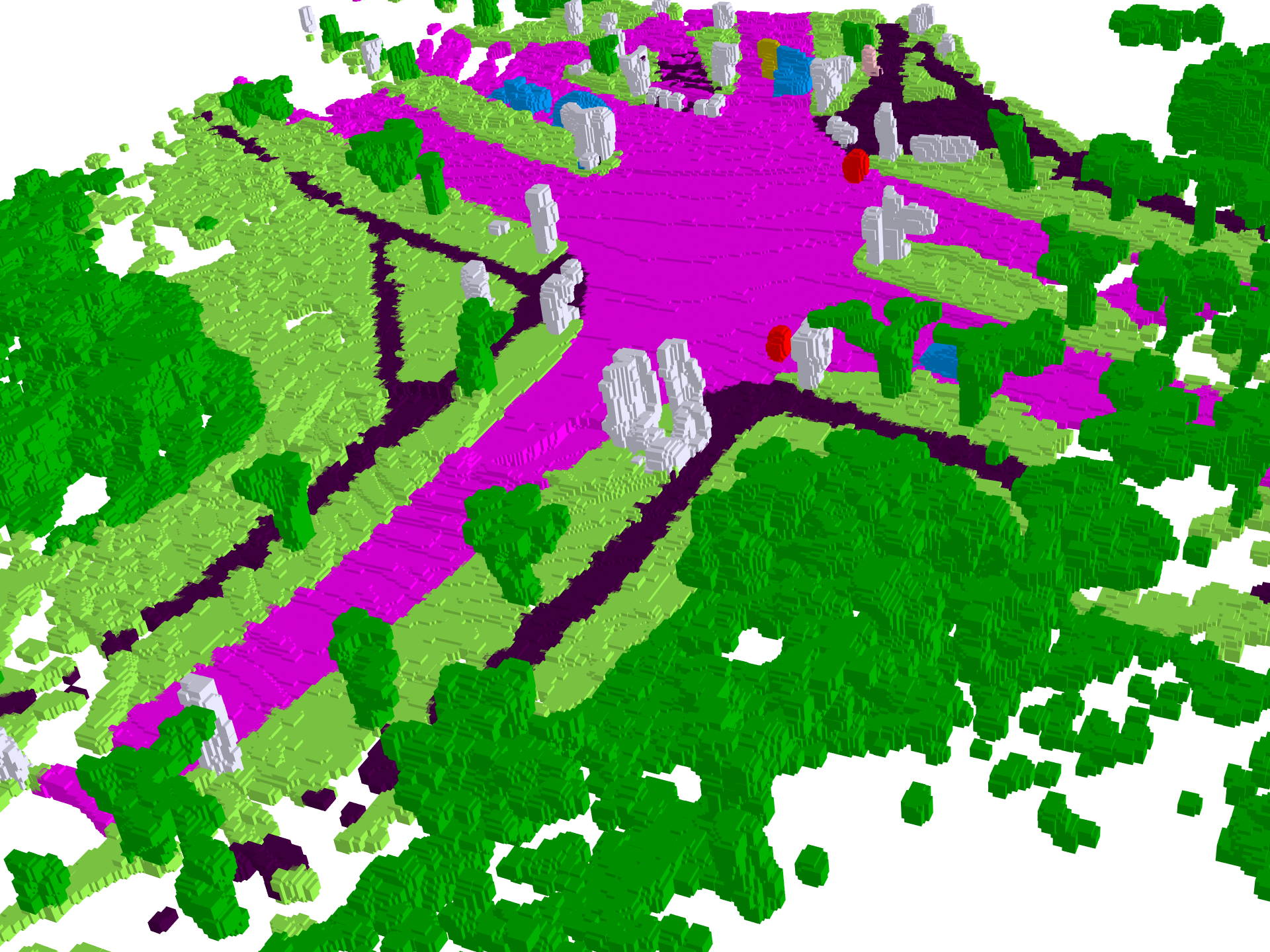}} \\
		\end{tabular}
		
		\begin{tabular}{cccccc}
			{\includegraphics[width=0.166\linewidth]{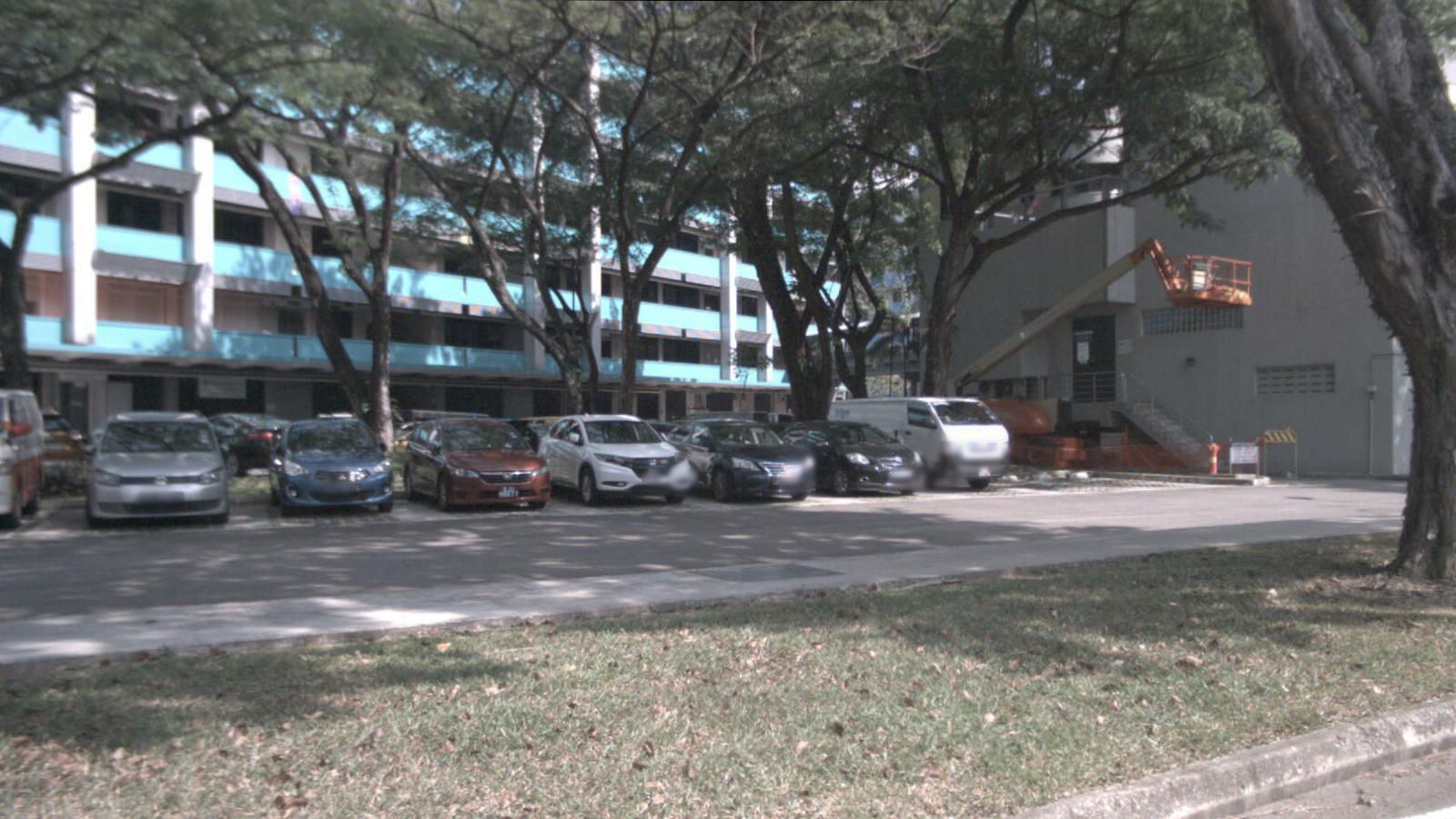}} &
			{\includegraphics[width=0.166\linewidth]{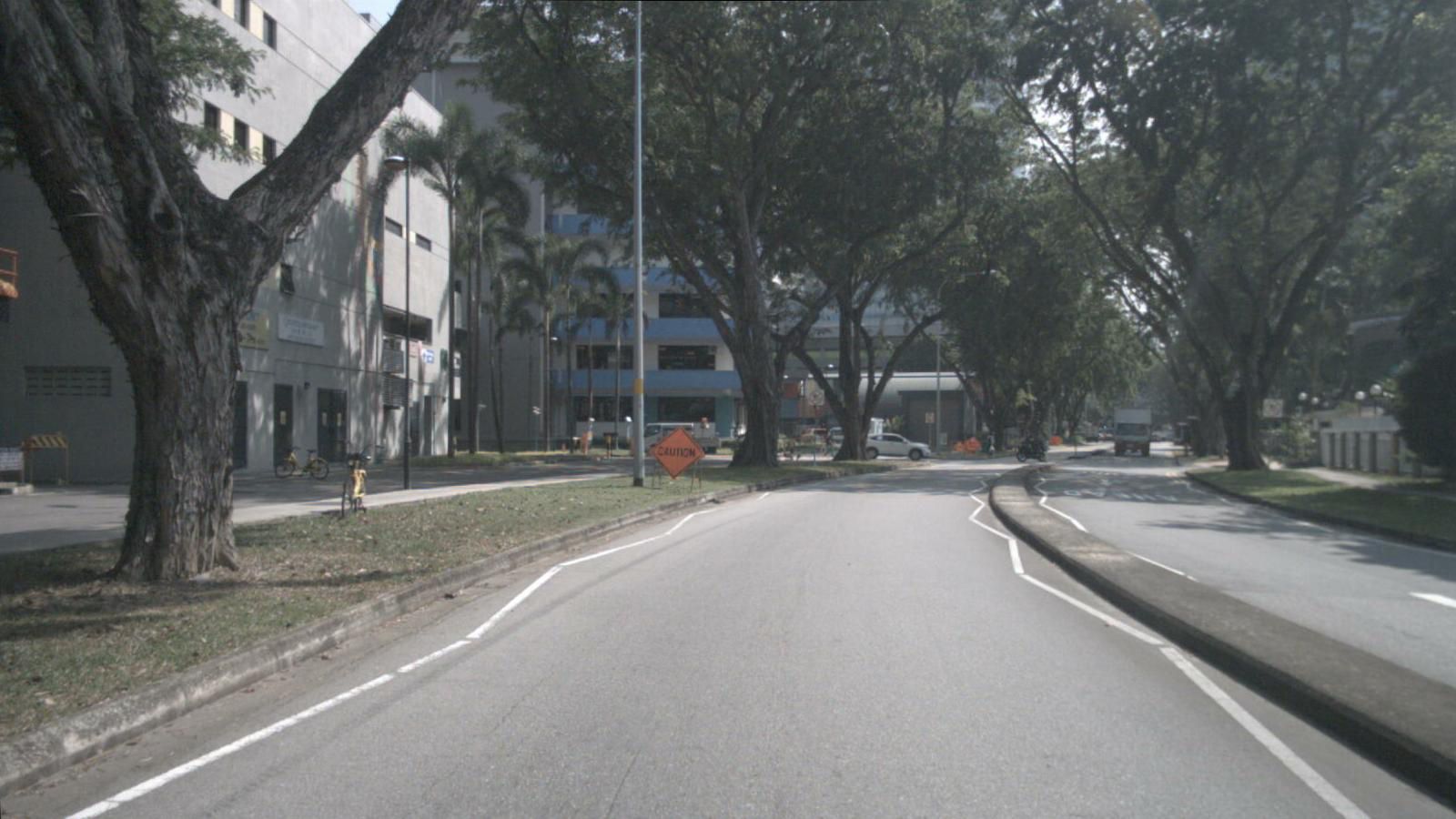}} &
			{\includegraphics[width=0.166\linewidth]{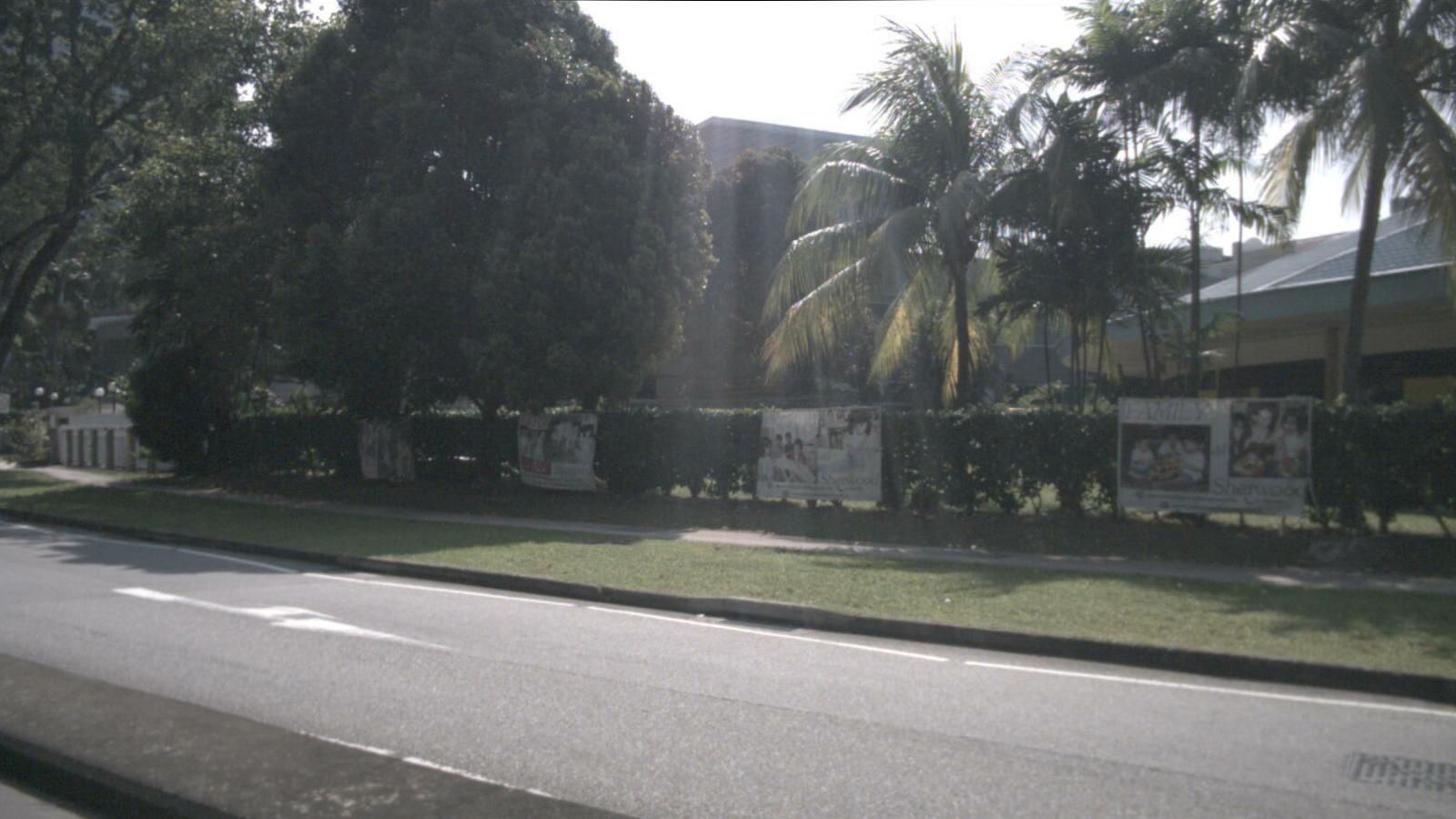}} &
			{\includegraphics[width=0.166\linewidth]{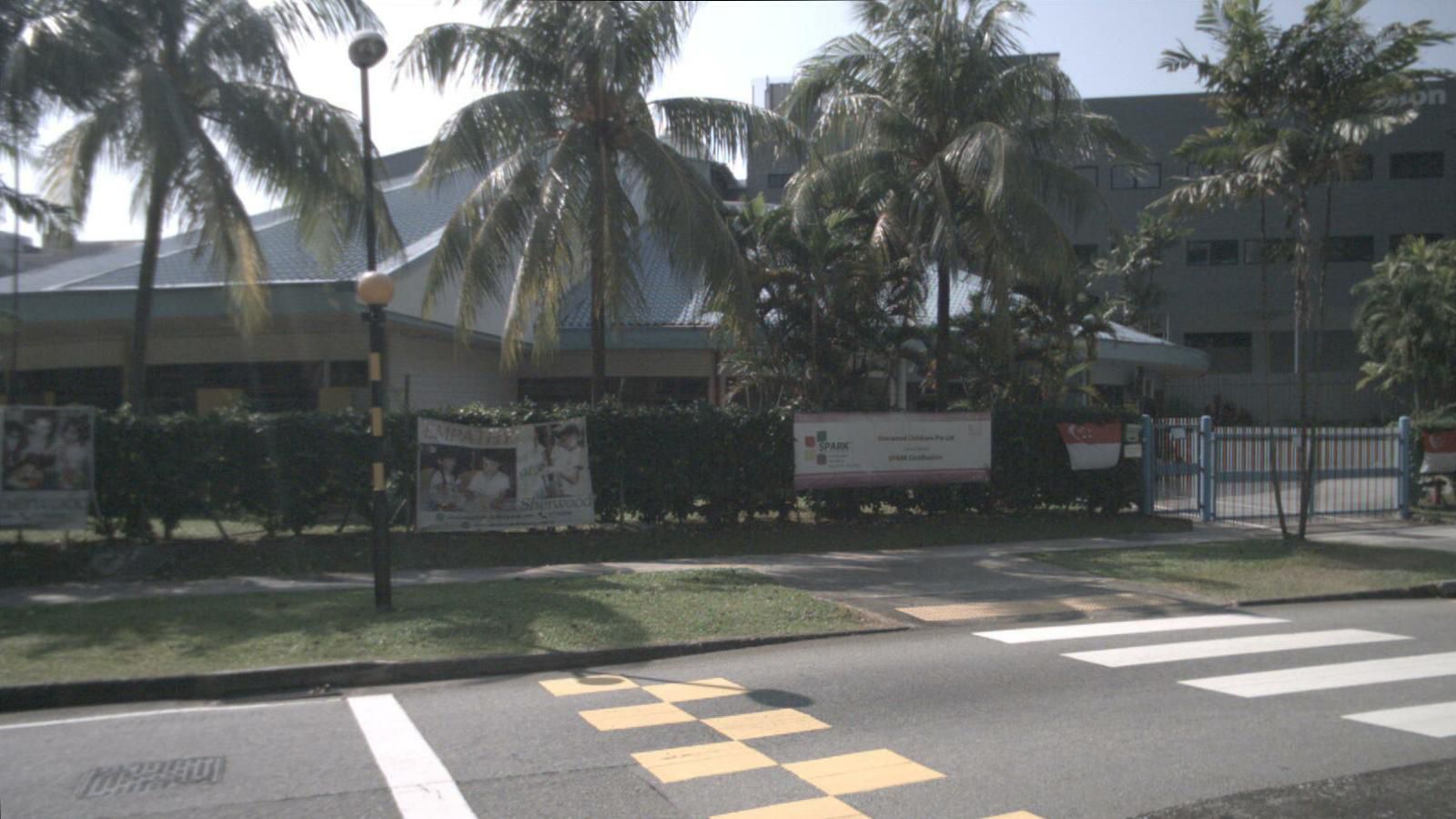}} &
			{\includegraphics[width=0.166\linewidth]{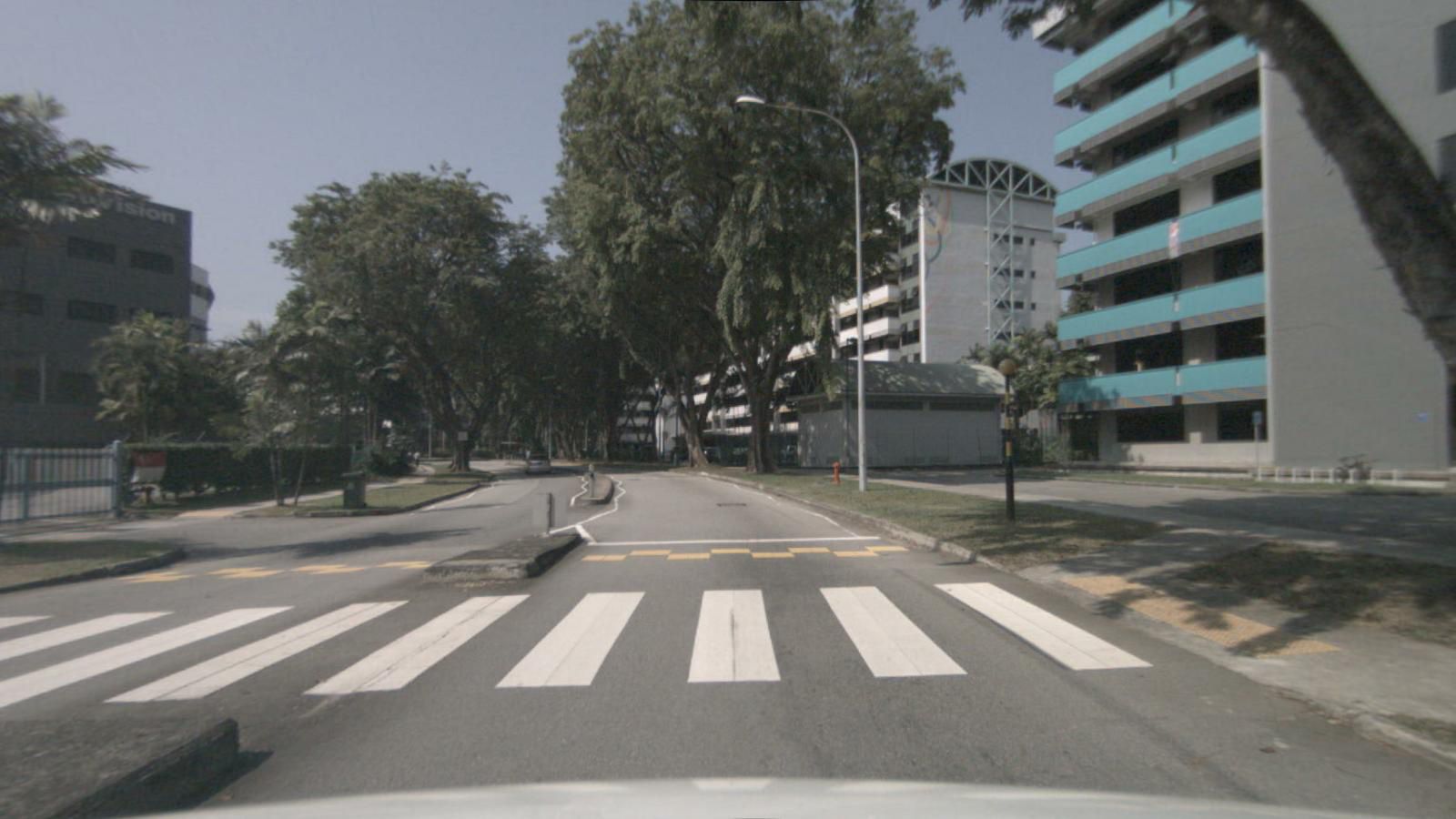}} &
			{\includegraphics[width=0.166\linewidth]{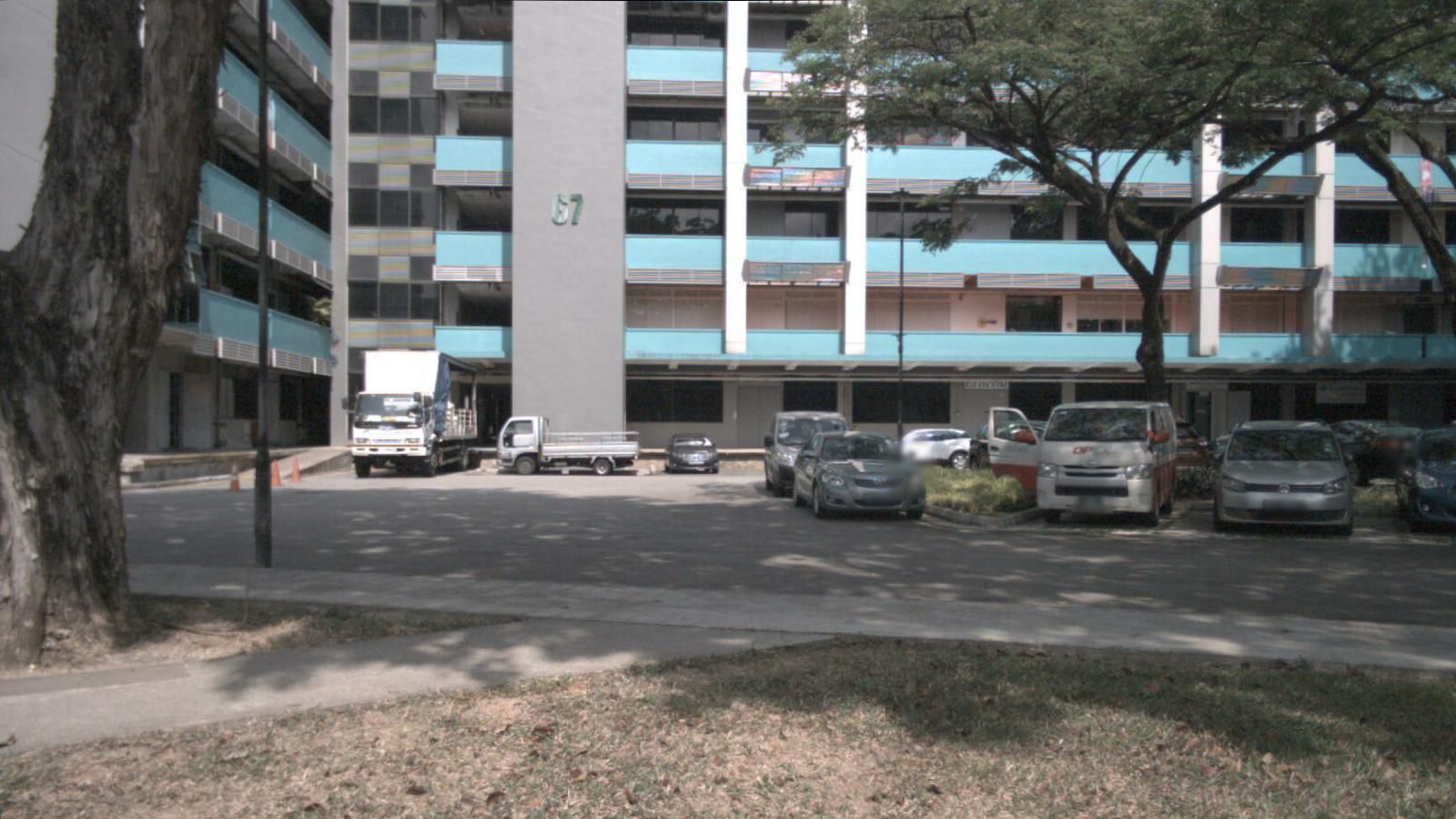}} \\
		\end{tabular}
		
		\begin{tabular}{cccc}
			{\includegraphics[width=0.25\linewidth]{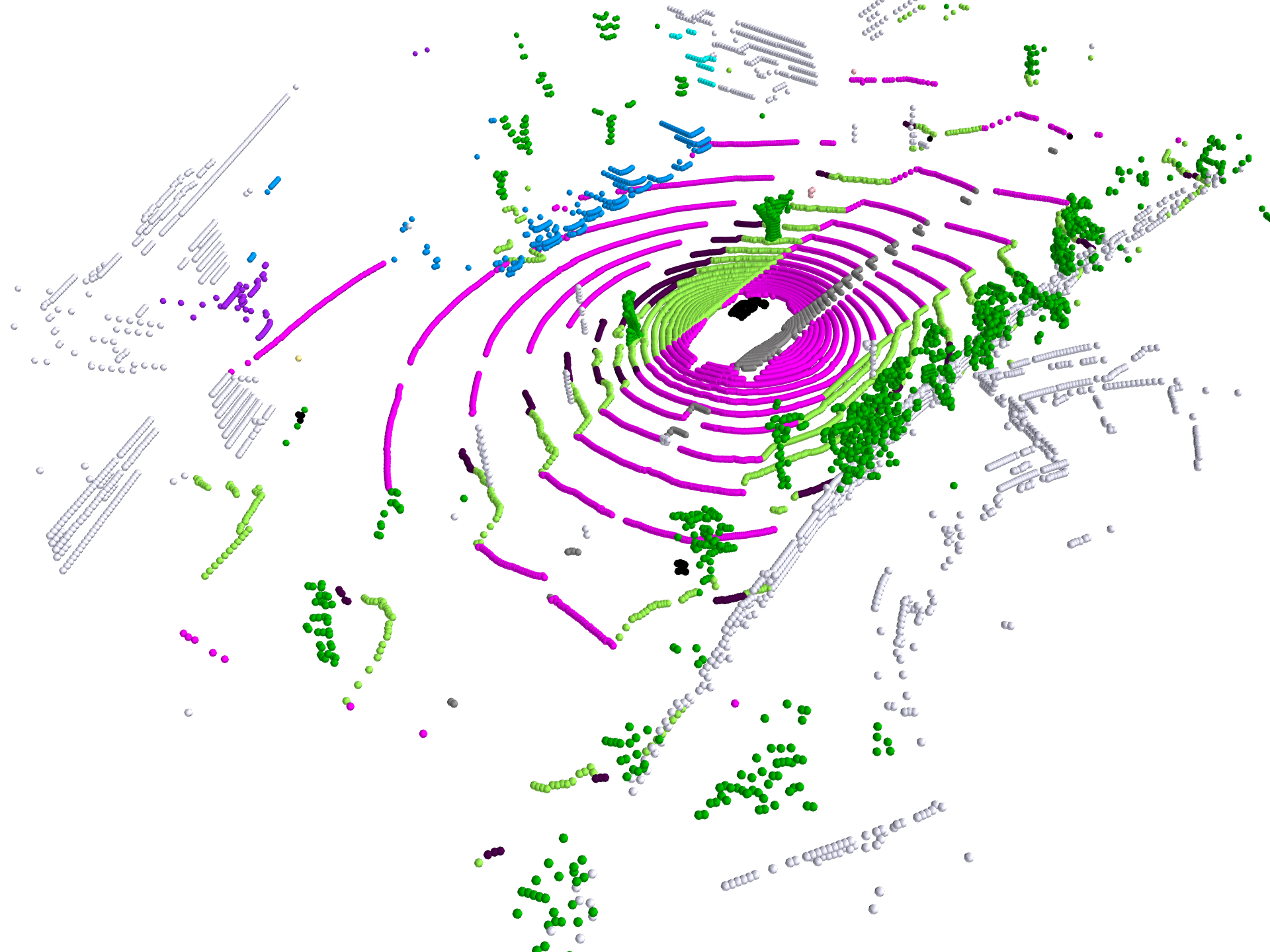}} &
			{\includegraphics[width=0.25\linewidth]{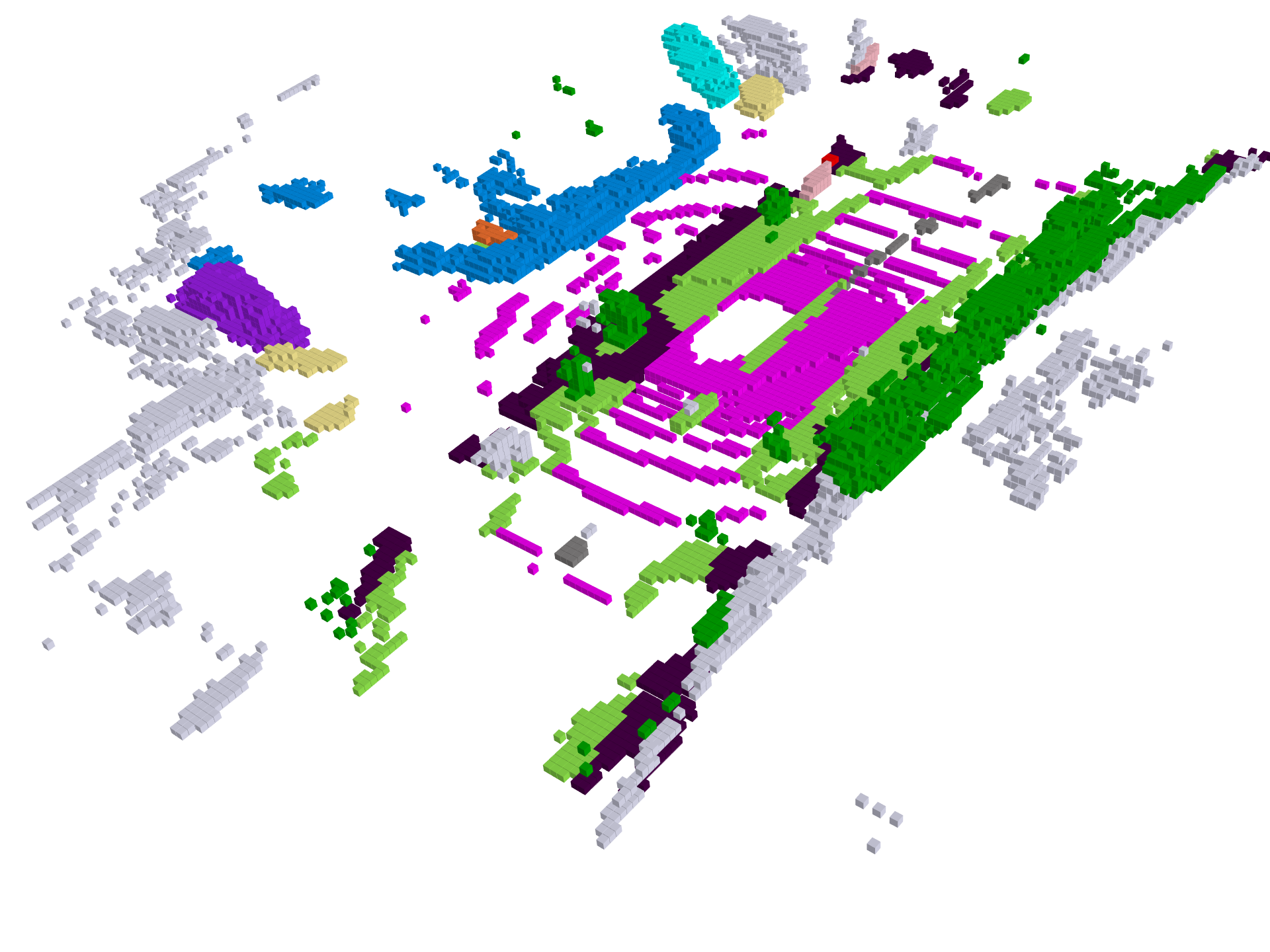}} &
			{\includegraphics[width=0.25\linewidth]{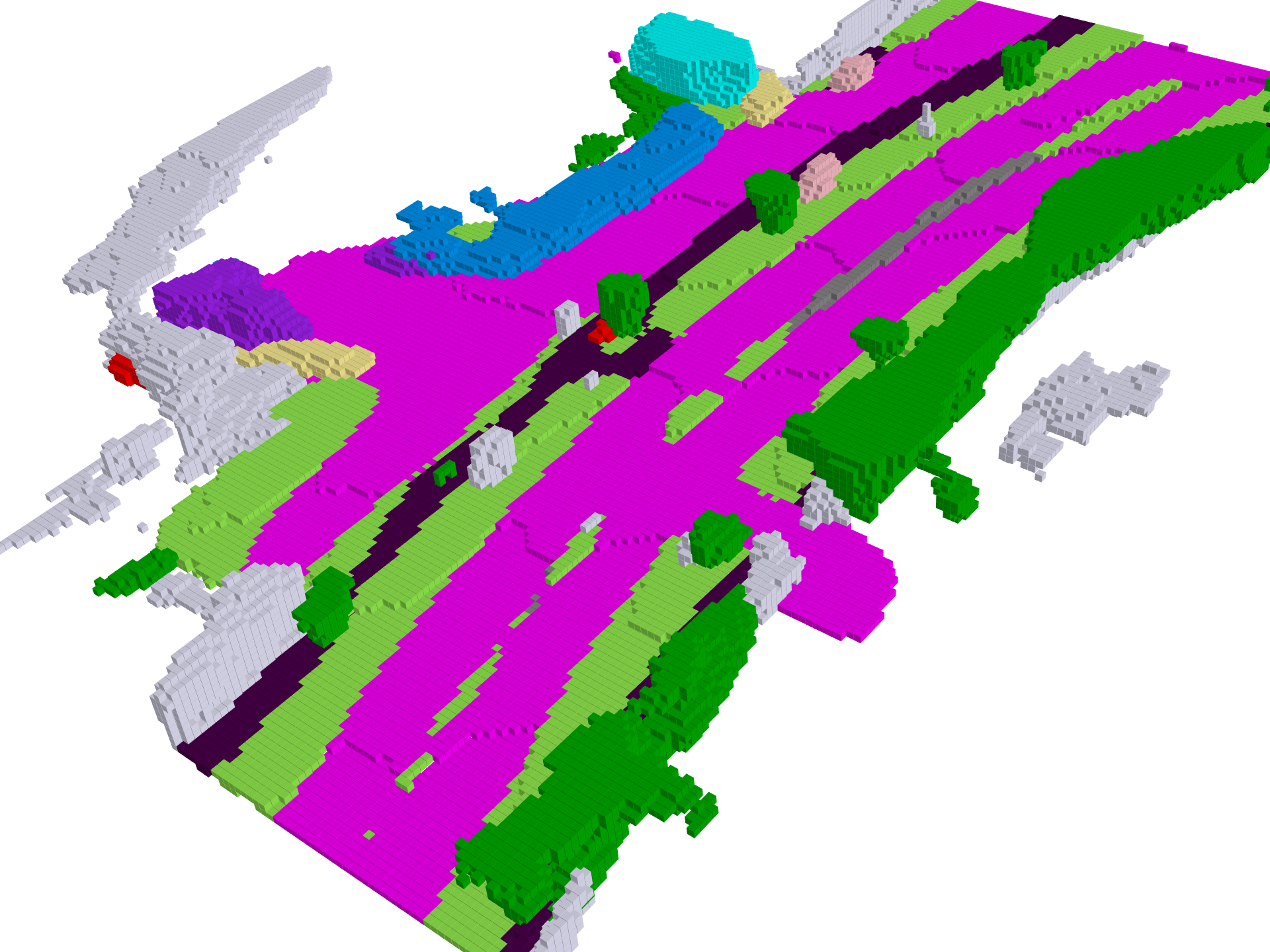}} &
			{\includegraphics[width=0.25\linewidth]{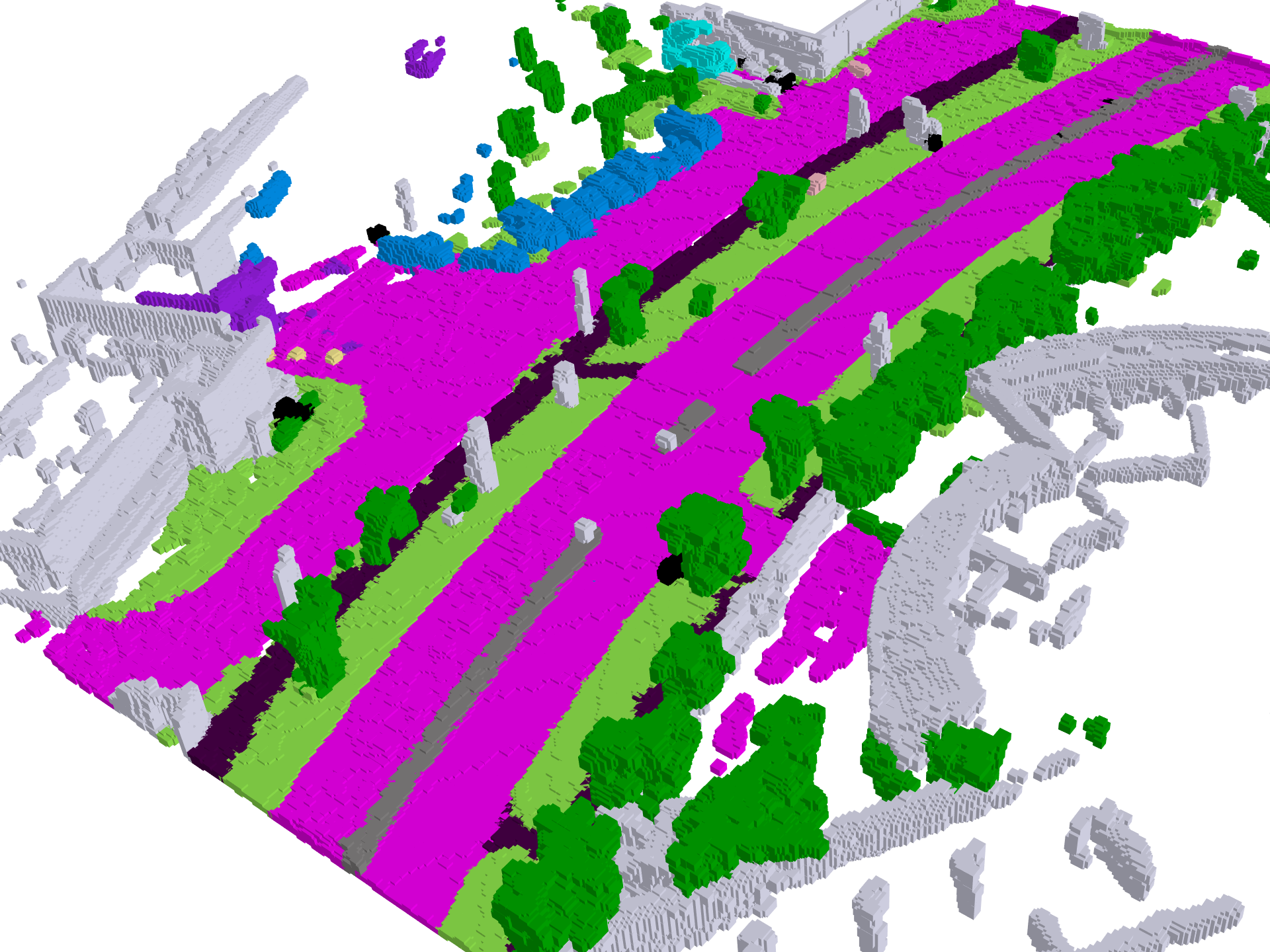}} \\
			Sparse LiDAR points & \makecell{Occupancy prediction \\ (with LiDAR GT)} & \makecell{Occupancy prediction \\ (with dense GT)} & Dense occupancy labels \\
		\end{tabular}

		\centering
		\caption{Visualizations on nuScenes validation set. Our generated dense occupancy labels are much denser than sparse LiDAR points. Trained with dense groundtruth, the network can predict better and denser occupancy. \textbf{Better viewed when zoomed in.}}
		\label{fig:seg2}
		\vspace{-3mm}
	\end{figure*}
	
	\subsection{Ablation Study}
	\noindent \textbf{2D-3D Spatial Attention:}
	Table \ref{tab:ab_1} shows the ablation results for 2D-3D spatial attention. Without spatial attention, we directly average all multi-camera features in a grid. However, we find this straightforward fusion method performs worse than spatial attention. The potential reason is that the contribution of each view is different for a 3D grid. Moreover, the ablation study shows that 3D-based cross-view attention is more effective than BEV-based cross-view attention since it can preserve 3D space information.
	
	\noindent \textbf{Multi-scale Occupancy Prediction:}
	We conduct an ablation study on multi-scale structure and multi-scale supervision in Table \ref{tab:ab_2}. The experimental results show that these two multi-scale designs can boost the performance. The multi-scale skip connection can help the network learn low-level fine-grained features, which is important for detailed high-resolution occupancy prediction. Moreover, the 3D volume features in all levels will be enhanced by multi-scale supervision.
	
	\noindent \textbf{Dense Occupancy Supervision:}
	The results in Table \ref{tab:ab_3} demonstrate the importance of using dense occupancy as ground truth. Compared with sparse LiDAR points, the sparse occupancy labels fuse multi-frame points and can provide more powerful supervision. The possion reconstruction and NN algorithm can fill the holes and further densify the occupancy labels, which will boost model's performance. Figure \ref{fig:seg2} can better illustrate the effectiveness of dense supervision. We can see that the model trained with our occupancy labels can predict much denser occupancy than that trained with LiDAR points.  

	\subsection{Model Efficiency}
	We compare the inference time and inference memory of different methods in Table \ref{tab:time}. The experiments are conducted on one RTX 3090 with six multi-camera images, whose resolutions are 1600x900. We find that our method can achieve both high performance and efficiency. Compared with BEVFormer, our method slightly
	increases inference time and memory and we think the increased burden is acceptable
	
	\begin{table}[ht]
		\centering
		\resizebox{0.43\textwidth}{!}{
			\begin{tabular}{l|cc}
				\hline
				Method & Latency (s) & Memory (G)\\ \hline
				SurroundDepth \cite{wei2022surrounddepth} & 0.73 & 12.4\\
				NeWCRFs \cite{yuan2022new} & 1.07 & 14.5\\
				Adabins \cite{bhat2021adabins} & 0.75 & 15.5\\
				BEVFormer \cite{bevformer} & \bf{0.31} & \bf{4.5}\\
				TPVFormer \cite{huang2023tri} & 0.32 & 5.1\\
				MonoScene \cite{monoscene} & 0.87& 20.3\\ \hline
				Ours &0.34 & 5.9\\
				\hline
			\end{tabular}
		}
		\vspace{1mm}
		\caption{The model efficiency of different methods. The experiments are conducted on one RTX 3090 with six multi-camera images, whose resolutions are 1600x900.}
		\label{tab:time}
	\end{table}
	
	\section{Limitations and Future Work}
	Currently, we only explore single-frame occupancy prediction. However, for the downstream modules, \eg motion prediction and planning, occupancy flow is more important. As the future work, we will design a framework to build occupancy flow dataset and utilize multi-frame surrounding images as the inputs. Moreover, LiDAR data is not always available. Self-supervised occupancy prediction with only RGB data is a valuable but challenging direction. 
	
	\section{Conclusion}
	In this paper, we propose SurroundOcc for multi-camera 3D occupancy prediction. We utilize 2D-3D spatial attention to integrate 2D features to the 3D volume in a multi-scale fashion, which is further upsampled and fused by the 3D deconvolution layer. Moreover, we devise a pipeline to generate dense occupancy ground truth. We stitch multi-frame LiDAR points of dynamic objects and static scenes separately and utilize Poisson Reconstruction to fill the holes. The comparison on nuScenes and SemanticKITTI datasets demonstrates the superiority of our method.
	
	\section*{Acknowledgement}
	This work was supported in part by the National Key Research and Development Program of China under Grant 2022ZD0160102 and in part by the National Natural Science Foundation of China under Grant 62125603.
	
	\section*{Appendix}
	\appendix

	\section{Baseline Method Details}
	We compare with several baseline methods on nuScenes dataset, which can be roughly classified as four categories: 
	
	\noindent \textbf{Depth estimation:} SurroundDepth \cite{wei2022surrounddepth}, AdaBins \cite{bhat2021adabins}, NeWCRFs \cite{yuan2022new}. Since SurroundDepth method is multi-camera self-supervised method, we use depth groundtruth to supervise the network along with self-supervised photometric loss. AdaBins \cite{bhat2021adabins} and NeWCRFs \cite{yuan2022new} are the state-of-the-art depth estimation methods both in outdoor and indoor scenes. To implement these two methods, we use their official released code with the dataloader in SurroundDepth. The depth results are fused by the TSDF fusion algorithm  \cite{curless1996volumetric, newcombe2011kinectfusion} with the voxel size $0.5m$, which is same to our method. 
	
	\noindent \textbf{3D scene reconstruction:} Atlas \cite{atlas} and Transformerfusion \cite{transformerfusion}. These two methods are state-of-the-art indoor scene reconstruction methods. We use our dense occupancy groundtruth to supervise them instead of tsdf ground truth.  To fairly compare, we also adopt ResNet101-DCN \cite{he2016deep, dai2017deformable} with the initial weight from FCOS3D \cite{fcos3d} as the backbone to extract image features. 
	
	\noindent \textbf{Occupancy reconstruction:} MonoScene \cite{monoscene} and TPVFormer \cite{huang2023tri}. To extend MonoScene to multi-camera setting, we project occupancy labels to each camera's coordinate and the shape of each camera's prediction is $(128, 104, 16)$ with 0.5m voxel size. We fuse multi-camera results in LiDAR coordinate with camera extrinsics. The final result has the same shape and voxel size with ours. For TPVFormer, the resolution is set as 200x200x16 and the feature dimension is 64.
	
	\noindent \textbf{BEV perception:} BEVFormer \cite{bevformer}. We use the full-resolution 200x200 BEV features. To lift BEV features to the 3D space, we split 256 dimensions BEV features as 16 grids and the feature of each grid has 16 dimensions. Then we adopt a 3D encoder-decoder network \cite{atlas} as a segmentation head to predict occupancy. Following the setting in TPVFormer, we employ both
	cross entropy loss and lovasz-softmax \cite{berman2018lovasz} as the supervision signals.

	\section{More Visualizations}
\begin{figure*}[tb]
		\centering
		\setlength\tabcolsep{1.0pt} 
		\renewcommand{\arraystretch}{1.0}
		
		\begin{tabular}{cccccc}
			
			\footnotesize FRONT LEFT & \footnotesize FRONT &  \footnotesize FRONT RIGHT & \footnotesize BACK RIGHT & \footnotesize BACK &\footnotesize BACK LEFT \\
			{\includegraphics[width=0.166\linewidth]{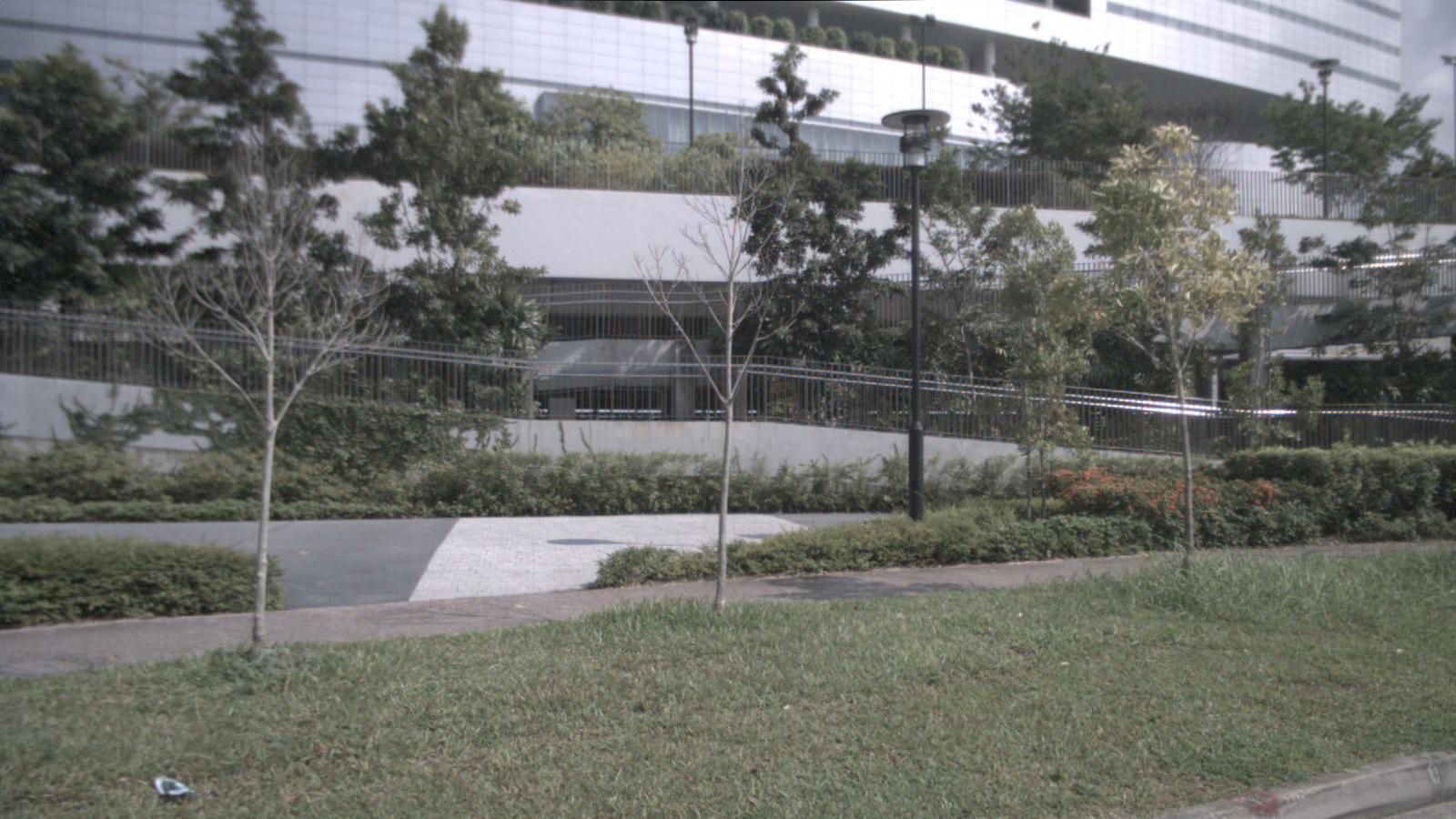}} &
			{\includegraphics[width=0.166\linewidth]{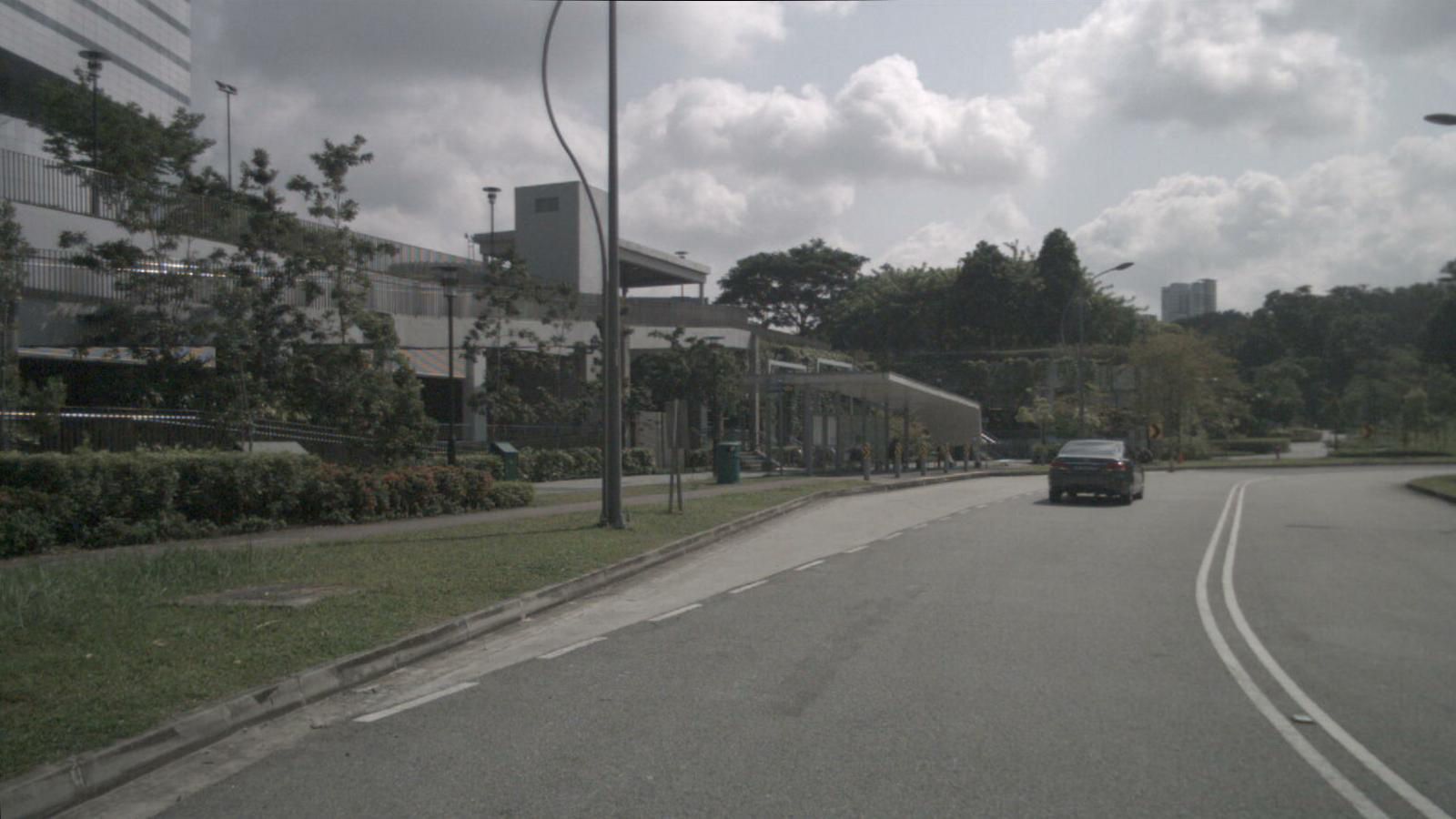}} &
			{\includegraphics[width=0.166\linewidth]{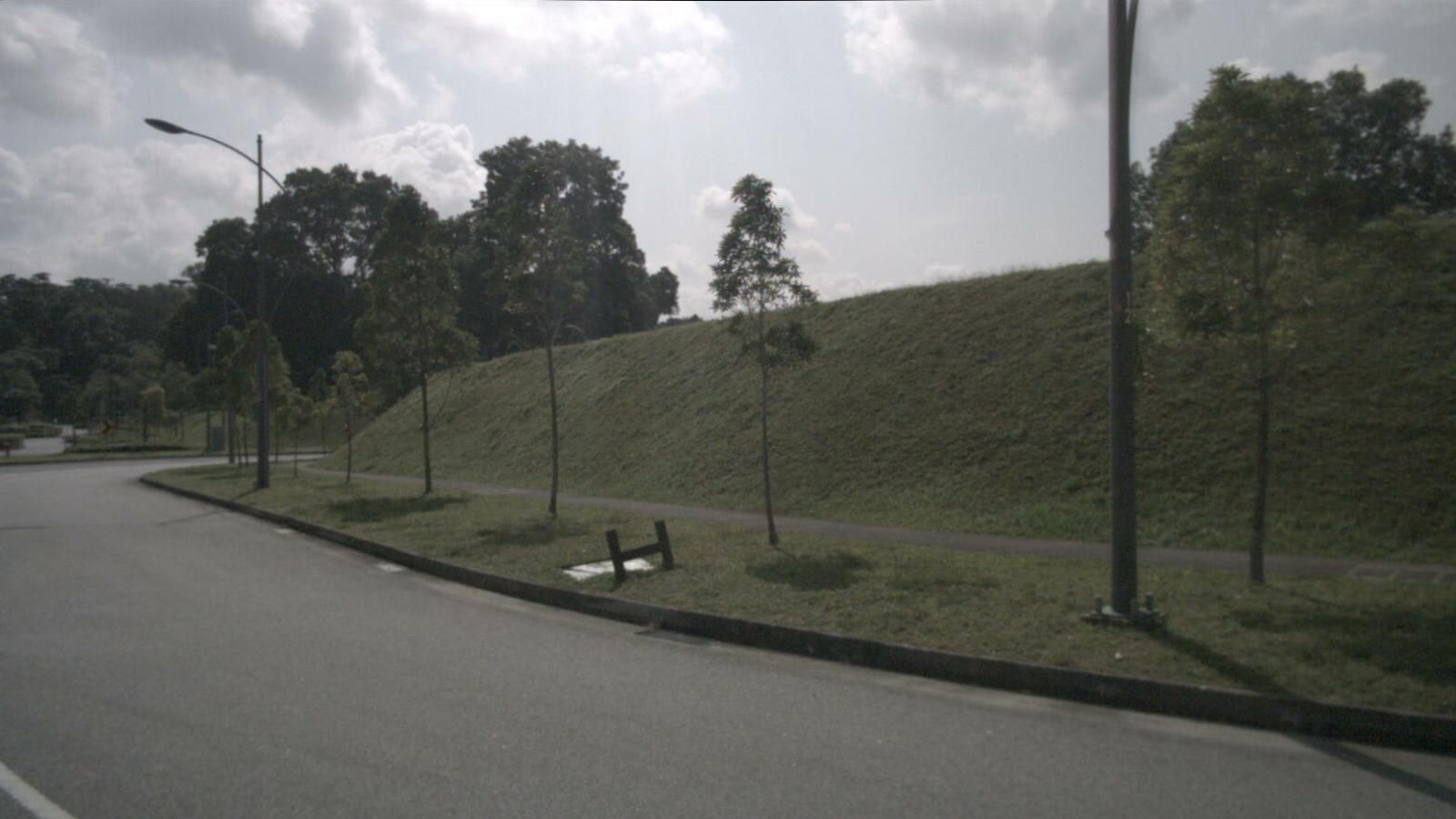}} &
			{\includegraphics[width=0.166\linewidth]{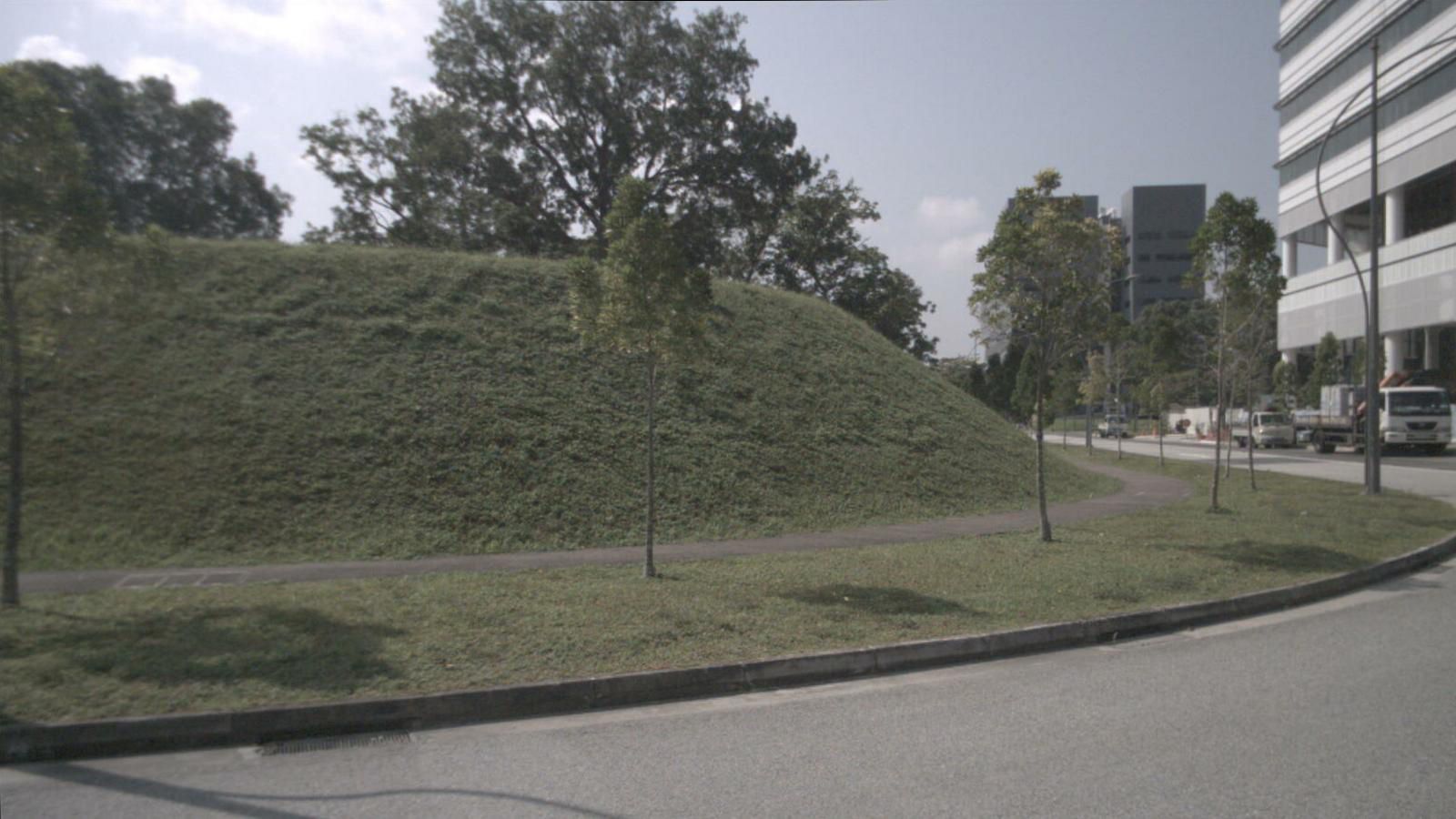}} &
			{\includegraphics[width=0.166\linewidth]{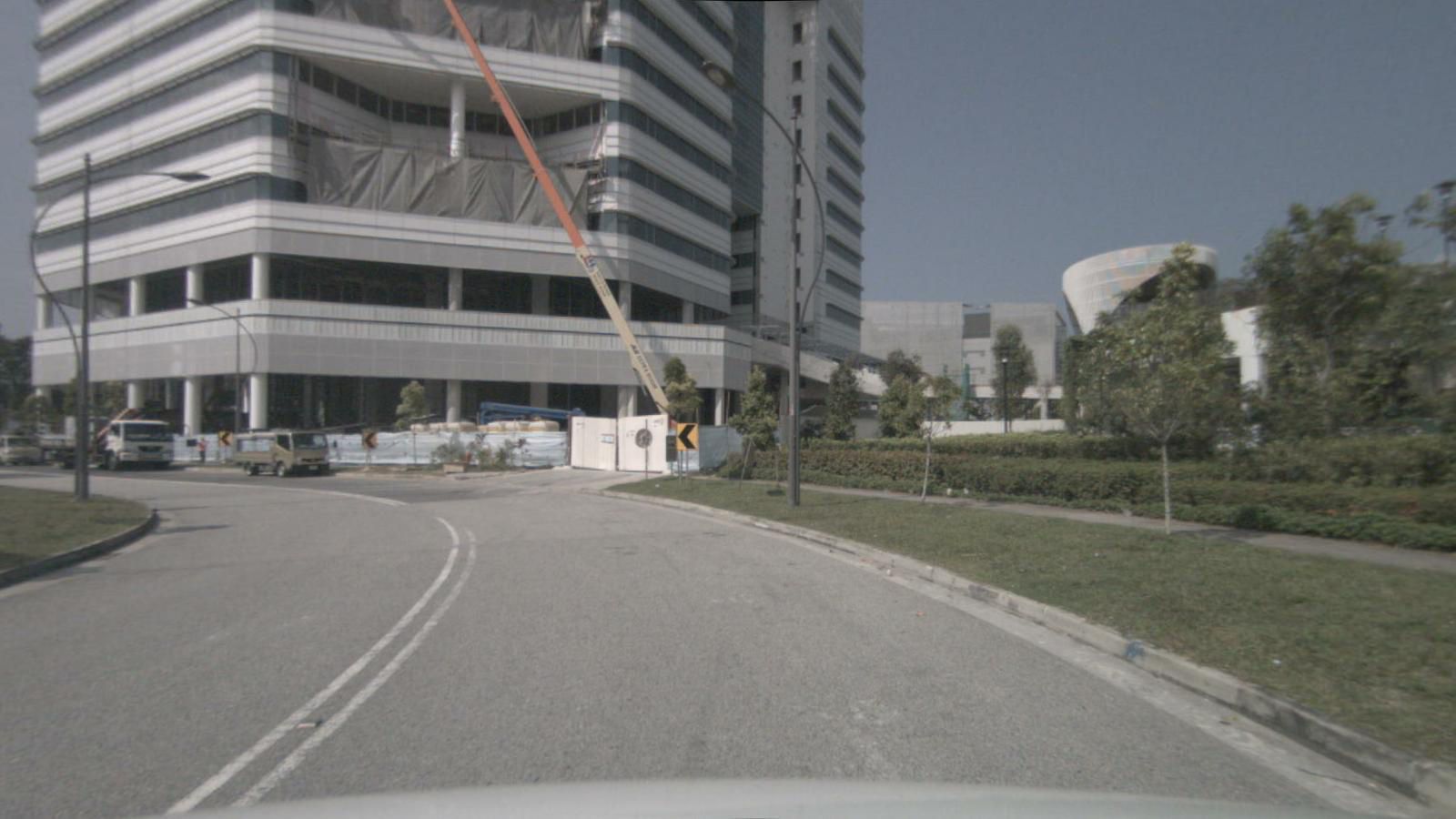}} &
			{\includegraphics[width=0.166\linewidth]{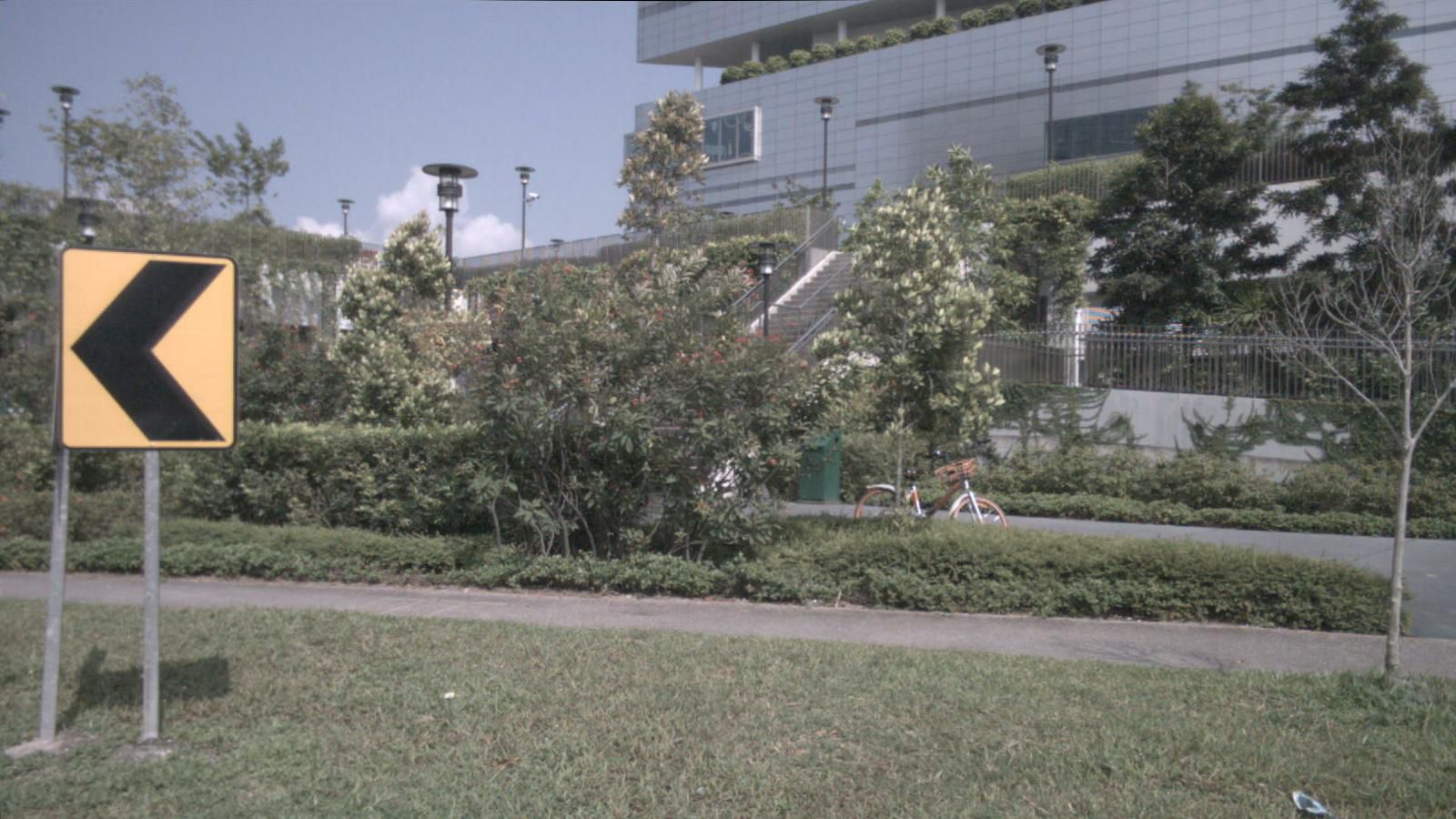}} \\
		\end{tabular}
		
		\begin{tabular}{cccc}
			{\includegraphics[width=0.25\linewidth]{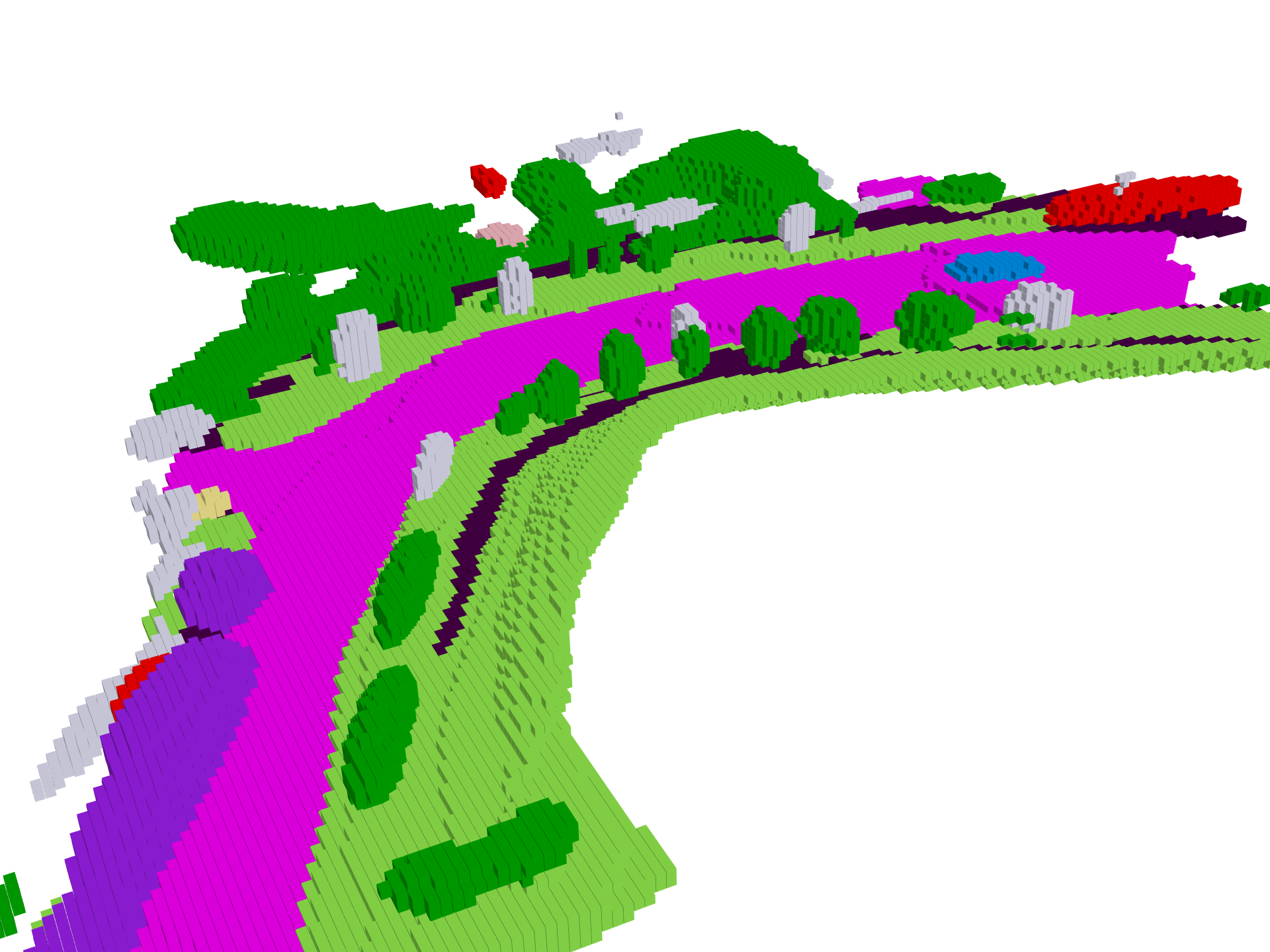}} &
			{\includegraphics[width=0.25\linewidth]{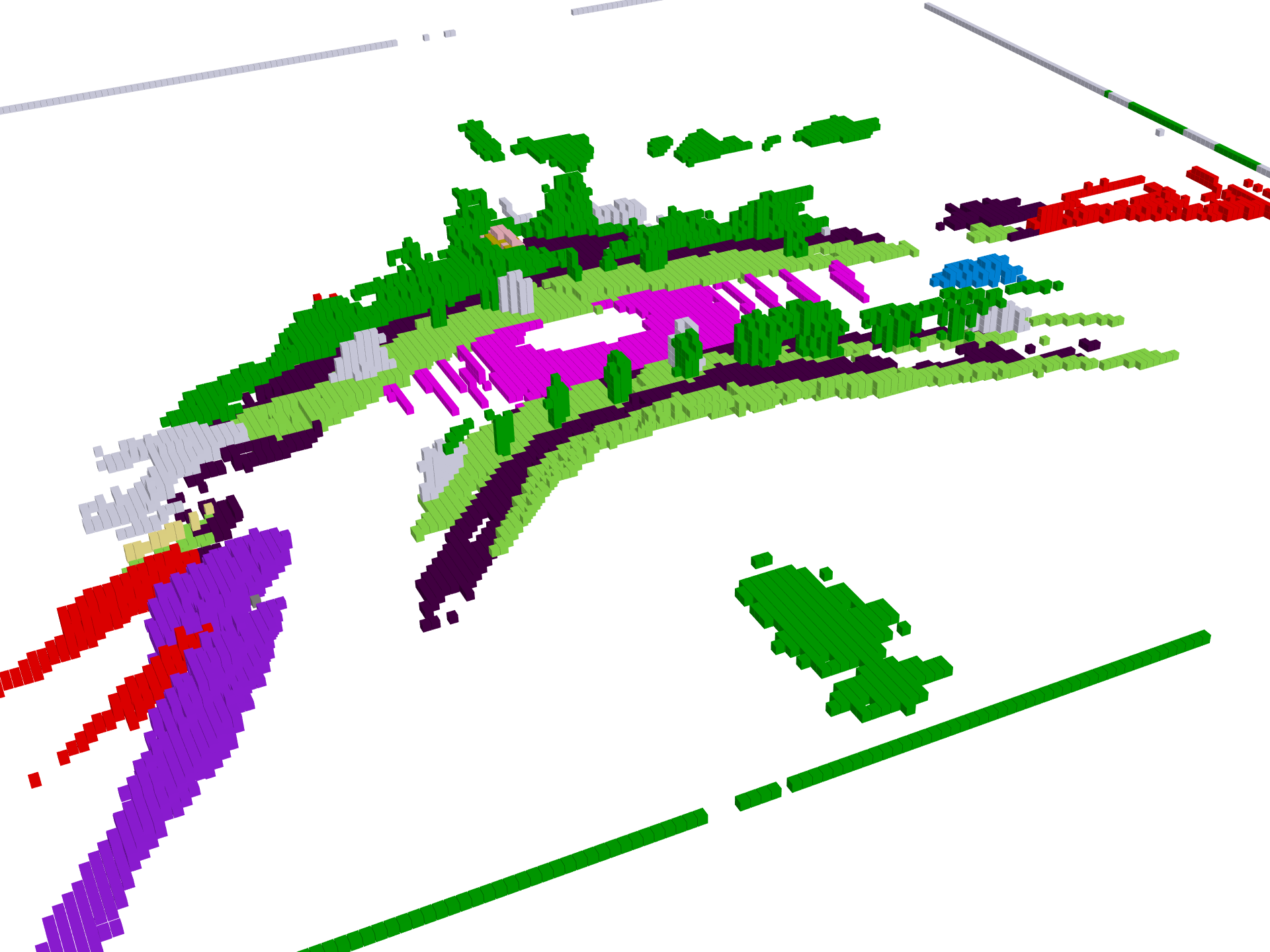}} &
			{\includegraphics[width=0.25\linewidth]{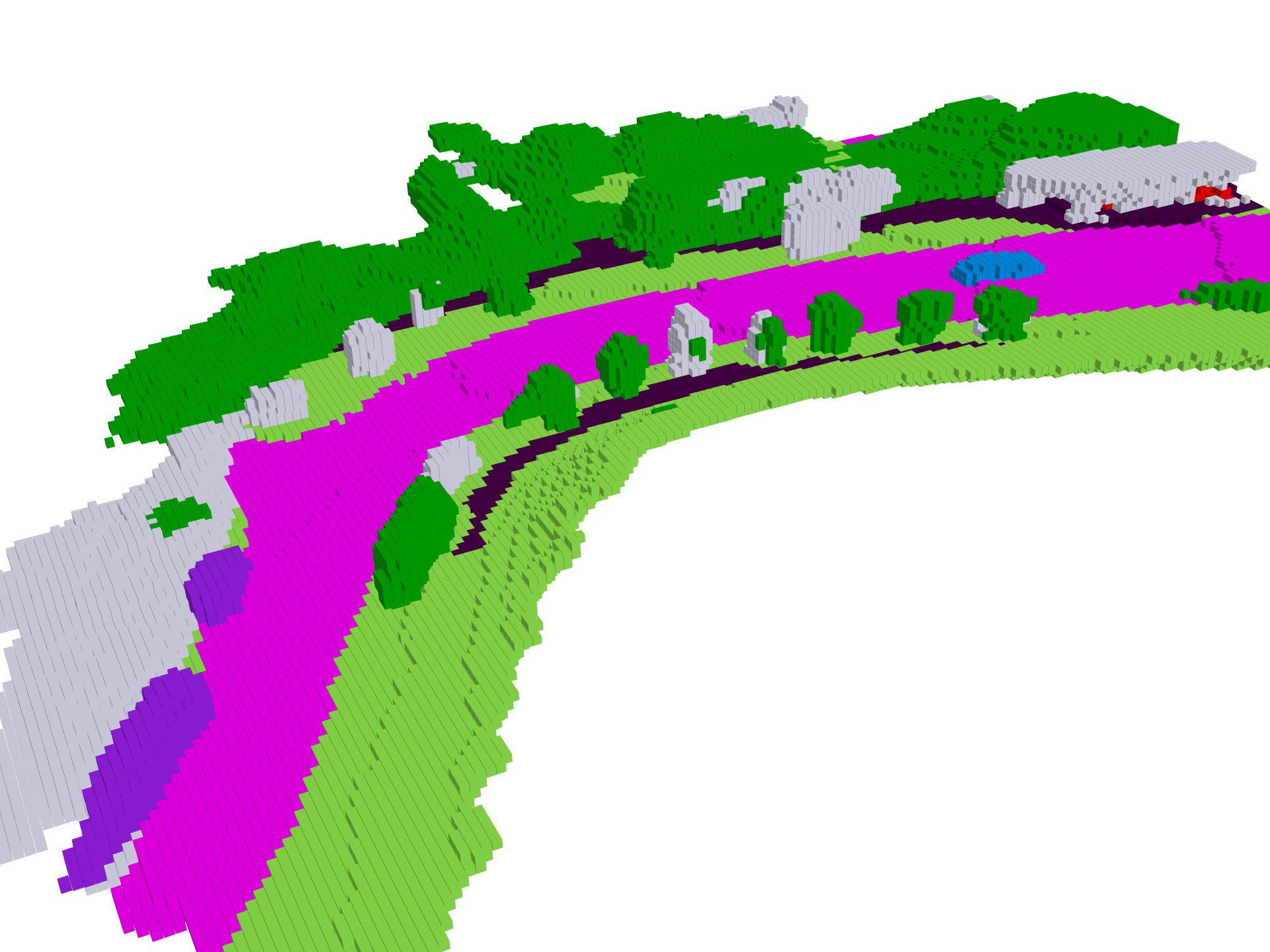}} &
			{\includegraphics[width=0.25\linewidth]{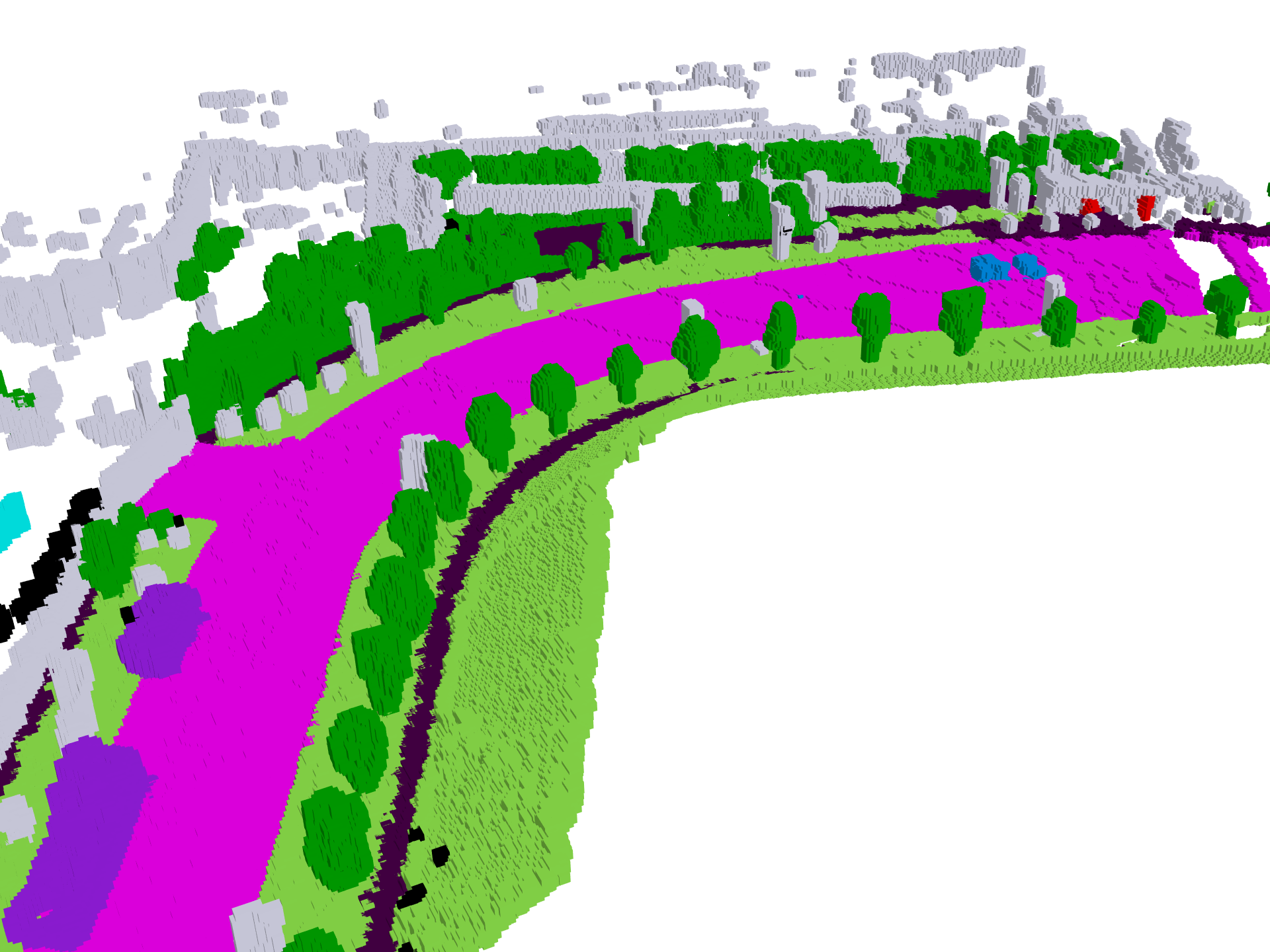}} \\
		\end{tabular}
		
		\begin{tabular}{cccccc}
			{\includegraphics[width=0.166\linewidth]{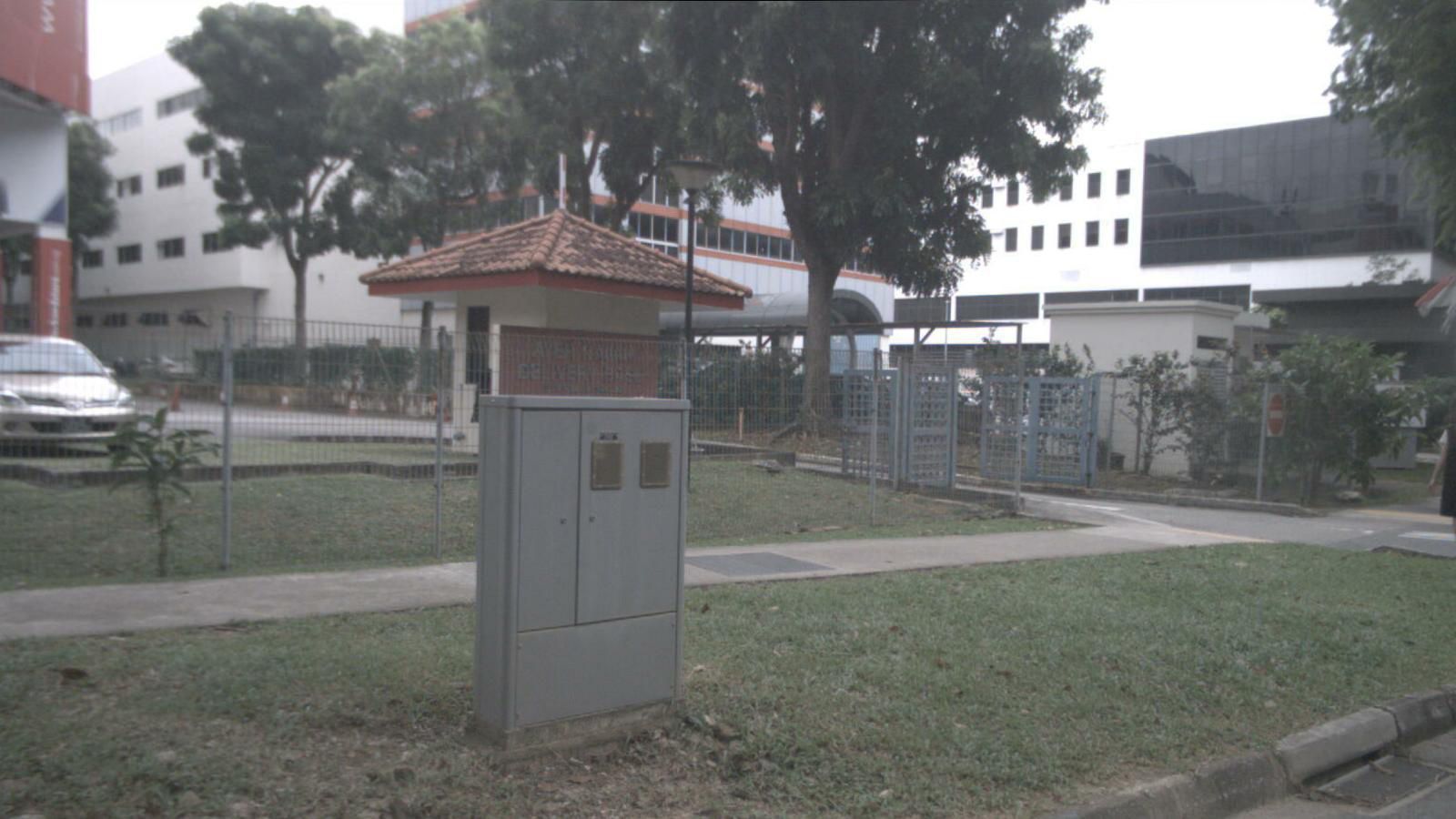}} &
			{\includegraphics[width=0.166\linewidth]{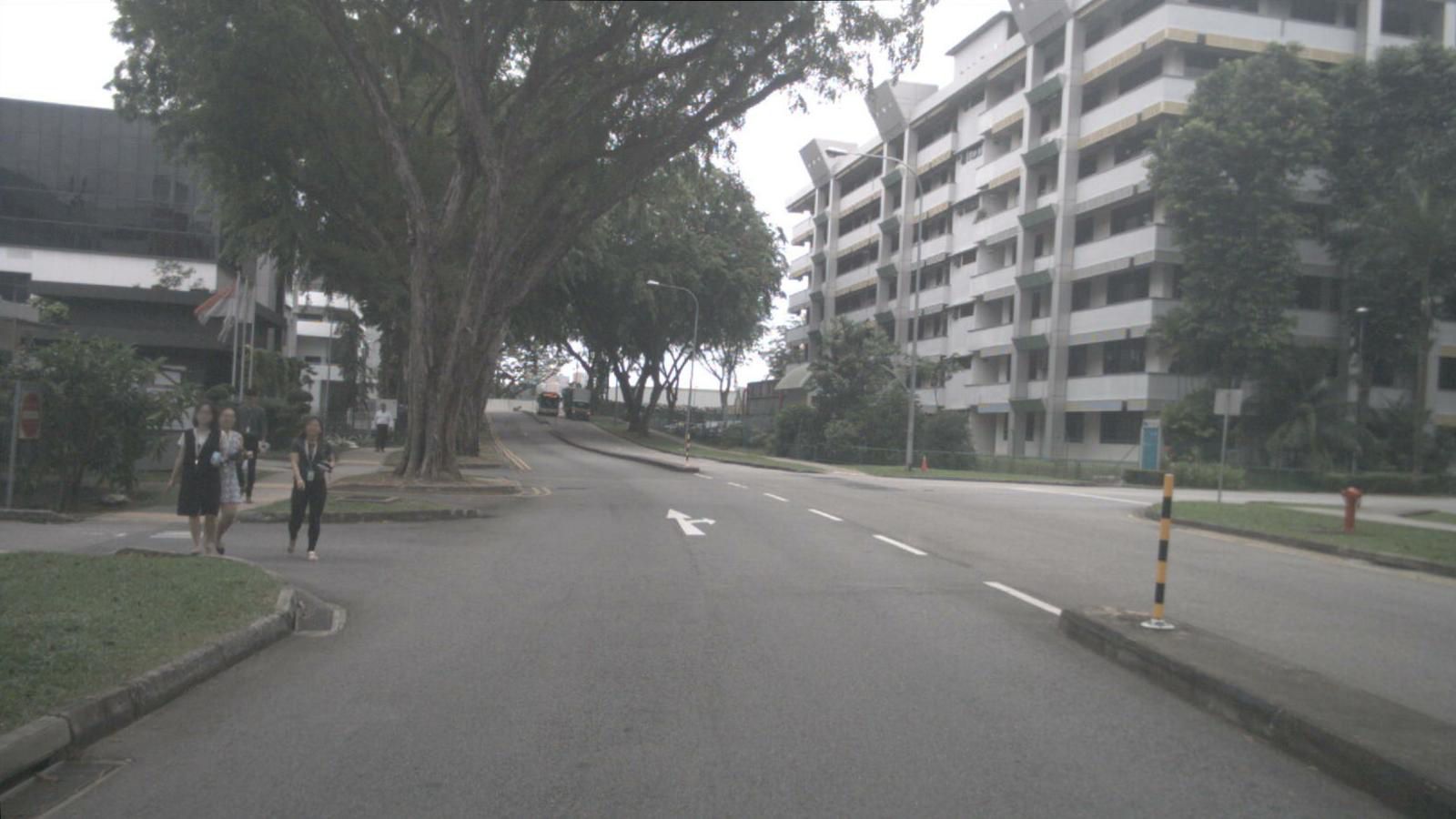}} &
			{\includegraphics[width=0.166\linewidth]{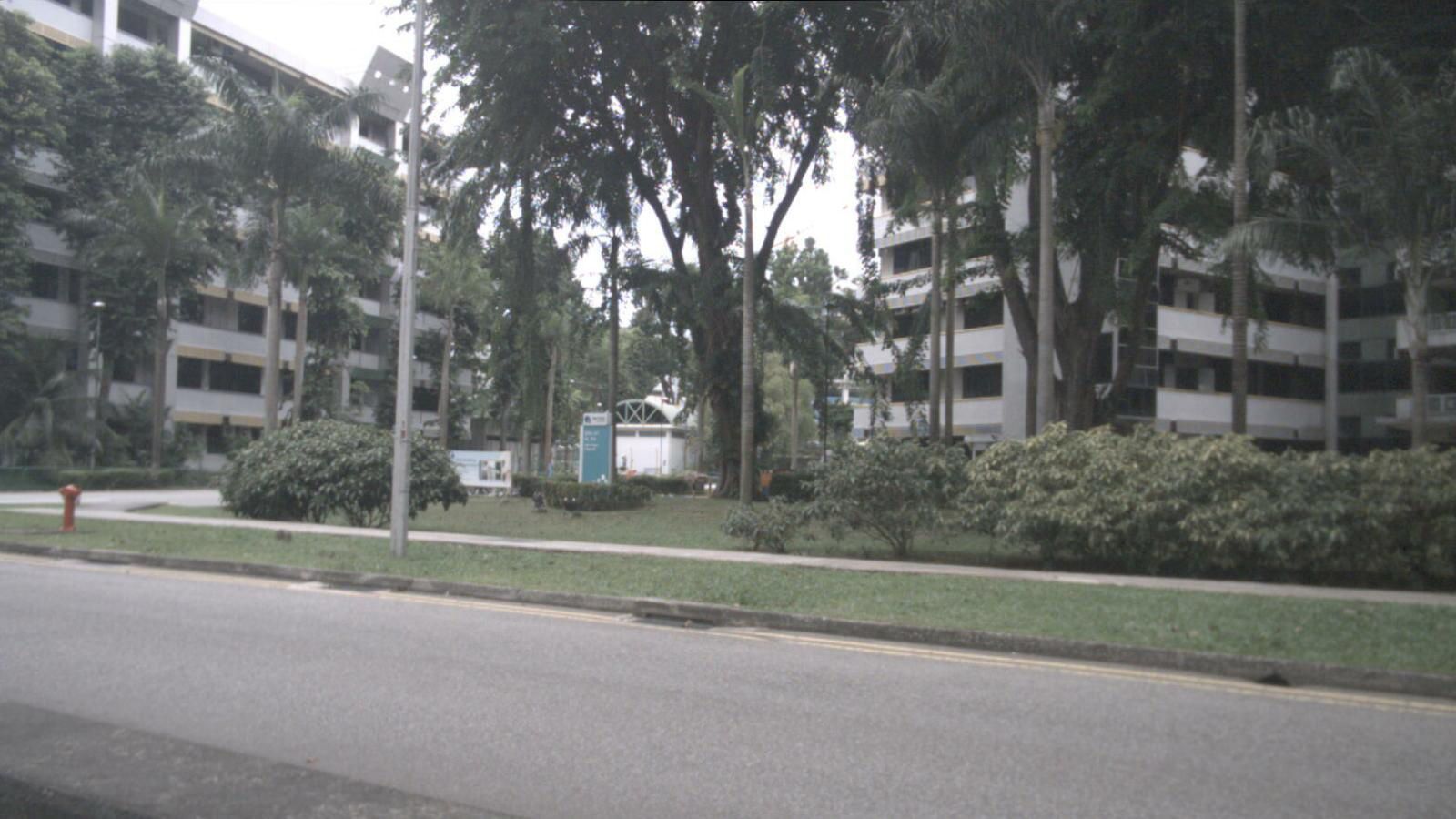}} &
			{\includegraphics[width=0.166\linewidth]{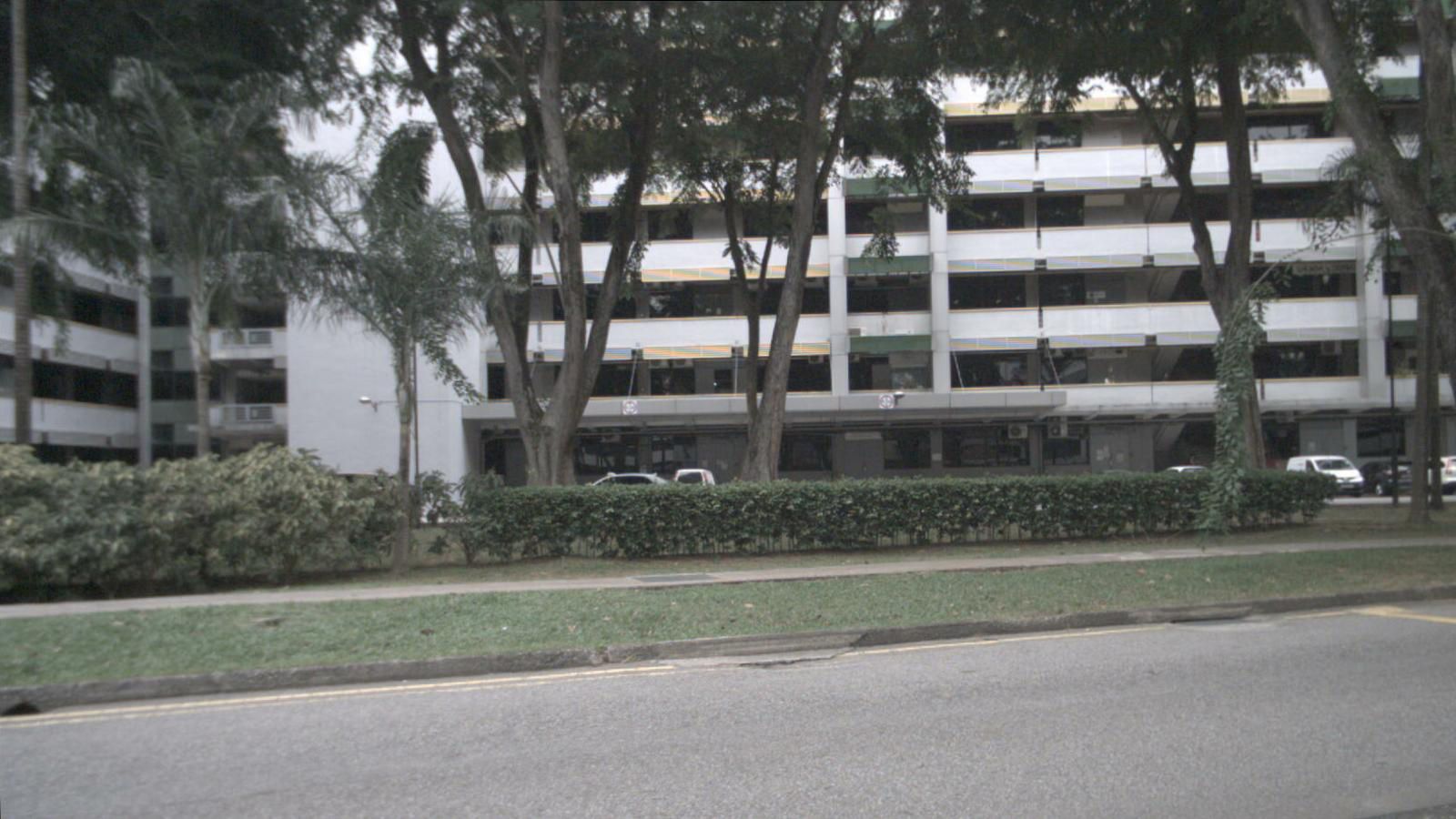}} &
			{\includegraphics[width=0.166\linewidth]{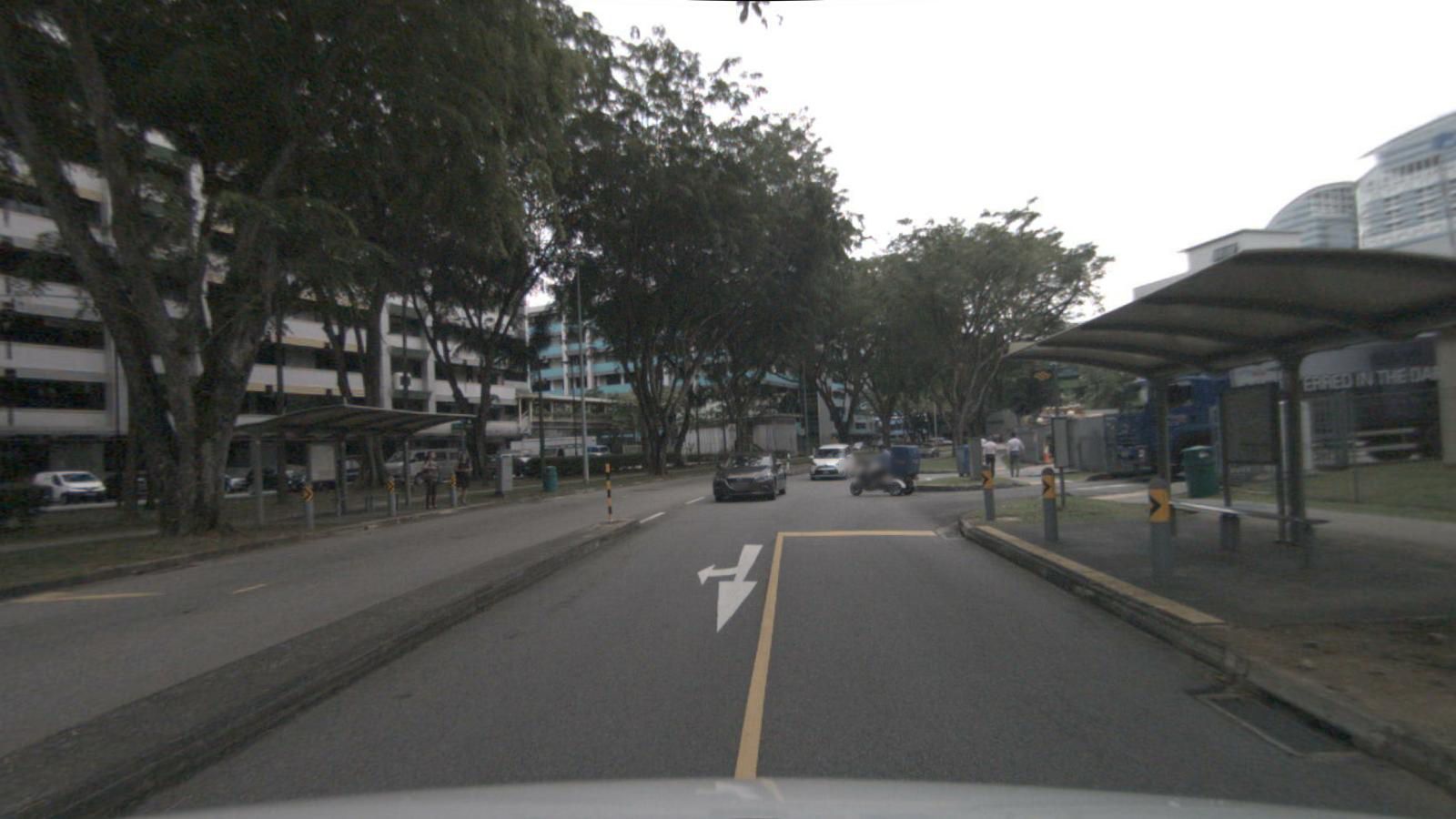}} &
			{\includegraphics[width=0.166\linewidth]{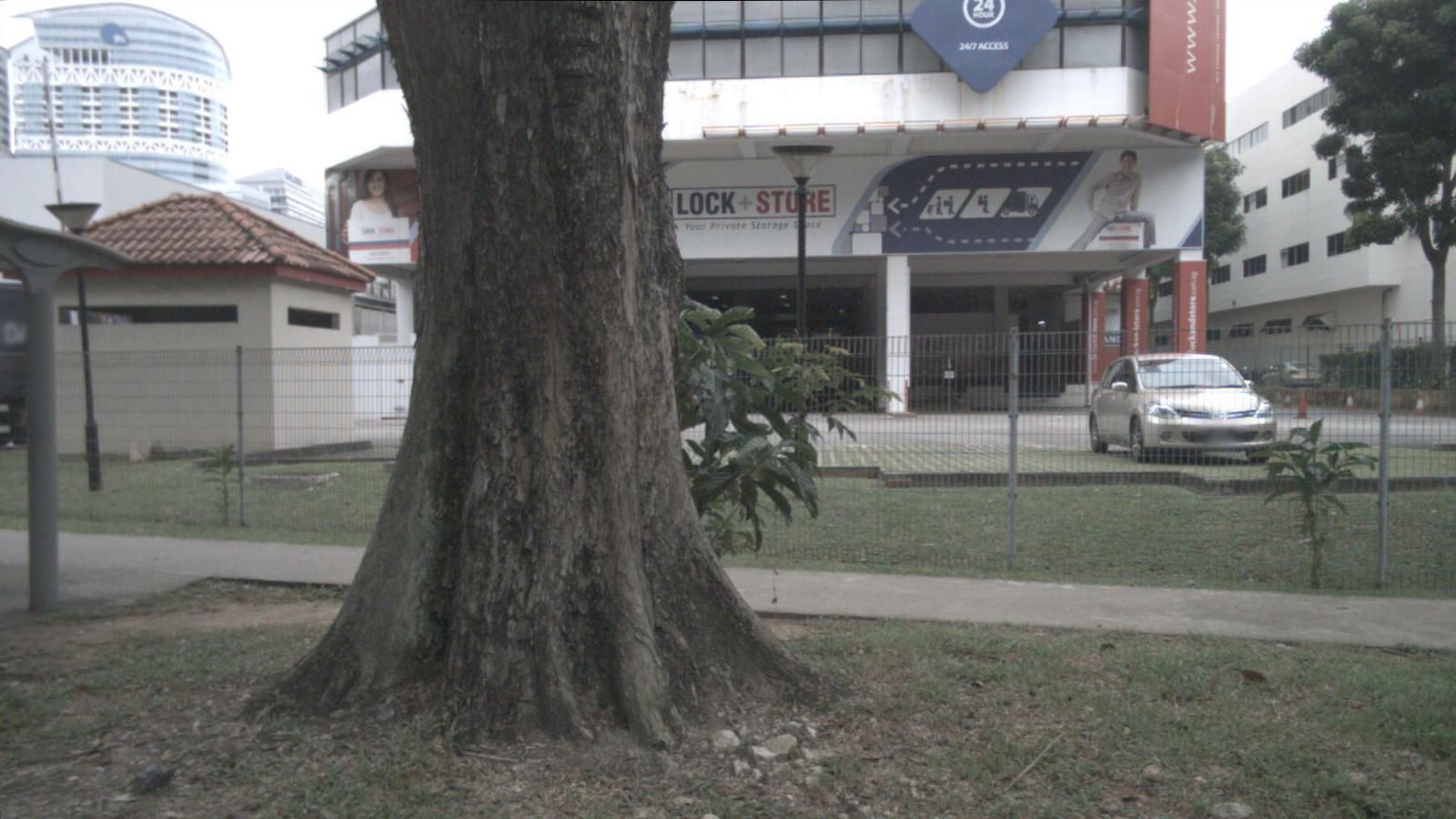}} \\
		\end{tabular}
		
		\begin{tabular}{cccc}
			{\includegraphics[width=0.25\linewidth]{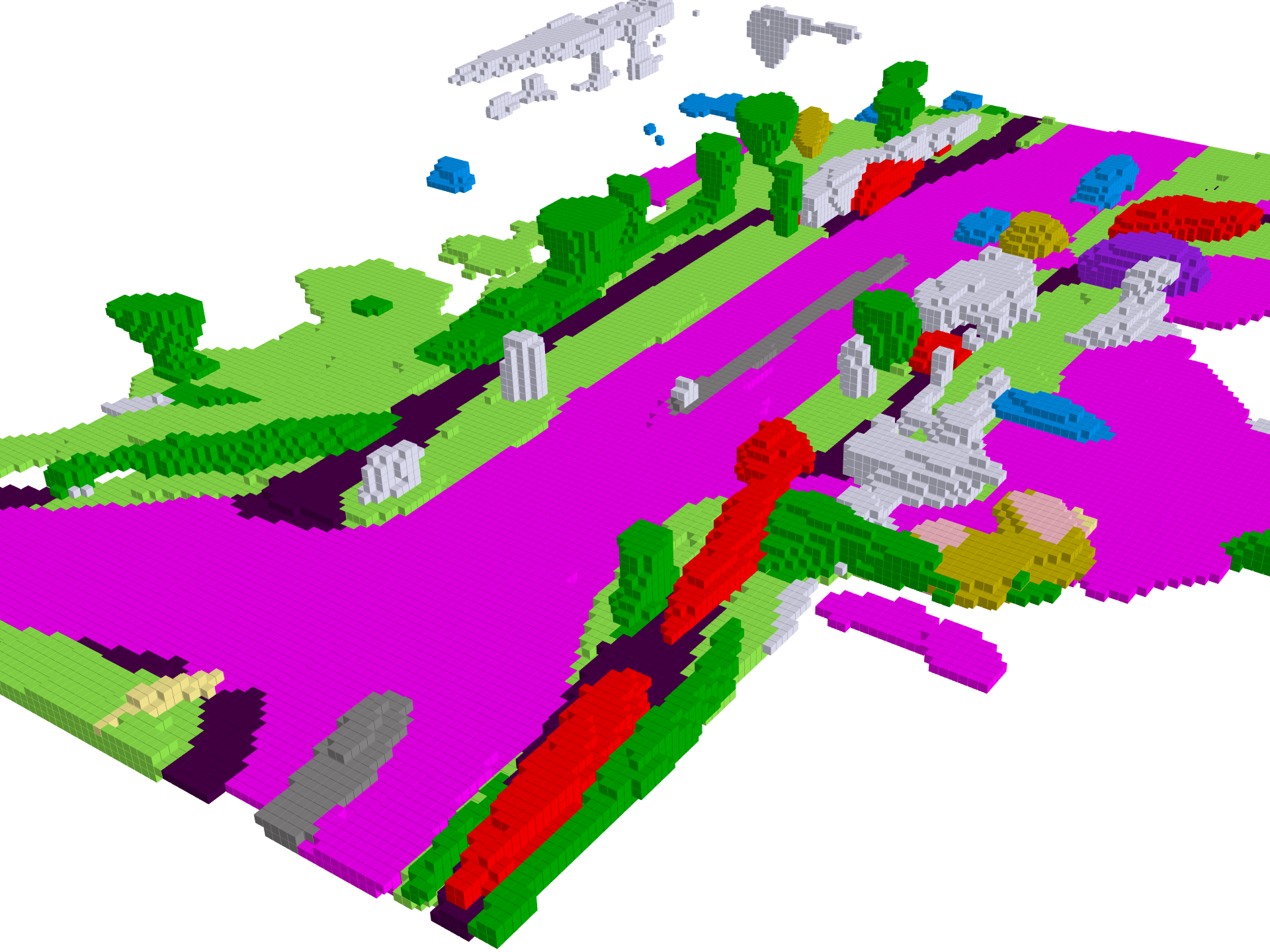}} &
			{\includegraphics[width=0.25\linewidth]{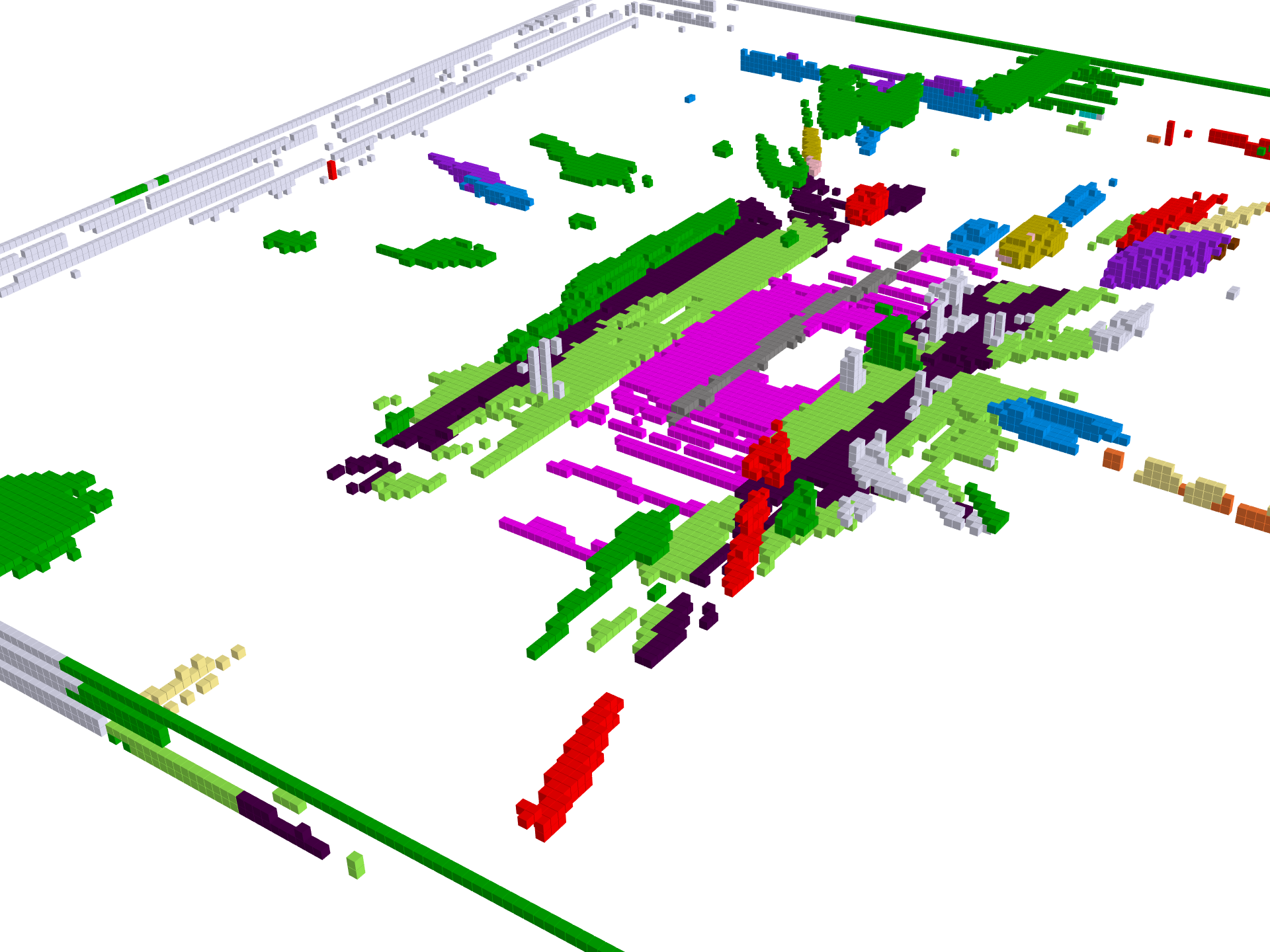}} &
			{\includegraphics[width=0.25\linewidth]{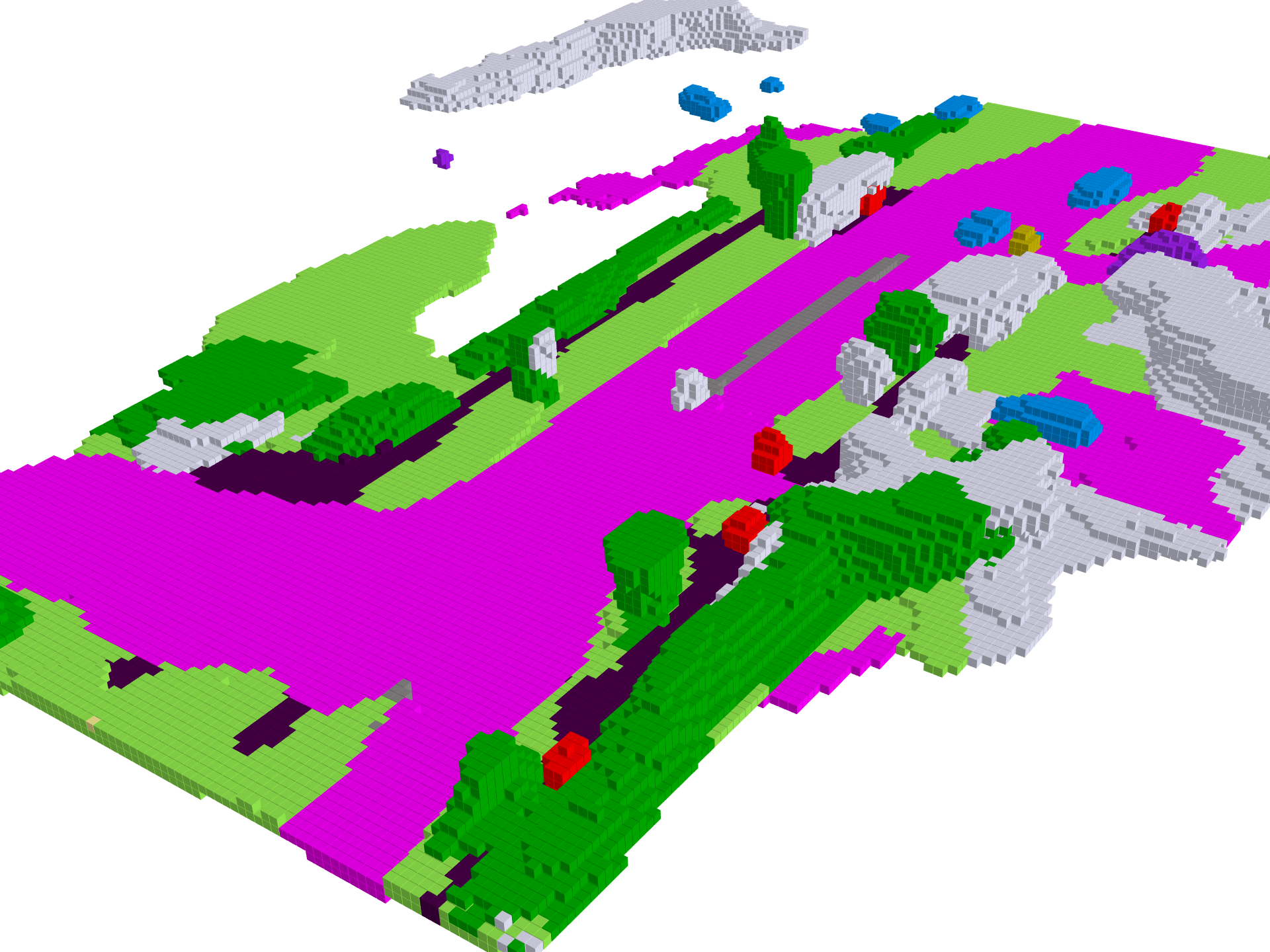}} &
			{\includegraphics[width=0.25\linewidth]{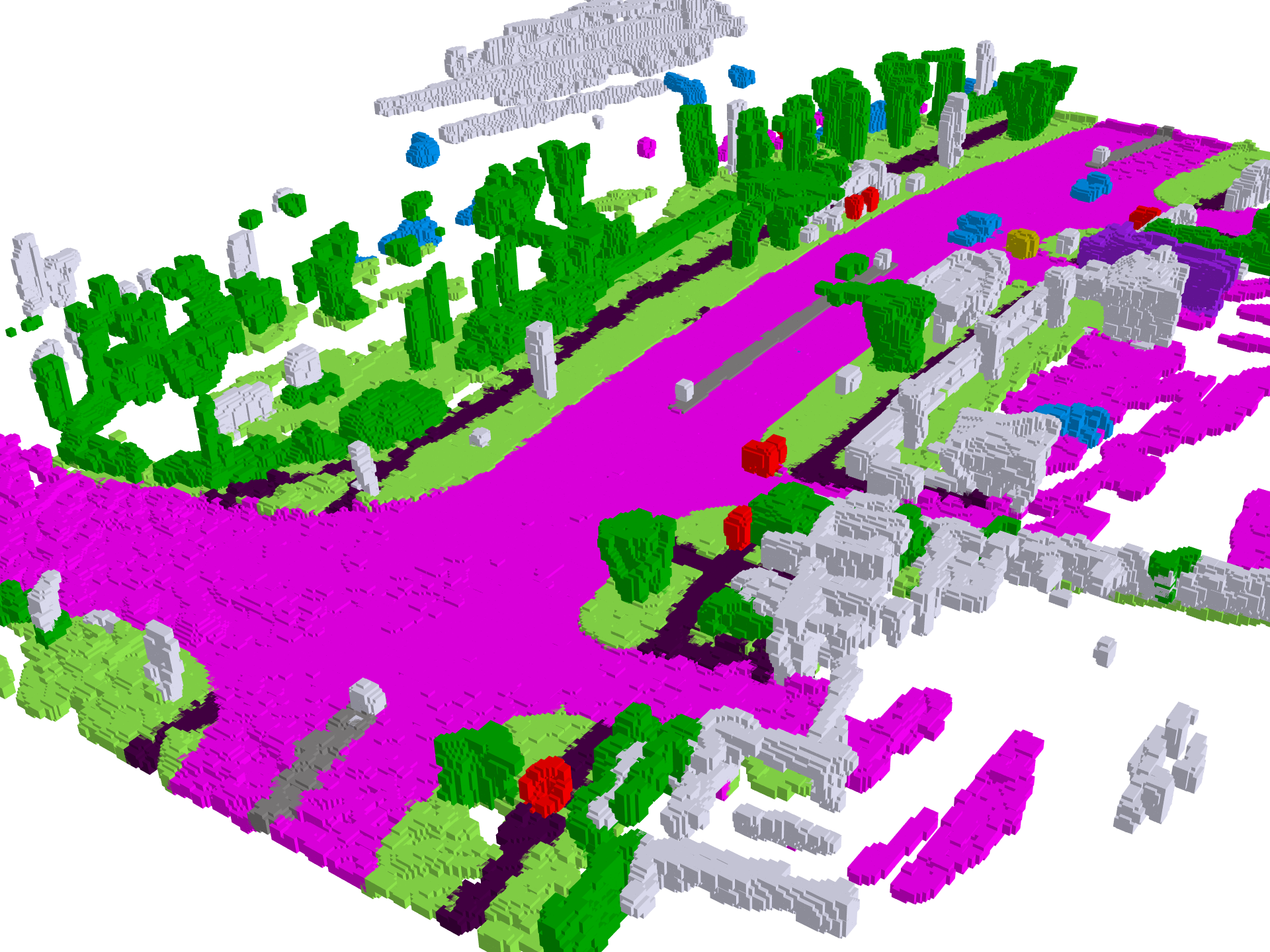}} \\
		\end{tabular}
		
		\begin{tabular}{cccccc}
			{\includegraphics[width=0.166\linewidth]{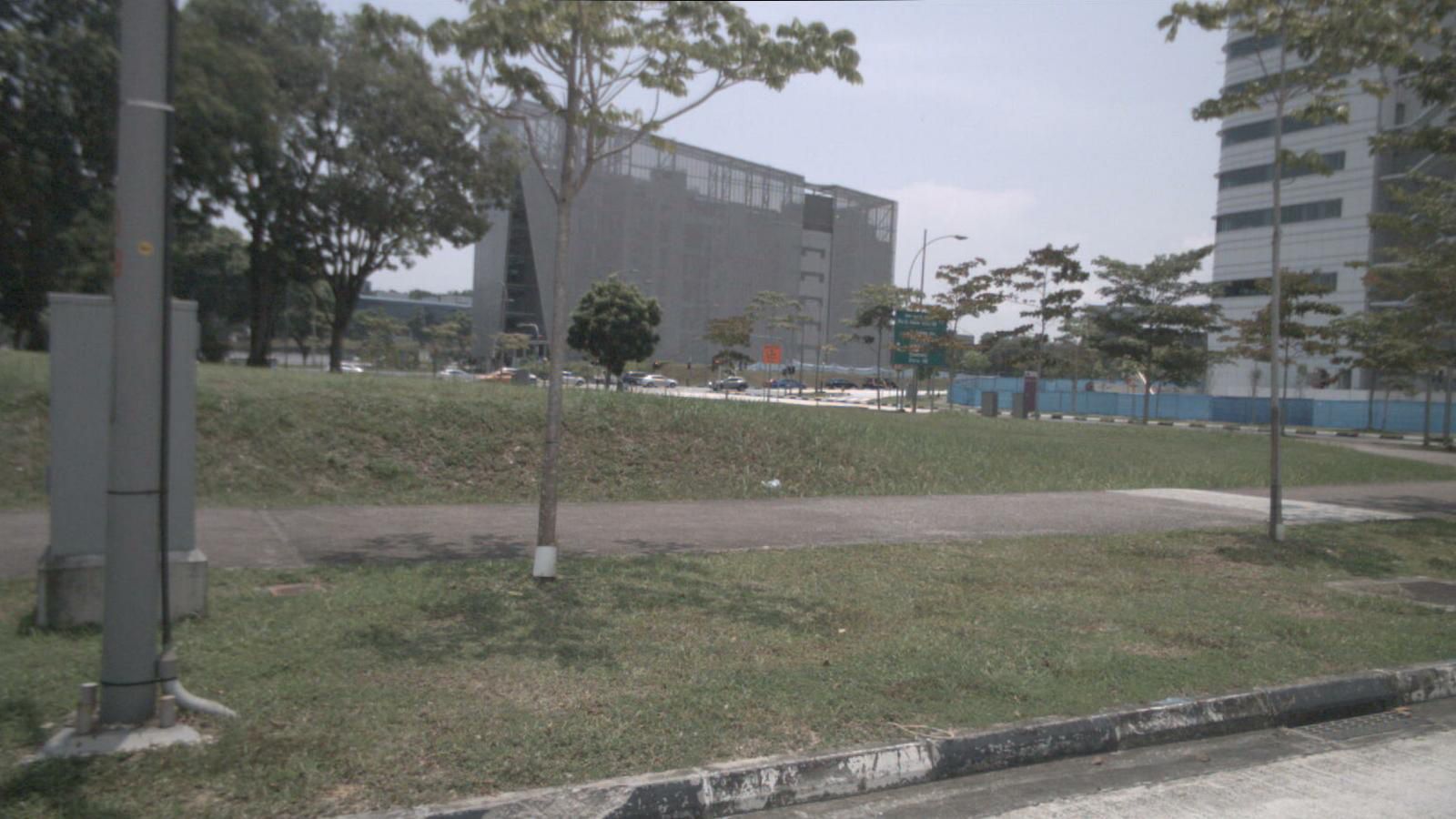}} &
			{\includegraphics[width=0.166\linewidth]{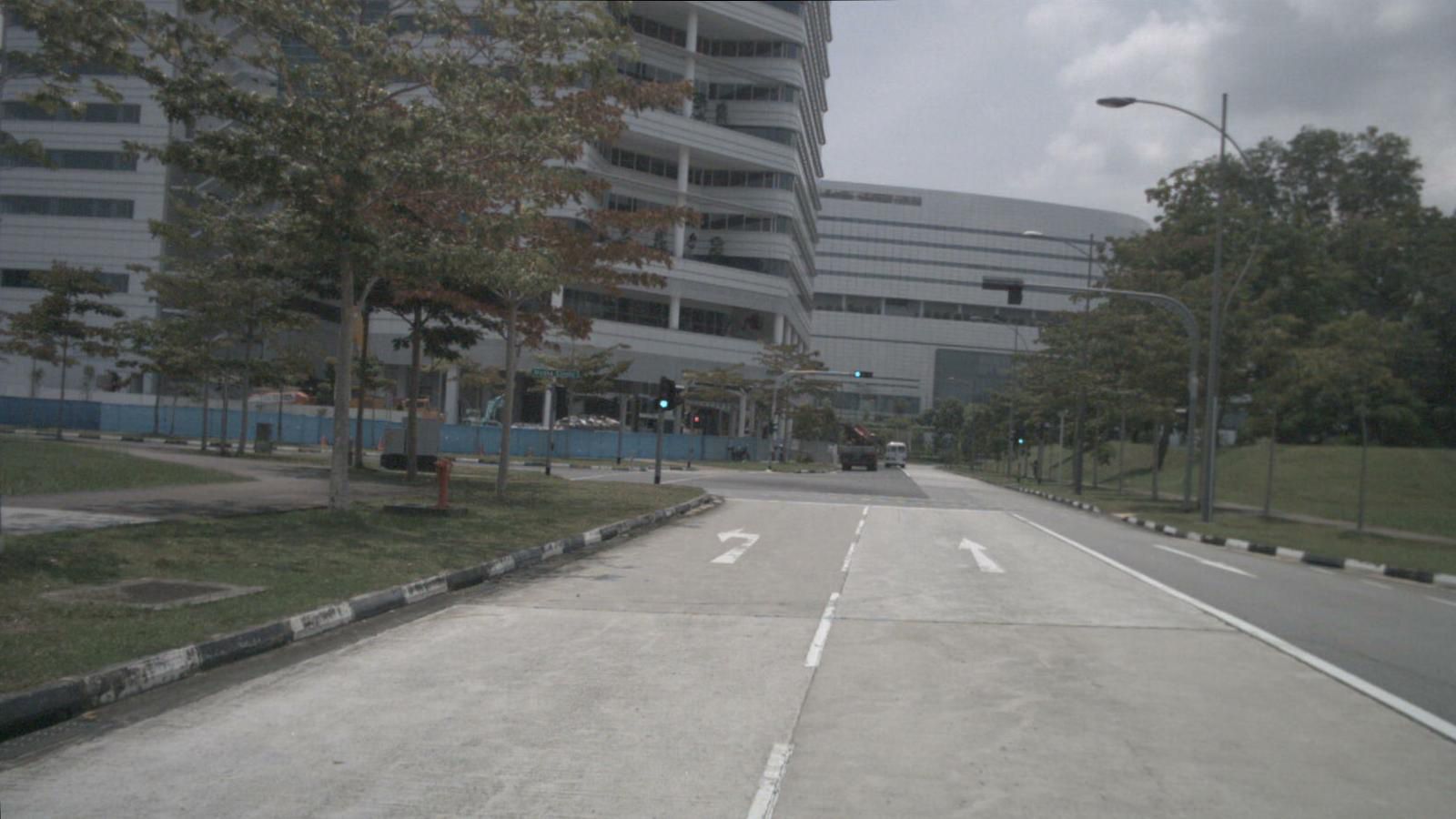}} &
			{\includegraphics[width=0.166\linewidth]{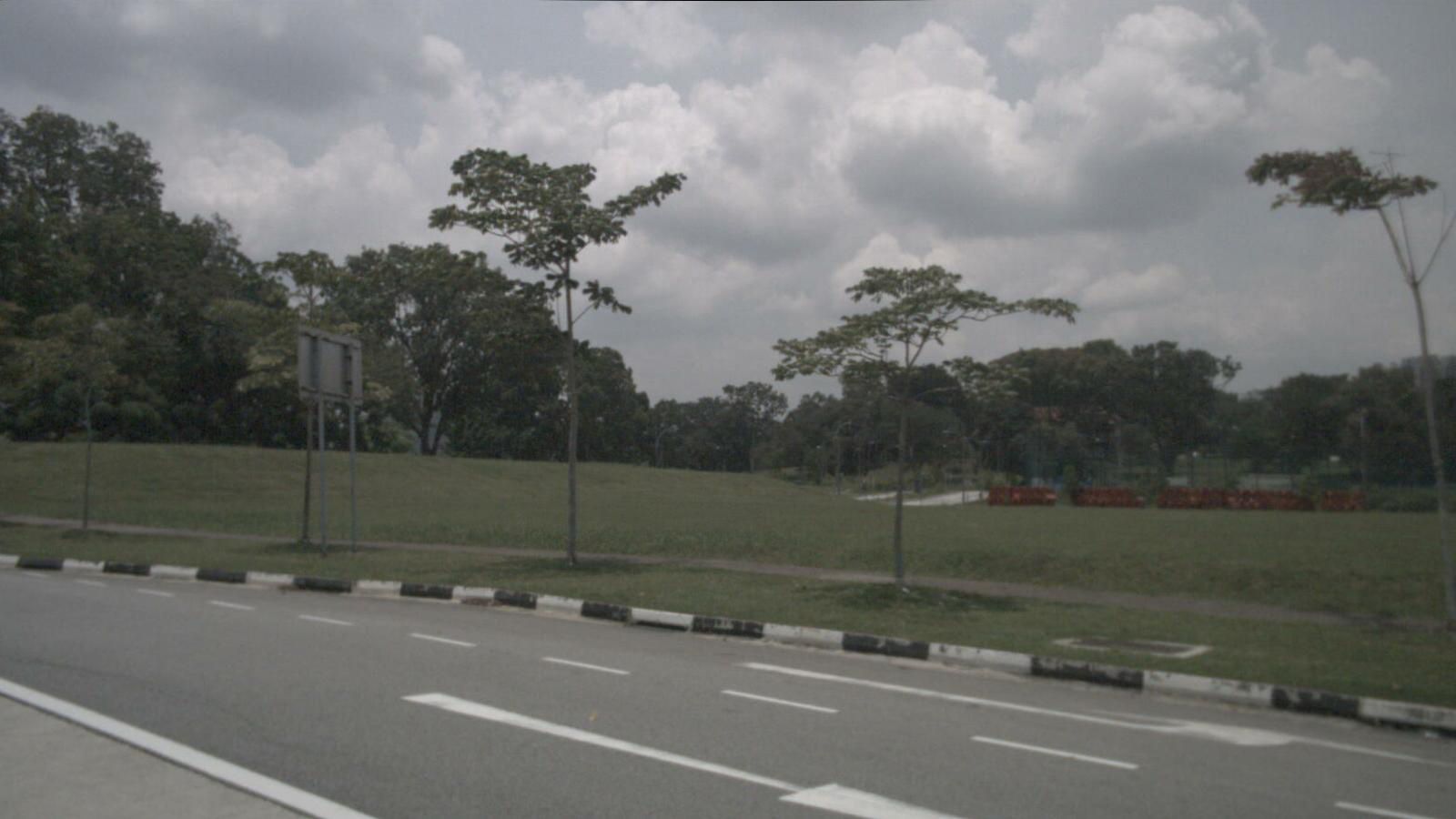}} &
			{\includegraphics[width=0.166\linewidth]{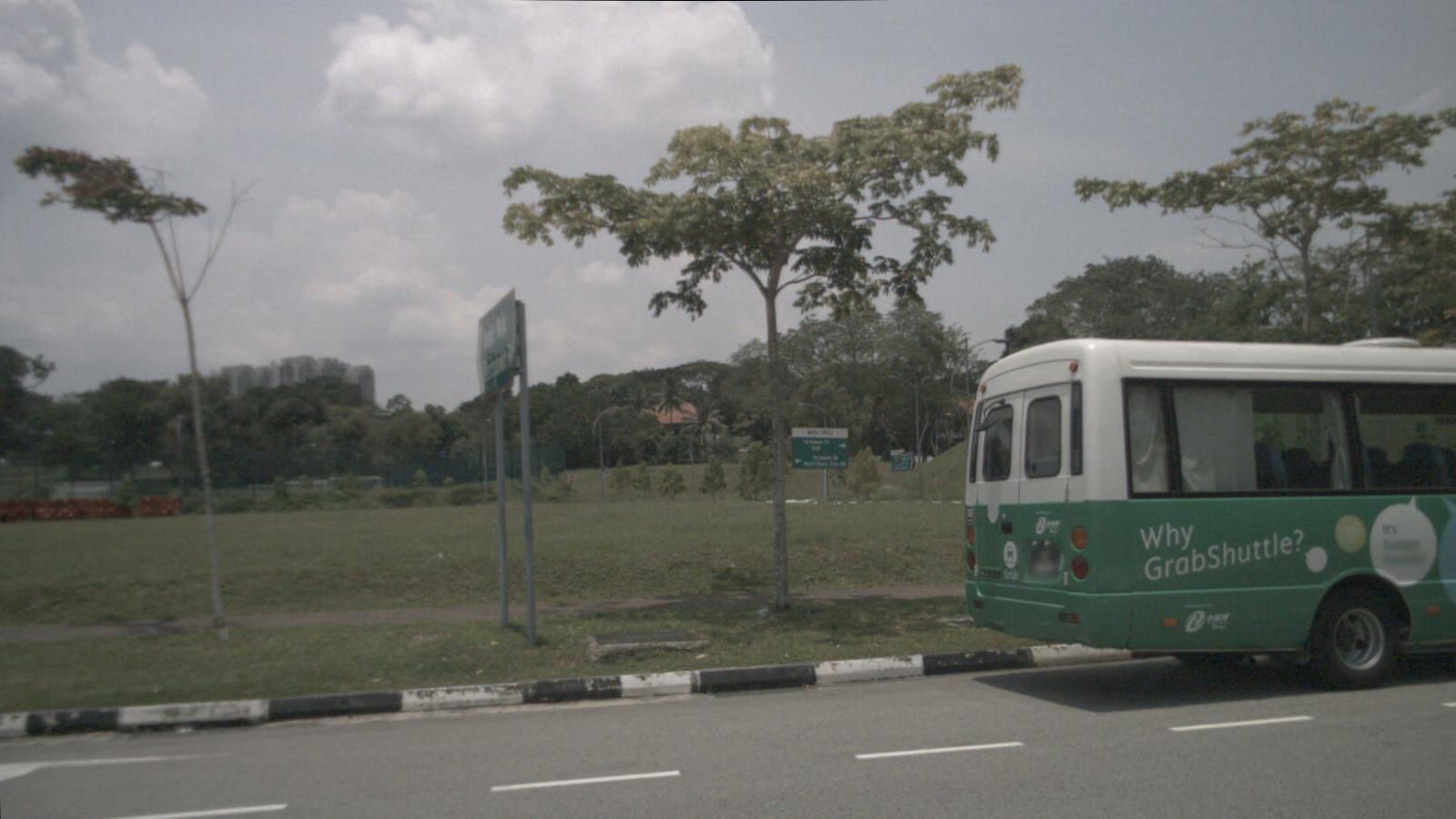}} &
			{\includegraphics[width=0.166\linewidth]{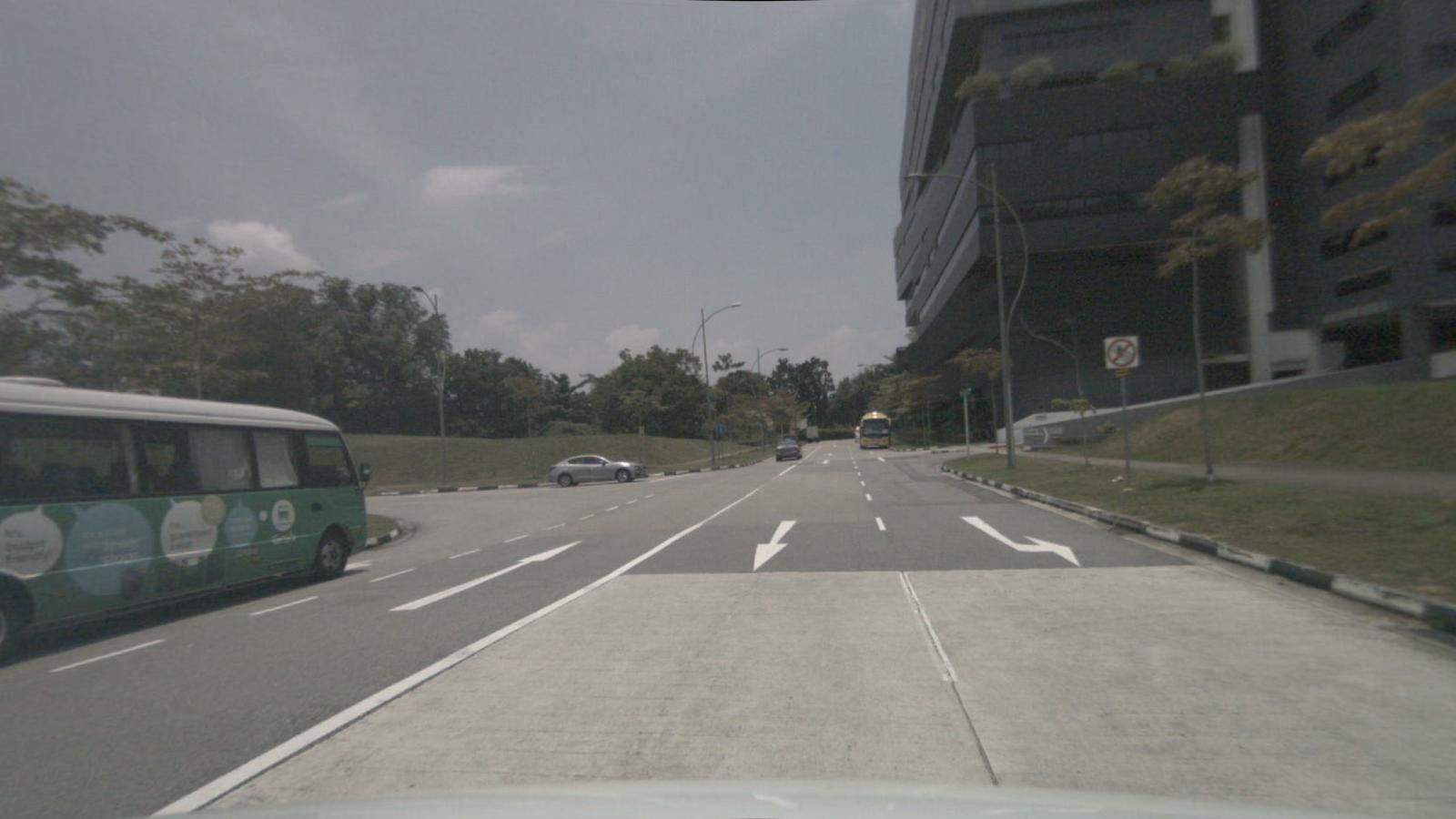}} &
			{\includegraphics[width=0.166\linewidth]{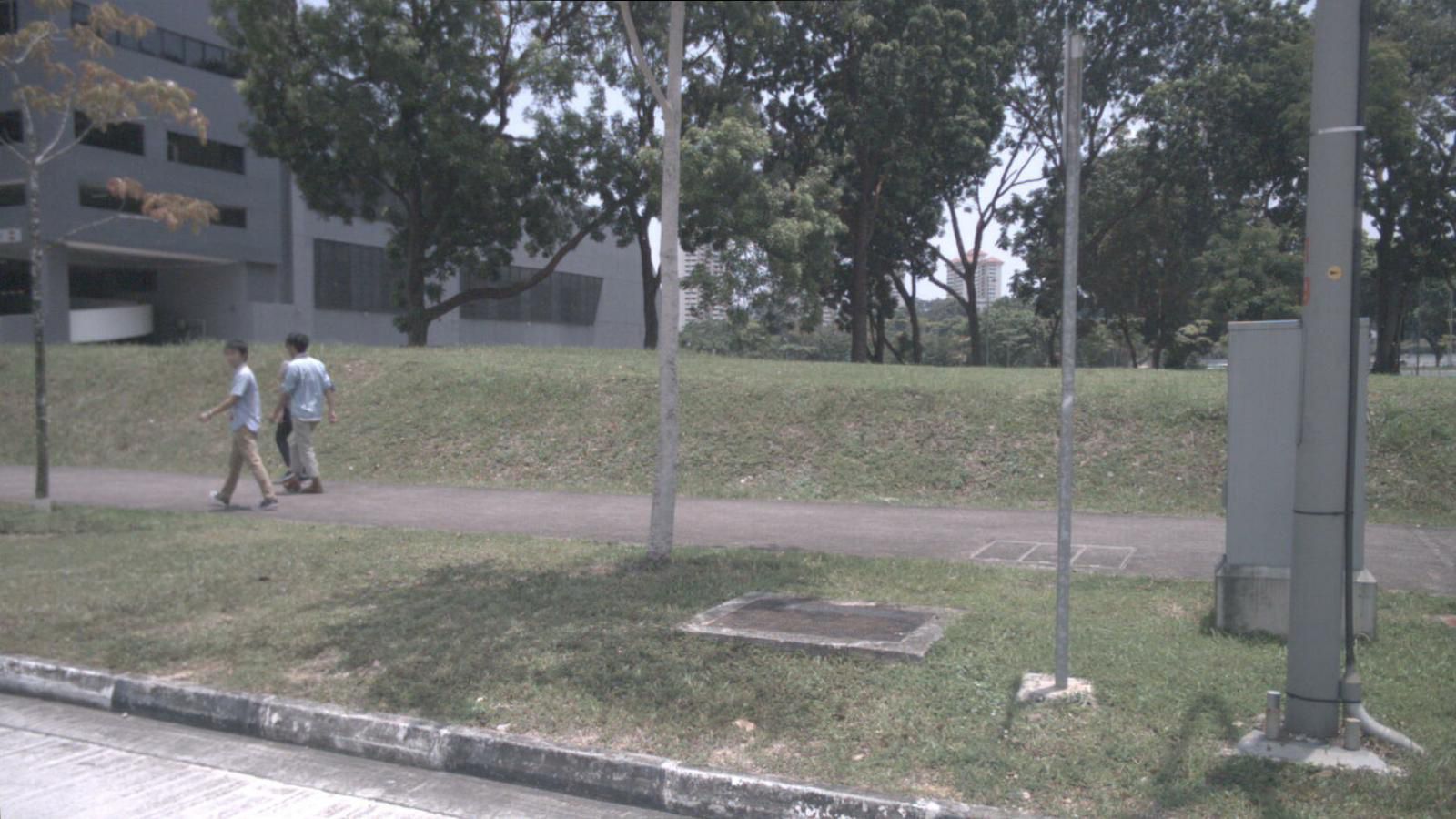}} \\
		\end{tabular}
		
		\begin{tabular}{cccc}
			{\includegraphics[width=0.25\linewidth]{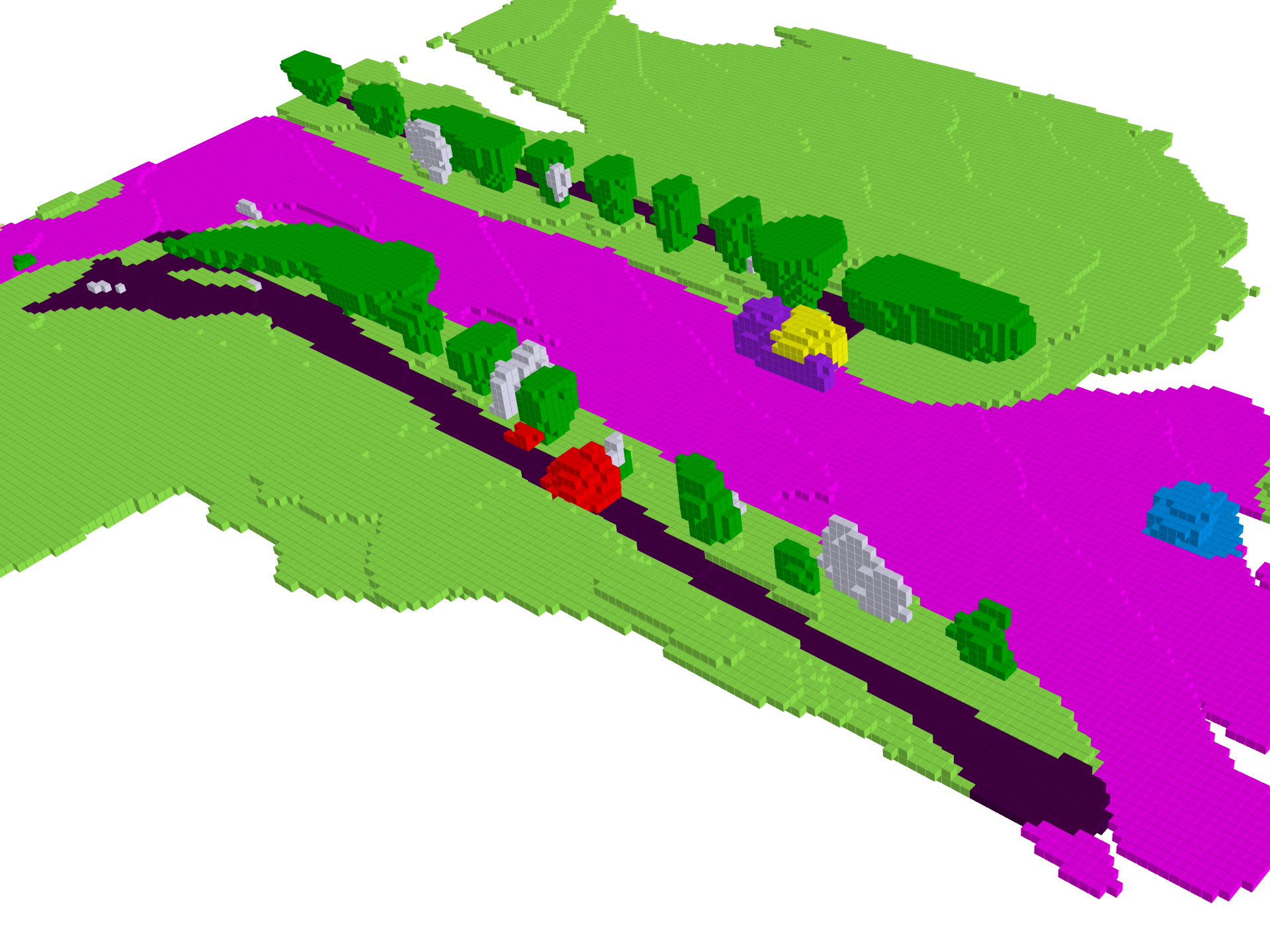}} &
			{\includegraphics[width=0.25\linewidth]{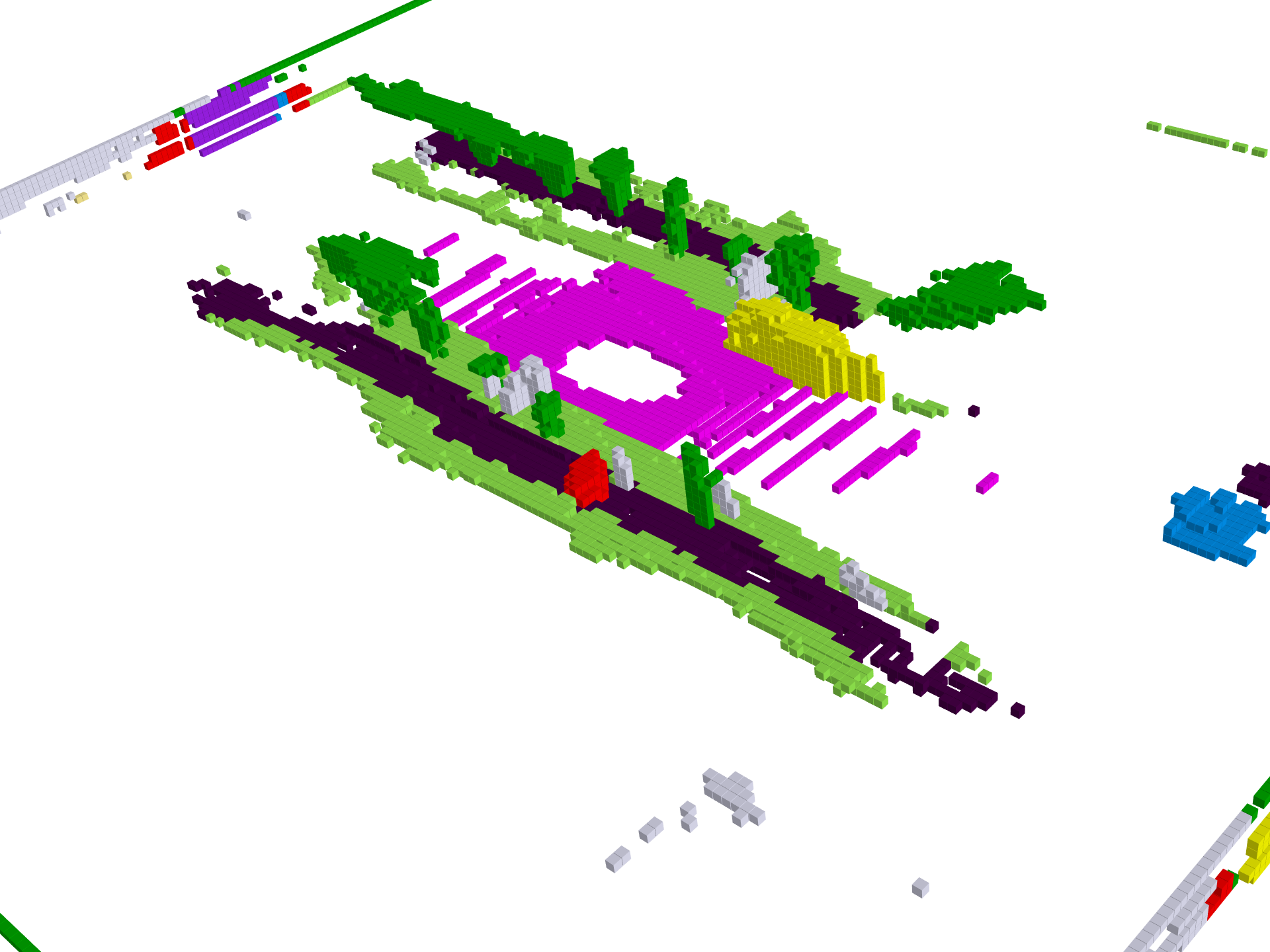}} &
			{\includegraphics[width=0.25\linewidth]{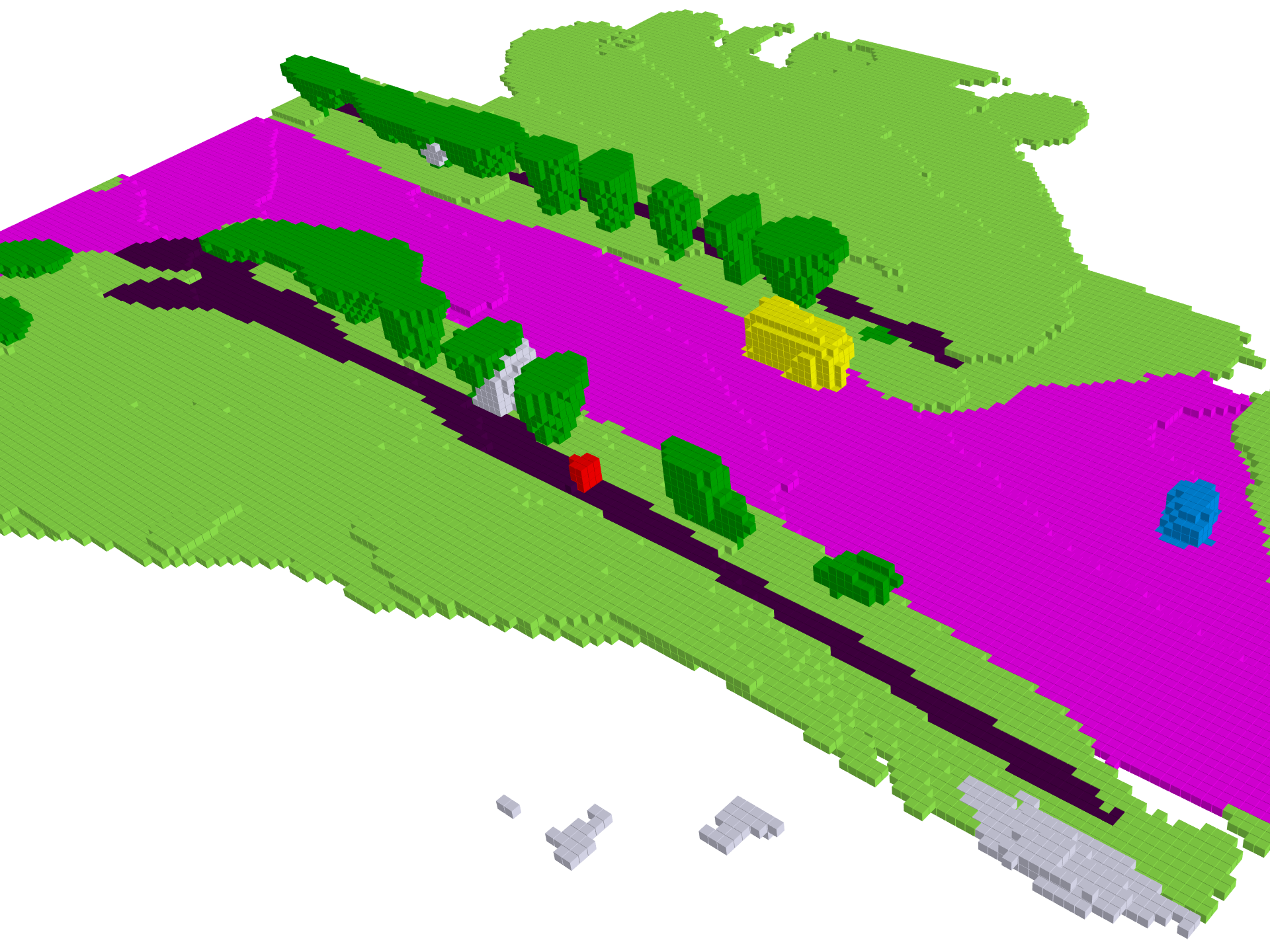}} &
			{\includegraphics[width=0.25\linewidth]{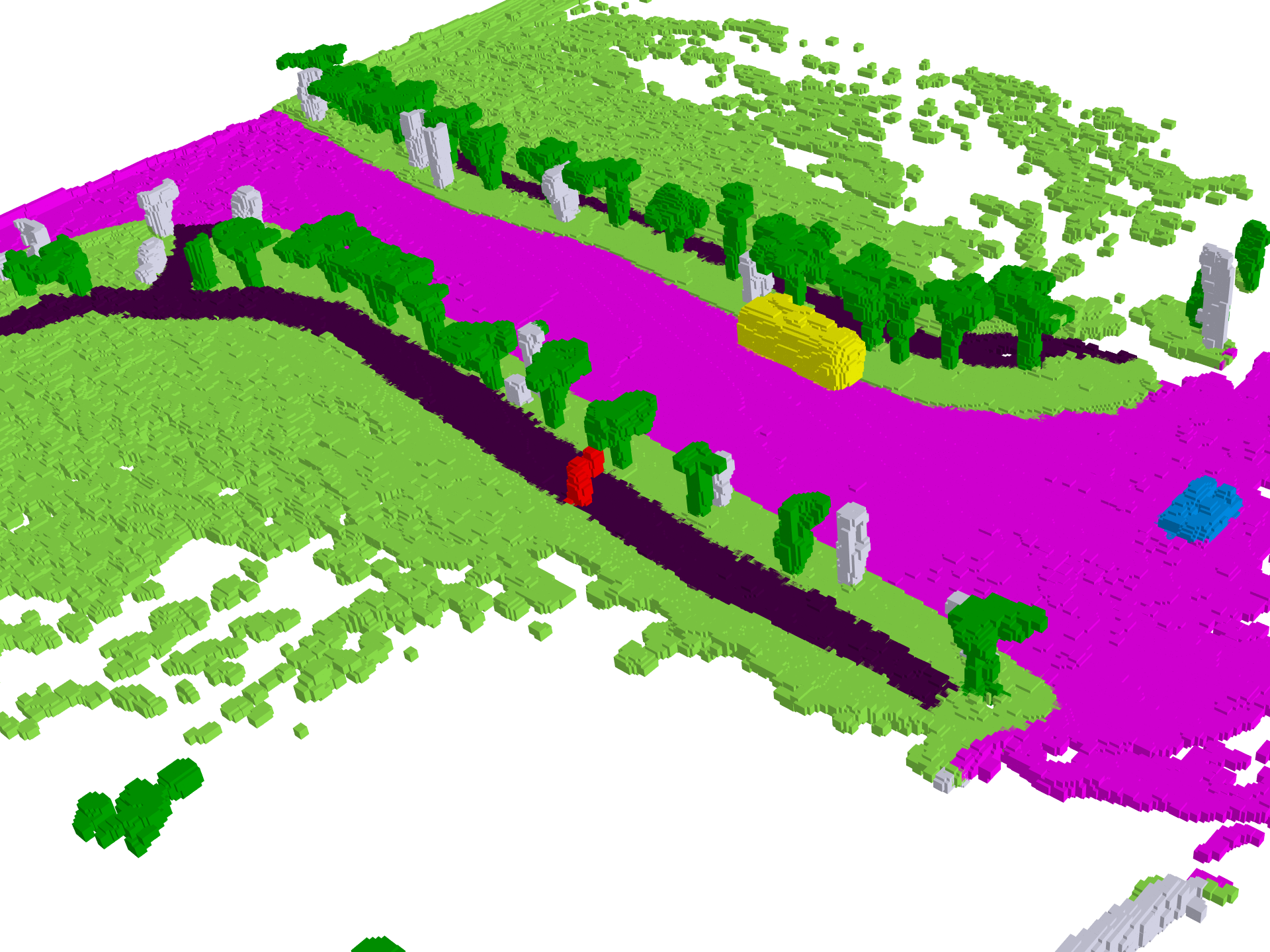}} \\
		\end{tabular}

		\begin{tabular}{cccccc}
			{\includegraphics[width=0.166\linewidth]{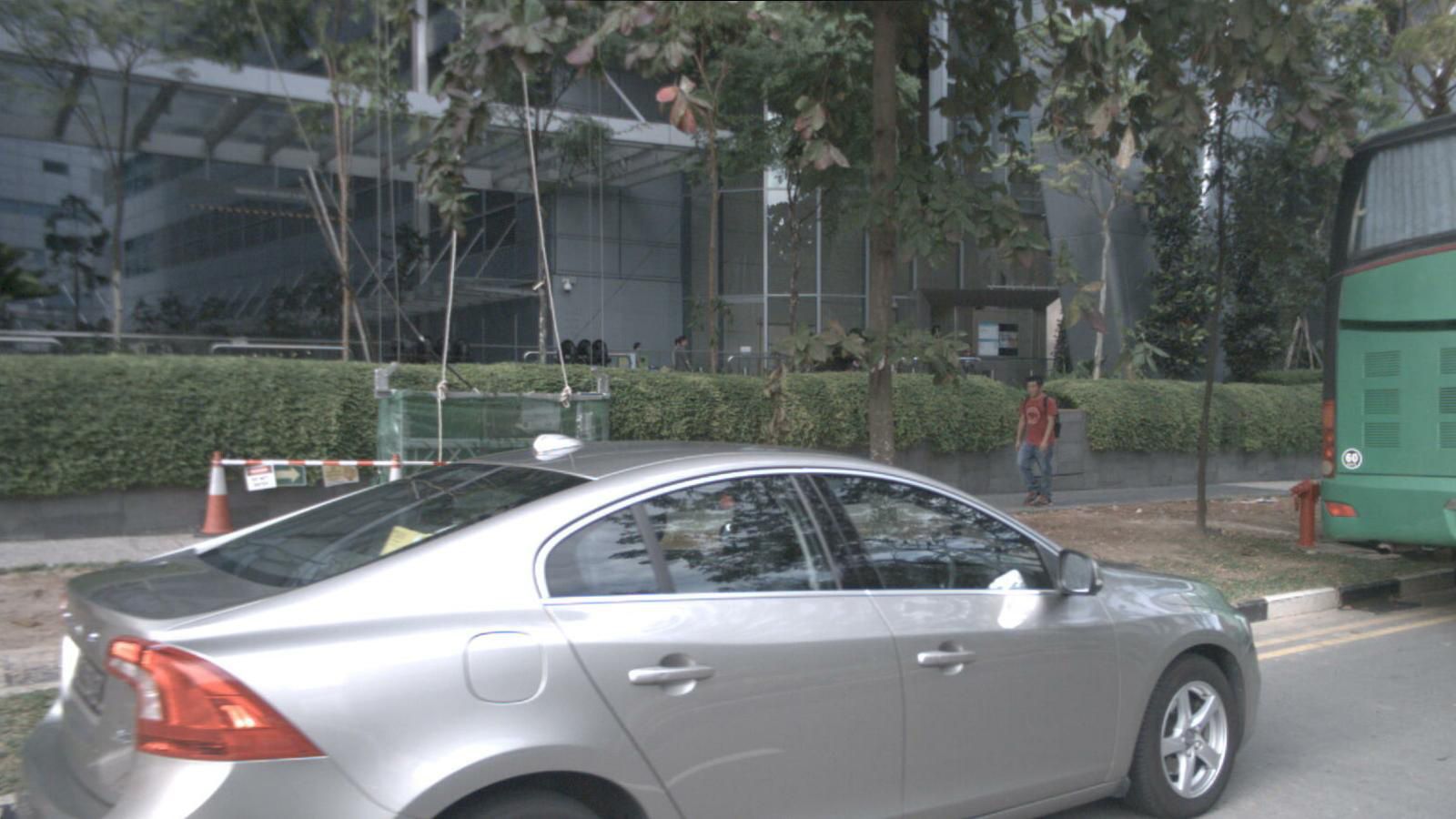}} &
			{\includegraphics[width=0.166\linewidth]{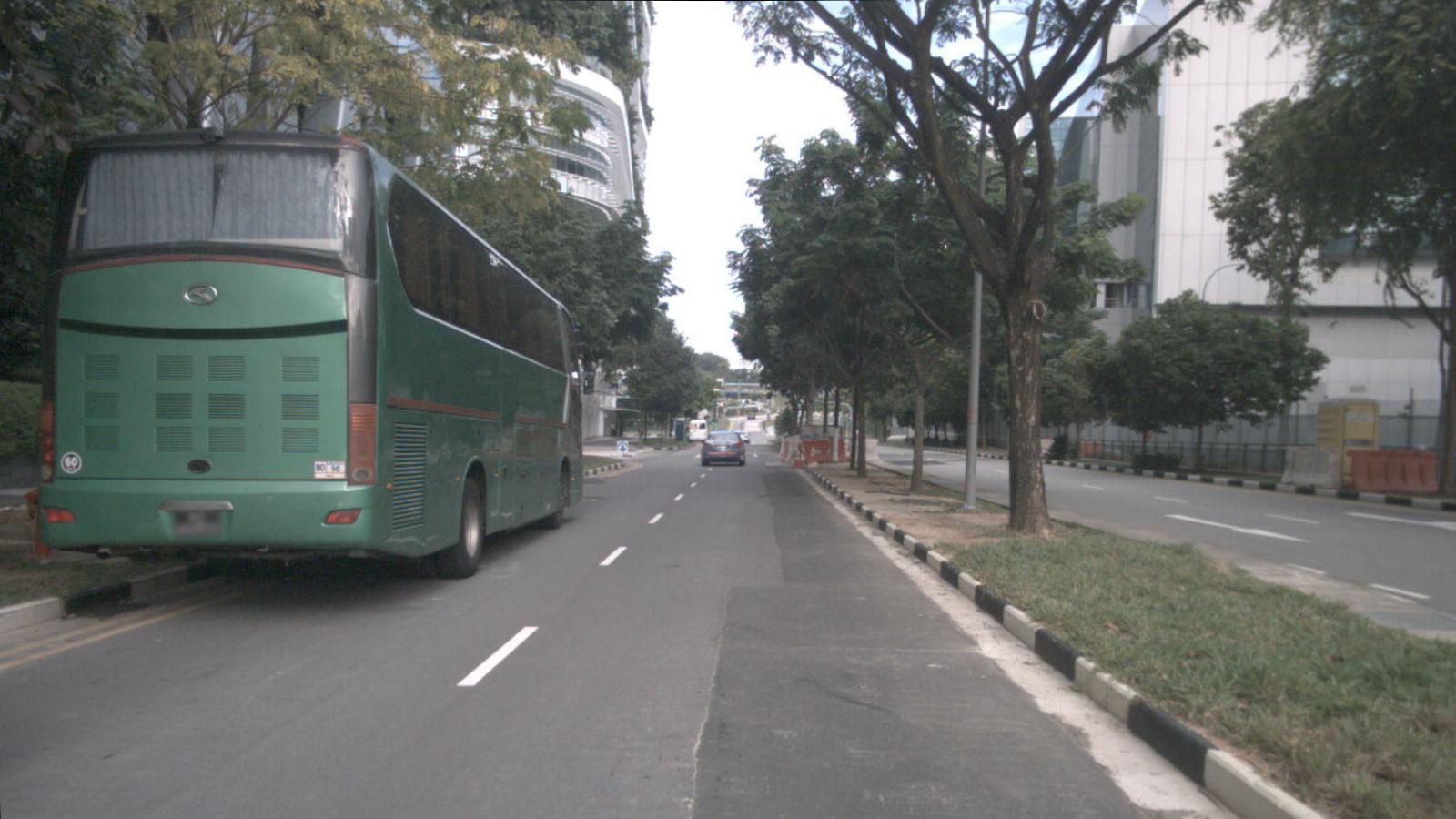}} &
			{\includegraphics[width=0.166\linewidth]{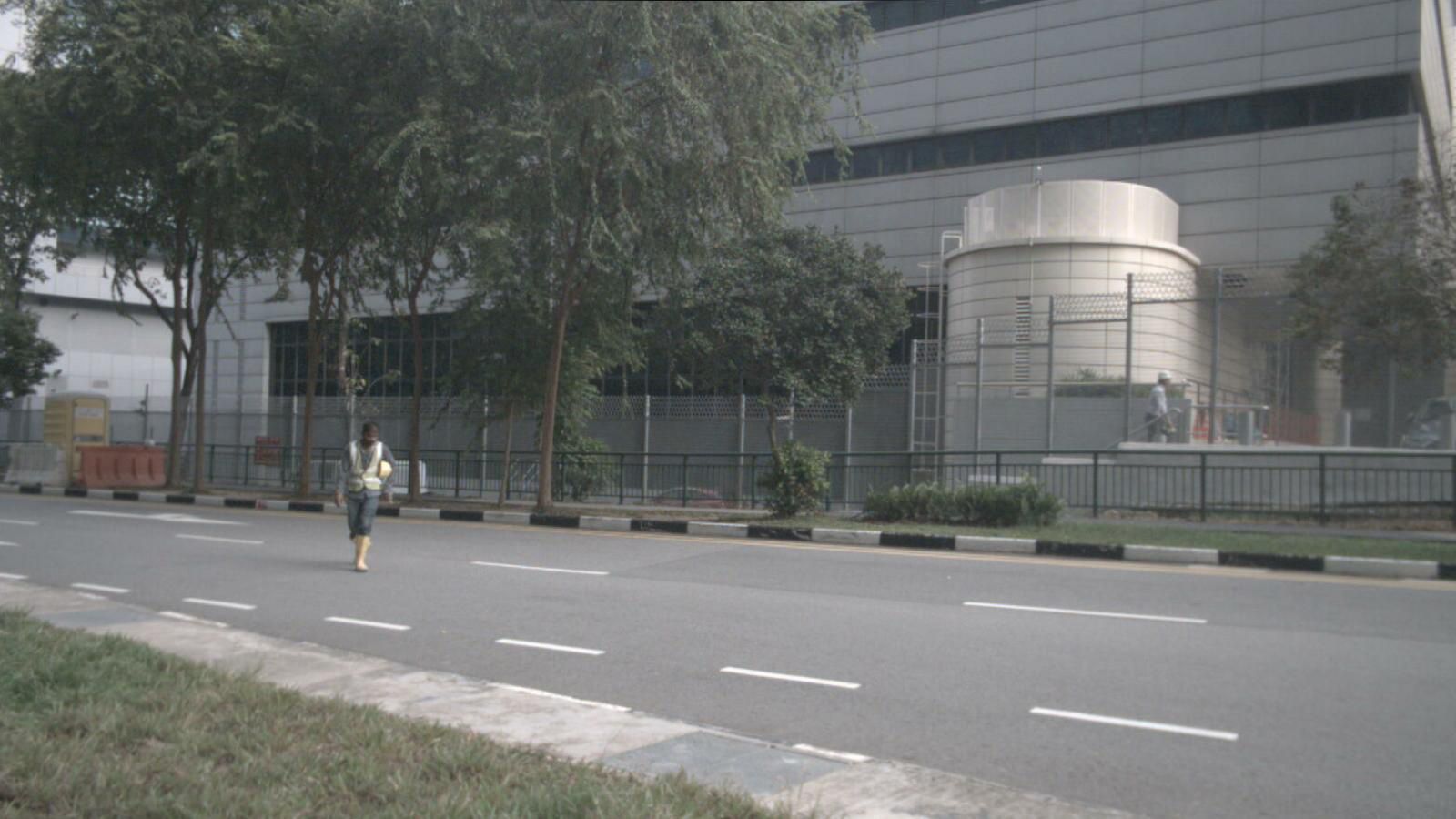}} &
			{\includegraphics[width=0.166\linewidth]{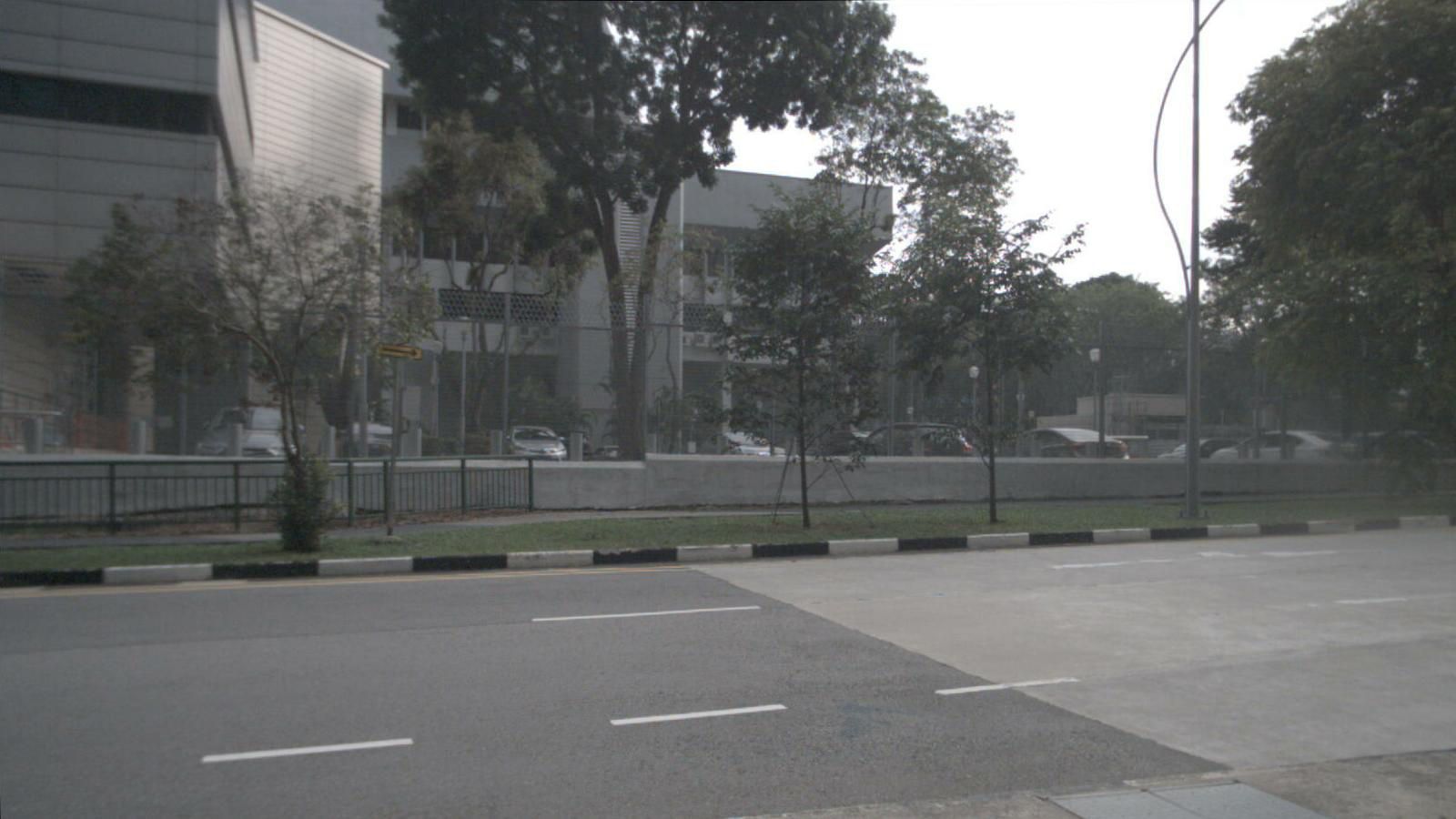}} &
			{\includegraphics[width=0.166\linewidth]{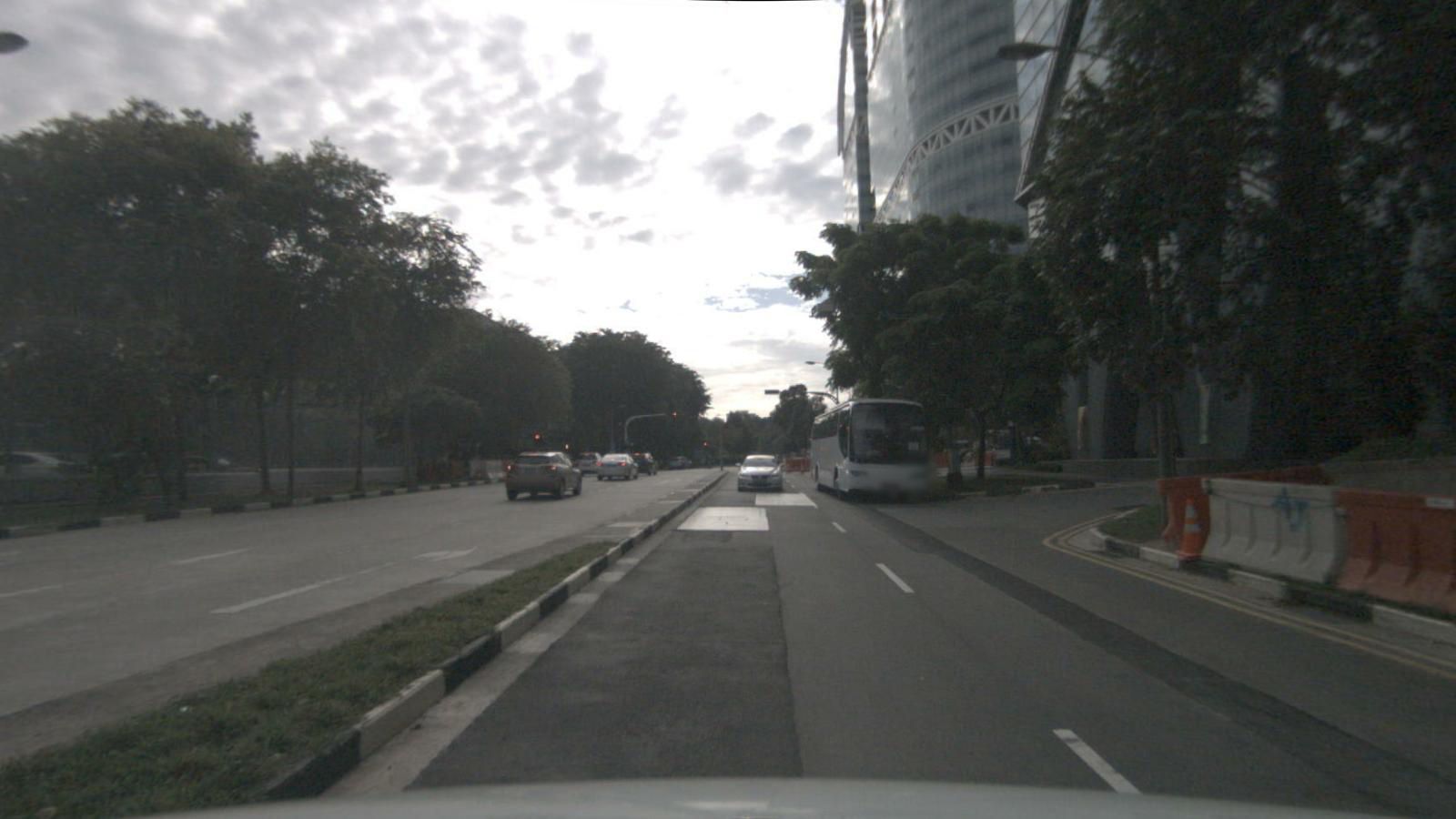}} &
			{\includegraphics[width=0.166\linewidth]{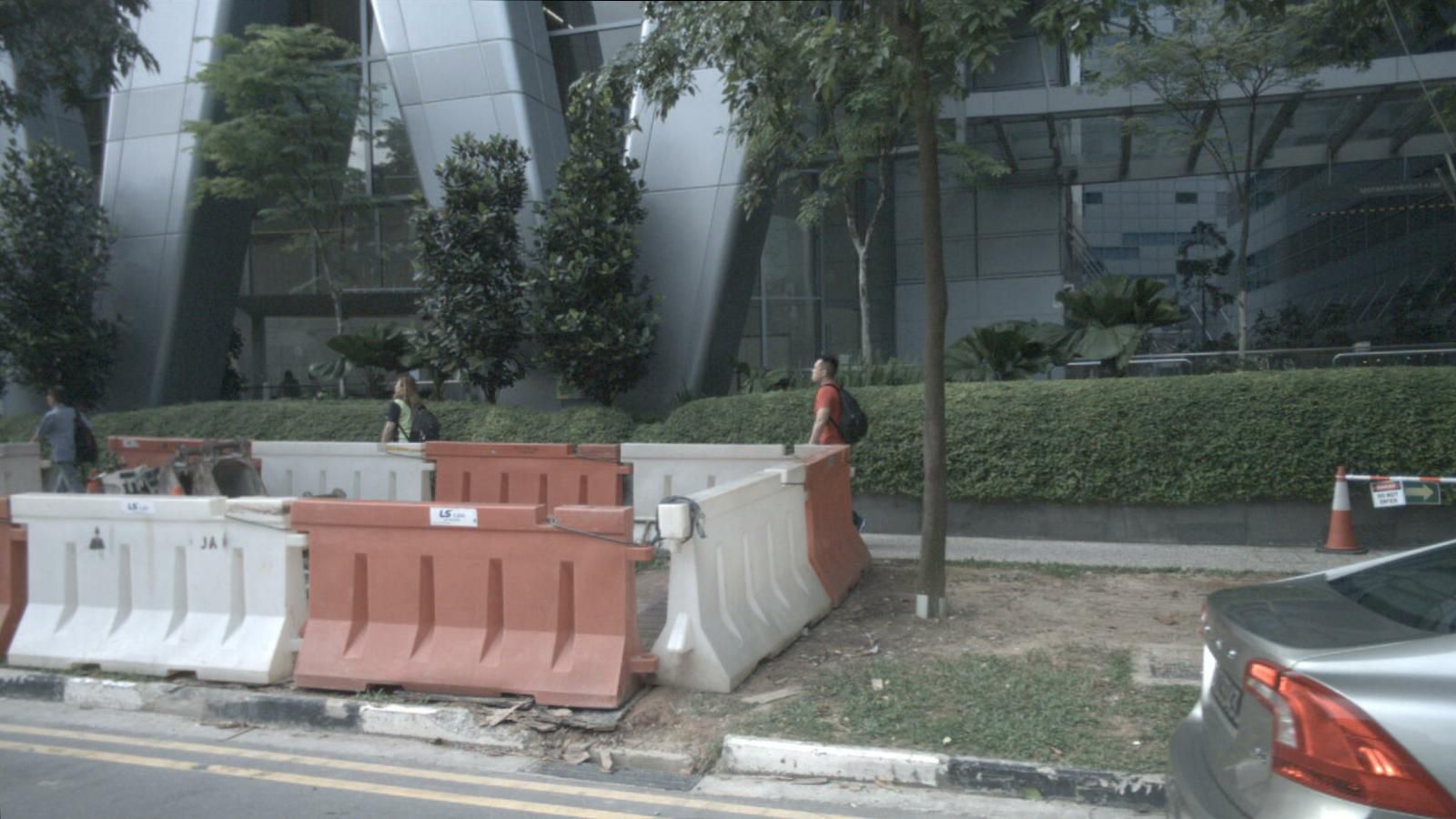}} \\
		\end{tabular}
		
		\begin{tabular}{cccc}
			{\includegraphics[width=0.25\linewidth]{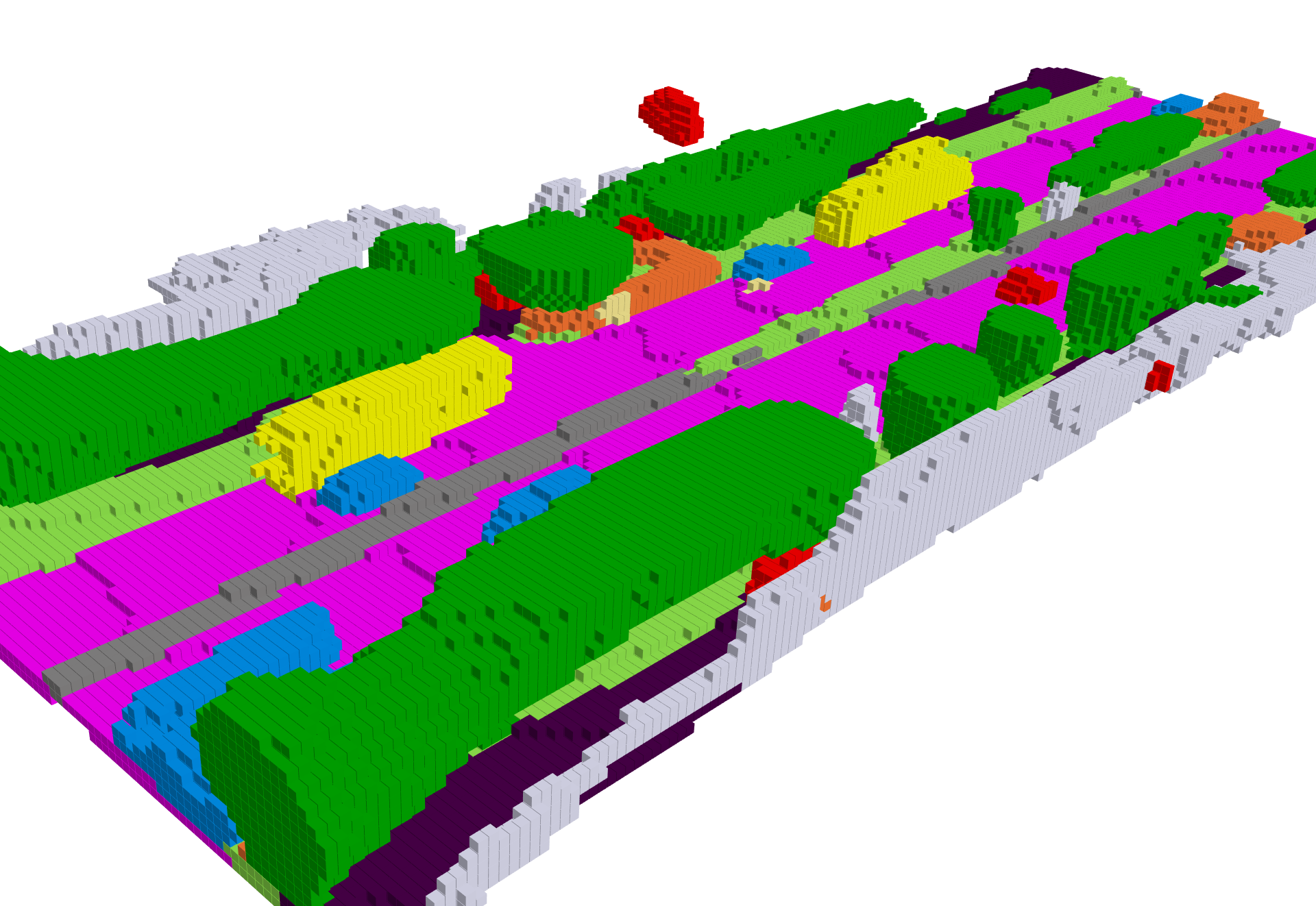}} &
			{\includegraphics[width=0.25\linewidth]{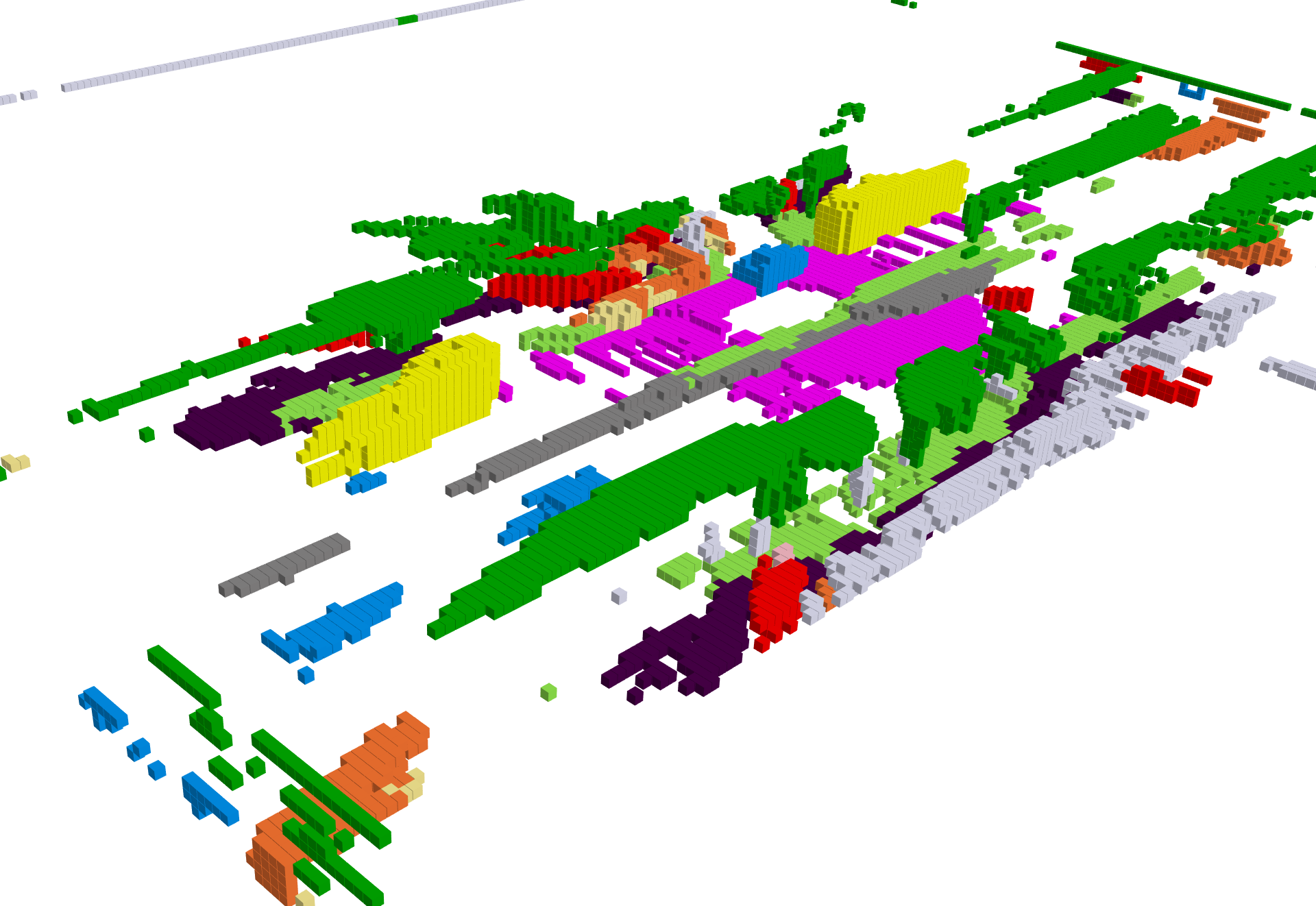}} &
			{\includegraphics[width=0.25\linewidth]{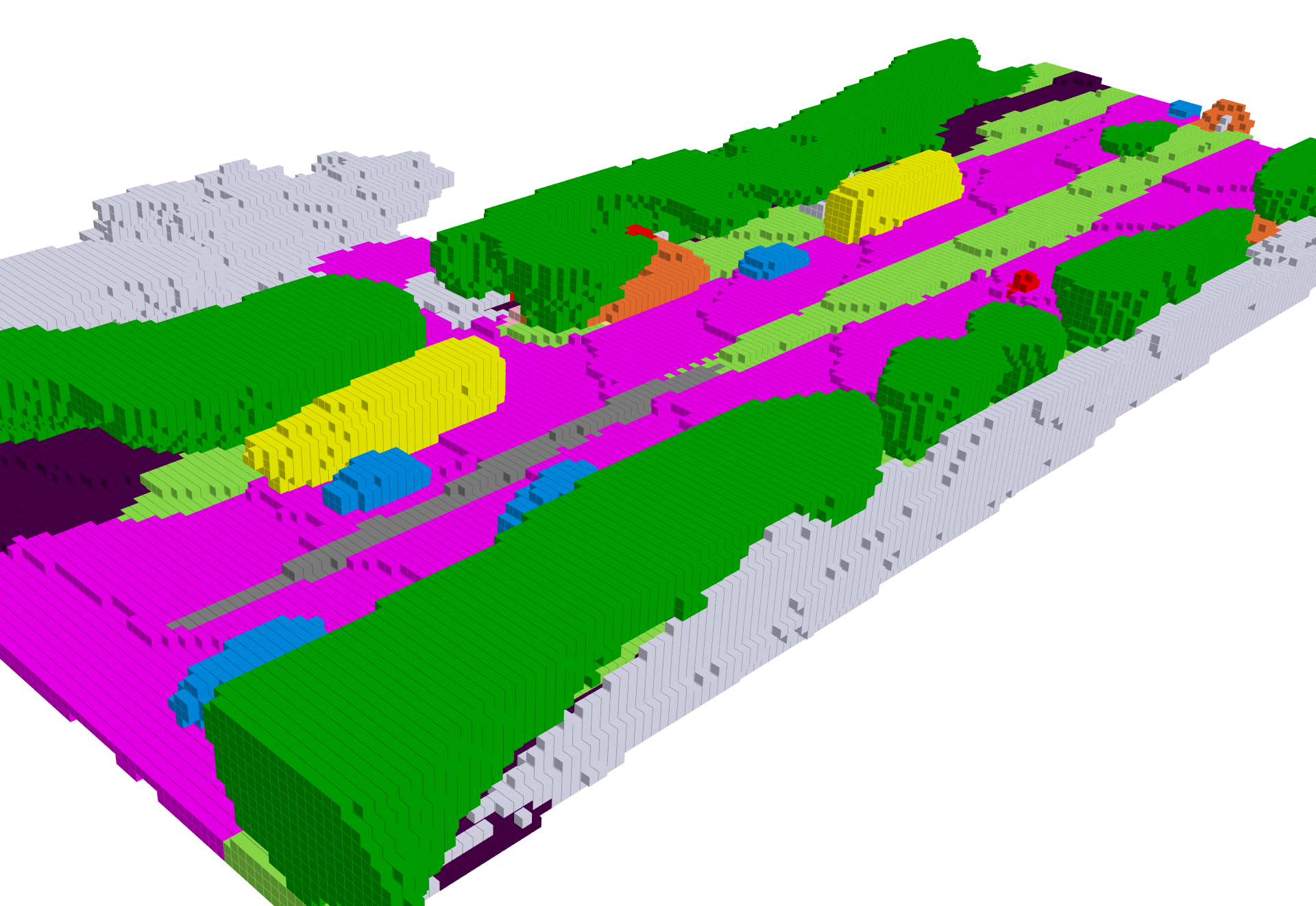}} &
			{\includegraphics[width=0.25\linewidth]{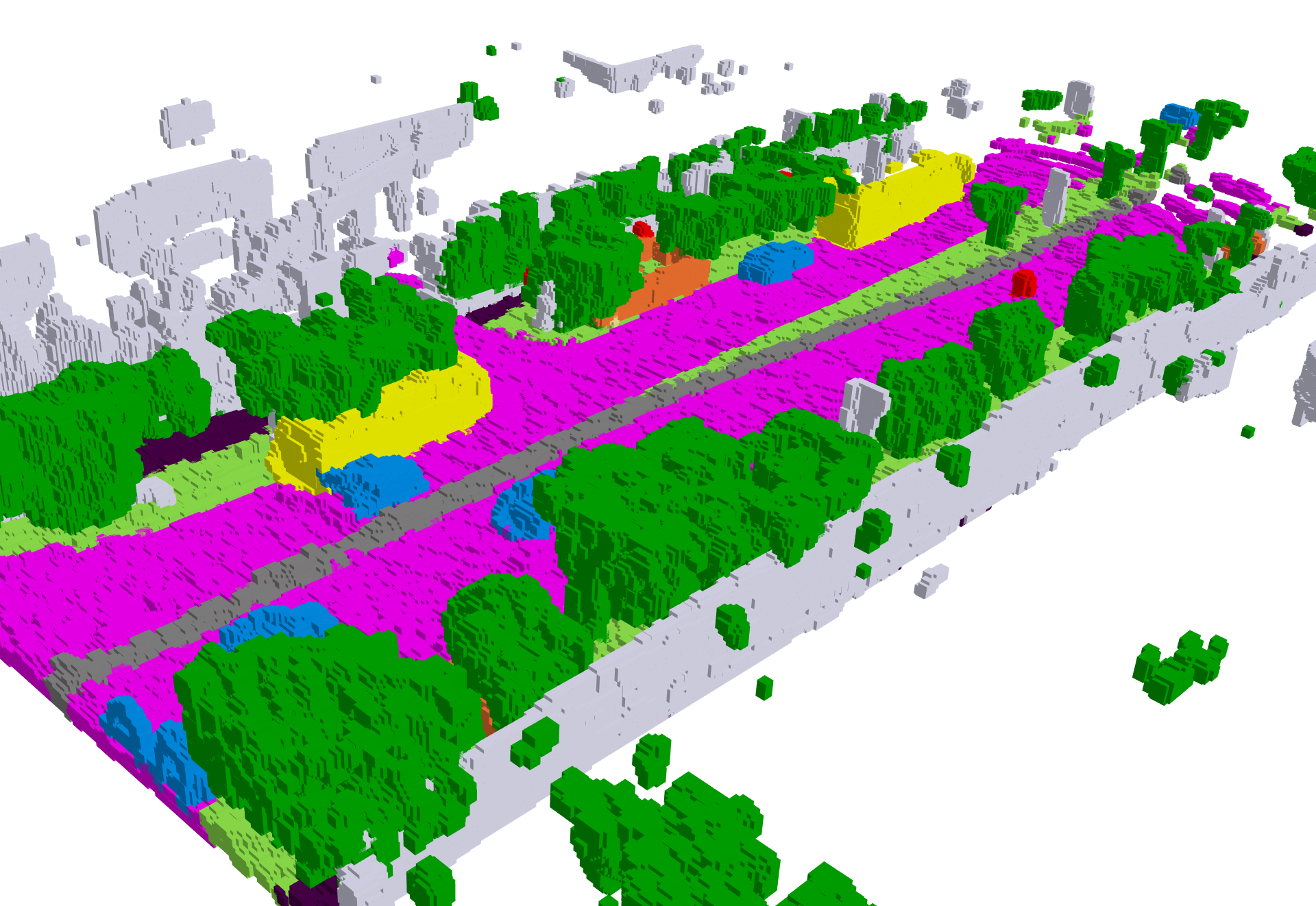}} \\
			BEVFormer \cite{bevformer} &  TPVFormer \cite{huang2023tri} & Ours & Dense occupancy labels \\
		\end{tabular}
		
		\vspace{2mm}
		\centering
		\caption{Qualitaative comparison on nuScenes validation set. Our mrthod can predict more accurate and denser occupancy. \textbf{Better viewed when zoomed in.}}
		\label{fig:quali_compa}
		
	\end{figure*}

	Figure \ref{fig:quali_compa} shows the qualitative comparison with other methods. We can see that our predictions are more accurate and denser. We also provide some video demos in the material. Specifically, 'demo-nuscenes' shows the results on nuScenes validation set and 'demo-gt' visualizes our generated groundtruth. 'demo-comparison' illustrates the comparison with other methods and 'demo-wild' shows the occupancy predictions on Beijing street (trained on nuScenes training set).

	\begin{table}[]
		\centering
		\resizebox{0.45\textwidth}{!}{
			\begin{tabular}{l|l}
				\hline
				Acc & $\mbox{mean}_{p \in P}(\min_{p^*\in P^*}||p-p^*||)$ \\
				Comp & $\mbox{mean}_{p^* \in P^*}(\min_{p\in P}||p-p^*||)$ \\
				Prec & $\mbox{mean}_{p \in P}(\min_{p^*\in P^*}||p-p^*||<0.5)$ \\
				Recal & $\mbox{mean}_{p^* \in P^*}(\min_{p\in P}||p-p^*||<0.5)$ \\
				CD & $\text{Acc} + \text{Comp}$ \\
				F-score & $(2 \times \text{Prec} \times \text{Recal}) / (\text{Prec} + \text{Recal})$ \\
				\hline
		\end{tabular}}
		\vspace{1mm}
		
		\caption{Evaluation metrics for 3D scene reconstruction. $p$ and $p^*$ are the predicted and ground truth point clouds.
		}
		\vspace{-3mm}
		\label{tab:metric_defs}
	\end{table}
	
	{\small
		\bibliographystyle{ieee_fullname}
		\bibliography{egbib}
	}
	
\end{document}